# A survey of Bayesian Network structure learning

Neville Kenneth Kitson[1], Anthony C. Constantinou[1, 2], Zhigao Guo[1], Yang Liu[1], and Kiattikun Chobtham[1]

1.  Bayesian Artificial Intelligence research lab, Risk and Information Management (RIM) research group, School of Electronic Engineering and Computer Science, Queen Mary University of London (QMUL), London, UK, E1 4NS.

    E-mail addresses: n.k.kitson@qmul.ac.uk (N.K. Kitson), a.constantinou@qmul.ac.uk (A. Constantinou), zhigao.guo@qmul.ac.uk (Z. Guo), yangliu@qmul.ac.uk (Y. Liu), and k.chobtham@qmul.ac.uk (K. Chobtham).

2.  The Alan Turing Institute, British Library, 96 Euston Road, London, UK, NW1 2DB, UK.

**ABSTRACT:** Bayesian Networks (BNs) have become increasingly popular over the last few decades as a tool for reasoning under uncertainty in fields as diverse as medicine, biology, epidemiology, economics and the social sciences. This is especially true in real-world areas where we seek to answer complex questions based on hypothetical evidence to determine actions for intervention. However, determining the graphical structure of a BN remains a major challenge, especially when modelling a problem under causal assumptions. Solutions to this problem include the automated discovery of BN graphs from data, constructing them based on expert knowledge, or a combination of the two. This paper provides a comprehensive review of combinatoric algorithms proposed for learning BN structure from data, describing 74 algorithms including prototypical, well-established and state-of-the-art approaches. The basic approach of each algorithm is described in consistent terms, and the similarities and differences between them highlighted. Methods of evaluating algorithms and their comparative performance are discussed including the consistency of claims made in the literature. Approaches for dealing with data noise in real-world datasets and incorporating expert knowledge into the learning process are also covered.

## 1. INTRODUCTION

The achievements of black-box machine learning, such as neural networks, are undeniable and have contributed to a renewed interest in machine learning and artificial intelligence in general. Nevertheless, it is now well understood that black-box solutions that are restricted to predictive optimisation are unsuitable to inform decision making in domains that require transparency and tractability, such as in government policy and healthcare. The recent book by Pearl and Mackenzie (2018) highlights the need for models to be capable of reasoning under causal representation, in order to offer solutions that go beyond prediction. They illustrate this by presenting a ladder of causation that consists of three levels:

- Level 1: Models restricted to associational relationships, or are capable of generating predictions only; e.g., "What symptoms should we expect to observe given disease $A$?".
- Level 2: Models that involve some form of causal representation and can answer questions about interventions; e.g., "What effect would taking drug $A$ have on symptoms $B$ given that they are caused by disease $C$?".
- Level 3: Models that offer a complete form of causal representation and can answer questions about causation that extend to counterfactual reasoning; e.g., "If I had taken drug $B$ instead of drug $A$, would my symptoms caused by disease $C$ be less severe?"

Pearl and Mackenzie also refer to the three above levels as seeing, doing and imagining, respectively. The Bayesian network (BN) framework that Pearl described a few decades back (Pearl, 1985, 1988) enables us to answer questions up to and including Level 3, although this requires that the BN model is employed under causal assumptions; also referred to as a 'causal BN'.

A BN is a probabilistic graphical model which provides a powerful general approach especially suited to modelling complex non-deterministic systems. A BN offers a compact probabilistic representation of the system and provides a means of applying probability theory to large collections, sometimes thousands or more, of variables. They have been used in many different domains, for example, in protein (Sachs et. al., 2005) and gene (Imoto et. al., 2004) networks in biology, pyschosis (Moffa et. al, 2017) and cancer care (Sesen et. al., 2013) in healthcare, engineering fault diagnosis (Cai et. al., 2017), and air pollution modelling (Vitolo et. al., 2018). Koller and Friedman (2009) and Darwiche (2009) provide two excellent introductions to the theory behind, and use of, BNs.

A BN consists of graph which shows the direct dependence relationships between variables, or in a causal BN, direct cause and effect relationships. The BN also defines parameters which specify the form and the strengths of these relationships. Determining these parameters is generally much easier than recovering the graph accurately and so we focus here on the latter. The graph may be specified by human experts in a domain of interest, but here, we describe structure learning algorithms which aim to learn the graph from data. We focus on combinatoric algorithms where the approach is to search or constrain the finite discrete space of possible graphs in some way. This paper aims to provide intuitive descriptions of a comprehensive range of these algorithms from the earliest, but often still competitive, algorithms, to some of the most recent advances.

Inevitably, in a field as broad and rapidly developing as this, we have had to omit, or only briefly refer to, some aspects of structure learning. Fortunately, there are other recent survey papers that cover some of these aspects more completely. For example, the paper by Glymour et al. (2019) provides more coverage of functional causal models where assumptions about the functional form of the effect, causes and noise relationships can be used to deduce causal relationships. Vowels et al. (2021) concentrate on approaches which treat structure learning as a continuous optimisation problem, optimising an objective function and handling the acyclicity constraint as a continuous function, and Zanga et. al. (2022) cover algorithms which learn from mixed observational and experimental data, and those which learn cyclic graphs. Moraffah et al. (2021) and Noguiera et al. (2022) deal with structure learning from time-series data which we do not cover in this paper.

The paper is structured as follows: the next section covers some preliminaries about BNs, Section 3 covers constraint-based learning, Section 4 score-based learning, and Section 5 hybrid learning and some non-combinatorial approaches. Figure 6 provides an overview of the evolution of structure learning algorithms that are covered in this paper, and will be referenced in subsequent sections. Section 6 covers various practical considerations when applying these algorithms to synthetic and real data, including how to evaluate their output, as well as a discussion of comparative reviews of algorithm performance. Section 6 also discusses some of the main approaches these algorithms may incorporate to handle noise in the data, methods for incorporating expert knowledge into the structure learning process, some open-source software packages, and some guidelines on choosing algorithms. Lastly, we provide our concluding remarks in Section 7.

## 2. PRELIMINARIES

A Bayesian Network, $\mathcal{B}$ , is defined by a tuple consisting of a Directed Acyclic Graph (DAG) $G$, and a set of parameters $\mathbf{\Theta}$, defining the strength and the shape of the relationships between variables (we shall denote sets in boldface throughout):

$$\mathcal{B} = (G, \mathbf{\Theta})$$

The DAG, $G$, consists of a set of *nodes* (also known as *vertices*) $\boldsymbol{X}$, each of which corresponds to one of the $n$ variables under consideration, $\boldsymbol{X} = \{X_1, \dots, X_n\}$, and a set of *directed edges* (or *arcs*) $\boldsymbol{E}$, so that:

$$G = (\boldsymbol{X}, \boldsymbol{E})$$

We will use plain capital letters to represent individual variables or nodes, e.g. $A, Y, X_i$ . A directed edge, for example $A \longrightarrow B$, represents a direct conditional relationship between $A$ and $B$, or under a causal assumption, means that $A$ is a *direct cause* of $B$. The BN may simply be considered as a compact representation of the conditional independence relations in observational data, and in this non-casual interpretation, it may be used to infer conditional and marginal distributions in the observational data to provide predictive analysis. However, if we interpret the BN to be a causal BN, then the BN is a unique DAG that enables us to reason about intervention and understand the system being modelled at a deeper level. Where we have a directed edge $A \longrightarrow B$ in a graph, we say that $A$ is a *parent* of $B$, or equivalently, $B$ is a *child* of $A$.

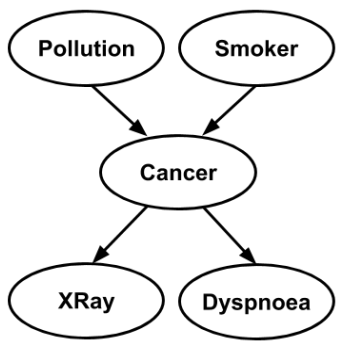

**Figure 1** – Hypothetical DAG on Cancer

Figure 1 shows a DAG representing a simple model of two causes of (lung) cancer and two effects of cancer. It encapsulates the relationships between the variables, in particular the *conditional dependence and independence relations* between the variables. *Conditional probability* tells us, for example, the probability that the person will have a cloudy X-ray given that we know they have Cancer, written as $P(XRay \mid Cancer)$[1] where "|" means "given". *Conditional independence* tells us which variables are irrelevant to that probability. For example,

$$P(XRay \mid Cancer, Smoker) = P(XRay \mid Cancer)$$

tells us that the probability of a cloudy X-Ray is independent of whether Smoker is true given that we know the person has Cancer. The symbol "⊥" means "is independent of" and so this is written as:

$$XRay \perp Smoker \mid Cancer.$$

Figure 2 shows the three causal classes possible with three variables, together with all the conditional independence or dependence relations between $A$ and $C$ given $B$ that they entail. The DAGs in Figure 2(a) and (b) entail the same conditional independence relationship which means they cannot be distinguished by their conditional independence relations solely from observational data. When this is the case, we say that they belong to the same *Markov Equivalence Class* (MEC, but simply referred to as equivalence class from now on). However, in Figure 2(c), $A$ and $C$ are independent but become ***dependent*** given $B$ (indicated by the symbol "⊥̸ "). This kind of relationship is known as *explaining away* and cannot be represented in undirected probabilistic graphs. A node which has multiple parents is known as a *collider*. The configuration shown in Figure 2(c), where $B$ is a collider, and there is no edge between $A$ and $C$, will be referred to here as a *v-structure*, although other authors use the term *unshielded collider*.

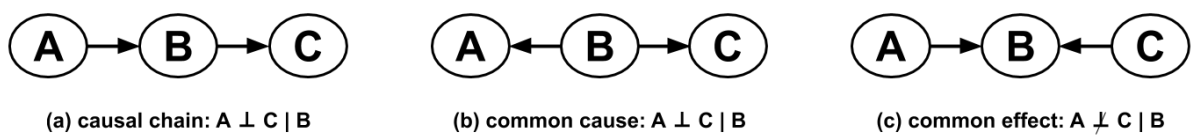

**(a) causal chain: A ⊥ C | B**        **(b) common cause: A ⊥ C | B**        **(c) common effect: A ⊥̸ C | B**

**Figure 2** – Causal classes containing three variables, and their corresponding conditional independence relationships.

Probabilistic graphical models represent conditional independence through the notion of *graphical separation*. For example, Cancer "separates" Smoker from XRay in the graph in Figure 1. DAGs use a special form of graphical separation known as *d-separation* to represent conditional independence relationships. D-separation is defined as follows (Pearl, 1988): If $X$ and $Y$ are nodes in DAG $G$, a subset of the remaining nodes, $S$, *d-separates* $X$ from $Y$ if $S$ *blocks all* paths between $X$ and $Y$. Each path between $X$ and $Y$ is blocked by $S$ if ***at least one*** of the nodes between $X$ and $Y$ on that path meets one of the following conditions, either:

---

[1] We use $P(XRay \mid Cancer)$ as a shorthand for the conditional probability distribution over all values of $XRay$ and $Cancer$, that is, $P(XRay = cloudy \mid Cancer = true)$, $P(XRay = clear \mid Cancer = true)$ etc.

- it is a collider and neither it, nor any of its descendants, are in **S**;
- or, it is not a collider and it is **in S**

If **S** does d-separate $X$ and $Y$, we say that **S** is a *Sepset* (also referred to as *cut-set* or *separating set*) for $X$ and $Y$. Figure 3 presents four examples of applying the d-separation rules to examine whether $X$ and $Y$ are d-separated, where the different conditioning sets are indicated by shaded nodes. Paths which are not blocked are known as *active paths* and are shown by green arrows, and conditioning sets which are minimal Sepsets are shaded in pink; otherwise, they are shaded in grey.

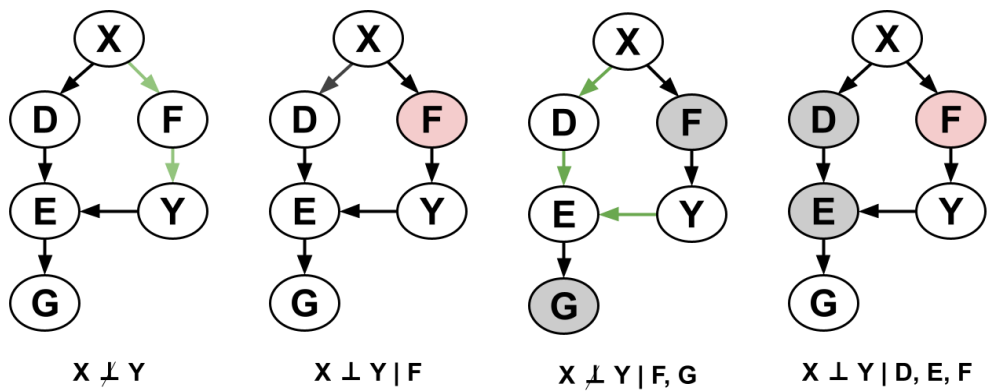

**Figure 3** - Examples illustrating the application of d-separation.

The BN represents the set of conditional independence relationships (and implicitly therefore, the dependence relationships too) in the joint probability distribution over all the variables, $P(\boldsymbol{X})$. Two assumptions about the DAG in a BN are made:

- **Markov Condition** – every variable $X$ in $G$ is conditionally independent of its non-descendants given its parents. This is equivalent to saying that every conditional independence implied by d-separation in the DAG is present in the joint probability distribution $P(\boldsymbol{X})$. Importantly, this condition means that the joint probability distribution $P(\boldsymbol{X})$ can be decomposed as follows (where $\boldsymbol{Pa}(X_i)$ are the parents of $X_i$):

$$P(\boldsymbol{X}) = \prod_{i=1}^{n} P(X_i \mid \mathbf{Pa}\,(X_i))$$

- **Minimality Condition** – we cannot remove any of the edges in the DAG without the graph implying a conditional independence relationship that is not present in $P(\boldsymbol{X})$.

Pearl (1988) expresses these two conditions by saying that $G$ is a *minimal Independence-Map (I-map)* of $P(\boldsymbol{X})$. If the DAG represents the causal structure of the variables, the Markov Condition is referred to as the *Causal Markov Condition* since it links the probabilistic and causal interpretations of the DAG.

We wish to learn the BN from a set of data $\boldsymbol{D}$, which consists of $N$ data instances $\boldsymbol{D} = \{\boldsymbol{d_1}, \ldots, \boldsymbol{d_N}\}$, each of which defines the value of each of the variables $X_1 .. X_n$, that is $\boldsymbol{d} = \{x_1, \ldots, x_n\}$ (lower case letters denote values or states of a variable). *Discrete BNs* allow variables which take discrete values each having a defined probability of occurring dependent upon the value of the parents. For example, in Figure 1, the probability of Dyspnoea occurring might be 0.9 if Cancer were true, but only 0.05 if Cancer were false. This set of conditional probabilities for a discrete variable is known as the *Conditional Probability Table (CPT)*.

*Linear Gaussian BNs* are based on continuous variables which are assumed to follow Gaussian distributions. Each value of a child variable is a linear combination of its parents' values plus a local noise component. These networks are parameterised with Conditional Probability Distributions (CPDs), as opposed to CPTs. Unless stated otherwise, we will assume linear relationships when we use the term Gaussian BN. *Hybrid BNs* support both discrete and continuous distributions. The most common form of hybrid BN is a *Conditional Linear Gaussian BN (CLGBN)* which allows discrete variables to be parents of a continuous variable, with a separate Gaussian Linear Model with different weighting coefficients for each set of discrete parent values (Geiger and Heckerman, 1994). While CLGBNs do not generally allow continuous variables to be parents of discrete ones, works such as those by Andrews et al. (2018) describe hybrid BNs which remove this restriction.

Constructing a BN involves two main phases: a) determining the graphical structure and b) determining the parameters **Θ**. The graph and the parameters of a BN model can be determined by expert knowledge, learnt from data, or a combination of both. This paper focuses on the problem of learning the structure of BNs from data, or from both data and expert knowledge.

Learning the structure of a BN represents an NP-hard problem partly because the solution space of graphs grows super-exponentially with the number of variables. Robinson (1973) showed that the recurrence relation:

$$|G_n| = \sum_{i=1}^{n} (-1)^{i-1} \binom{n}{i} 2^{i(n-i)} |G_{n-i}|$$

computes the number of possible DAGs for $n$ variables, $|G_n|$, with $|G_0|$ defined as 1. Using this recurrence formula, Table 1 illustrates how the number of possible DAGs grows super-exponentially as $n$ increases. Clearly, a naïve exhaustive search is not a solution for any problem with a reasonable number of variables.

**Table 1** - Number of directed graphs, directed acyclic graphs, and the percentage of directed graphs which are acyclic for different number of variables.

| Number of variables, $n$ | Number of directed graphs ($3^{n(n-1)/2}$) | Number of DAGs, $|G_n|$ | Percentage of directed graphs which are acyclic[2] |
|---|---|---|---|
| 2 | 3 | 3 | 100.0% |
| 3 | 27 | 25 | 92.59% |
| 4 | 729 | 543 | 74.49% |
| 5 | 59,049 | 29,281 | 49.59% |
| 6 | $1.4349 \times 10^7$ | $3.7815 \times 10^6$ | 26.35% |
| 7 | $1.0460 \times 10^{10}$ | $1.1388 \times 10^9$ | 10.89% |
| 8 | $2.2877 \times 10^{13}$ | $7.8730 \times 10^{11}$ | 3.42% |

In general, structure learning algorithms fall into two main classes. The first class is constraint-based methods that eliminate and orientate edges based on a series of conditional independence (CI) tests. The second class, score-based methods, represent a traditional machine learning approach where the aim is to search over different graphs maximising an objective function. The graph that maximises the objective function is returned as the preferred graph. Additionally, hybrid algorithms that combine score-based and constraint-based approaches are often viewed as a third class of structure learning. Chickering et. al. (1994)

---

[2] This calculation assumes that the directed graph has at most one arc between each pair of nodes.

demonstrate that score-based learning is NP-hard, and Chickering et al. (2004) show that constraint-based learning is as well. This is true even under favourable conditions such as limiting the number of parents to 3 and having a constant time method of computing scores from the data.

## 3. CONSTRAINT-BASED LEARNING

Constraint-based learning uses CI tests on the data to determine the conditional independence relationships between the variables under investigation, and hence construct a graph consistent with the data. Constraint-based learning is often assumed to discover causal relationships under the assumptions of *causal faithfulness* and *causal sufficiency* which we cover later in this section. As the simple examples in Figure 2 have already shown, a set of independence relationships may be consistent with multiple DAGs. Hence, rather than producing a single DAG, constraint-based algorithms return the set of DAGs consistent with the independence relationships in the data. That is, the *equivalence class* referred to in Section 1.

Verma and Pearl (1990) show that two DAGs belong to the same equivalence class if they have the same adjacencies (same *skeleton*) and the same set of *v-structures*. The adjacencies and v-structures are represented by a *Partially Directed Acyclic Graph (PDAG)* which has a mixture of directed and undirected edges, with the directed edges indicating the v-structures. Figure 4(a) shows three DAGs which entail the same set of independence relationships even though the arrow orientations vary between $A, B$ and $C$. Figure 4(b) shows the corresponding PDAG, with directed edges indicating the v-structure $B \longrightarrow D \longleftarrow C$. Implicit in that PDAG is that $B - D - E$ and $C - D - E$ are *not* v-structures, so we can deduce that there must be a directed edge $D \longrightarrow E$, and filling in all the additional implicit directed arrows such as this one creates a Complete PDAG (CPDAG), as shown in Figure 4(c)[3]. The CPDAG represents the equivalence class. A directed edge in the CPDAG means that all the equivalent DAGs must have that same directed edge, but undirected edges in the CPDAG indicate that the equivalent DAGs can have a directed edge in either direction.

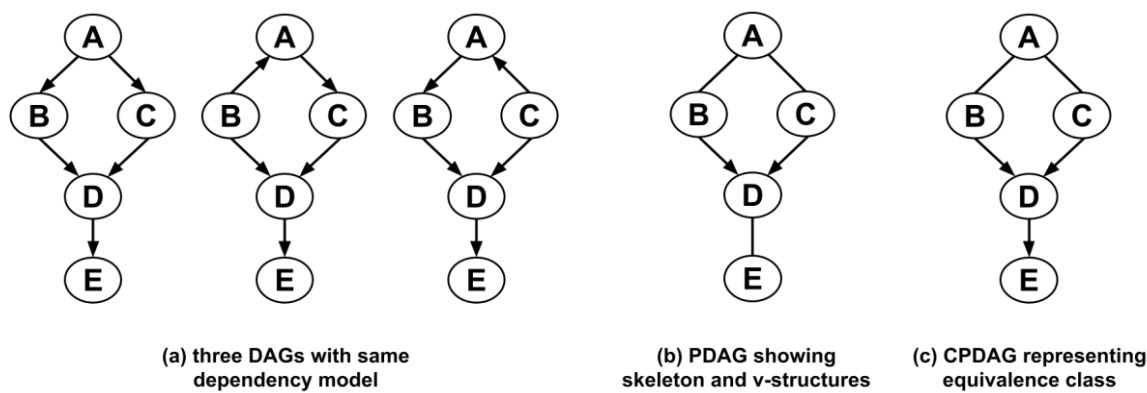

(a) three DAGs with same
dependency model

(b) PDAG showing
skeleton and v-structures

(c) CPDAG representing
equivalence class

**Figure 4** - Illustration of the equivalence classes, PDAGs and CPDAGs, based on an example in Verma and Pearl (1990).

---

[3] PDAGs are also referred to as *rudimentary patterns* (Verma and Pearl, 1990), and CPDAGs are referred to as *completed patterns* (Verma and Pearl, 1990)*,* sometimes simply *patterns* (Spirtes and Glymour, 1991), or even *essential graphs* (Andersson et al., 1997) or *maximally orientated graphs* (Meek, 1995).

We noted in the Preliminaries that two assumptions are made when formally defining a Bayesian Network: the Markov and Minimality Conditions. To recap, this means that all conditional independence relationships implied from the DAG by d-separation are present in the probability distribution. In general, however the converse is not necessarily true, in that there may be conditional independence relationships present in the probability distribution that are ***not*** reflected by the DAG. If this is the case, we say that the DAG and the probability distribution are *unfaithful* to one another.

Figure 5 shows an example in which the network is unfaithful. Applying d-separation rules to the DAG would indicate that $A$ and $C$ are ***not independent***. However, the ***particular values*** chosen for the CPTs shown give rise to a probability distribution where $A$ and $C$ are ***independent***. Thus, there is an independence in the probability distribution which is not reflected by the DAG, and so it is unfaithful. In other words, this example shows that it is possible to have causation without association. Note that constraint-based algorithms do often make the additional assumption that all independence relationships present in the distribution ***are*** reflected in the DAG. In this case, we say that the DAG and distribution are *faithful* to each other, or that the DAG is a *perfect map (P-map)* of the distribution.

The next subsection of this section describes the CI tests used to determine the set of independence relationships, and the remaining three subsections each discuss a group of constraint-based algorithms. Subsection 3.2 describes the prototypical constraint-based algorithms that learn the graph structure globally and make the assumption of causal sufficiency which is also explained there. Subsection 3.3 describes local learning algorithms which learn the graph structure local to each variable which can then be merged to produce the overall graph. Finally, subsection 3.4 describes algorithms which assume the existence of latent variables (i.e., causal insufficiency) and which are represented by a new kind of graph covered in that subsection. The main constraint-based algorithms covered in these subsections are shown in red hues in Figure 6, which also illustrates the evolution of structure learning algorithms covered in this review. Lastly, Table 2 summarises the constraint-based algorithms covered in terms of whether they are global or local, the type of output[4] they produce, and the key assumptions the algorithms make. Note that faithfulness assumptions that are stronger than the normal are marked in red text, and those that are weaker marked in blue text.

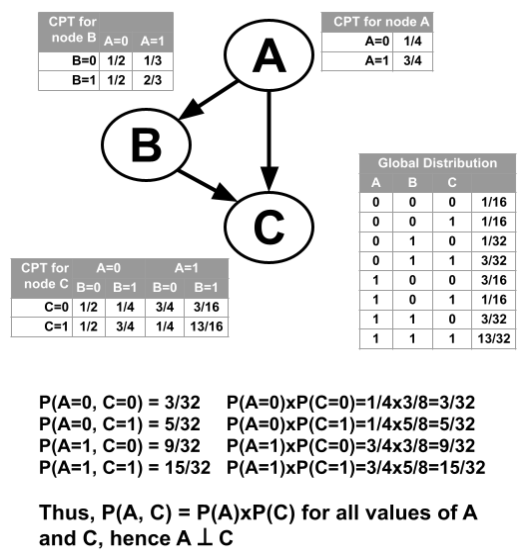

**Figure 5** - Example unfaithful network.

---

[4] Table 2 reports the type of output produced by the algorithm *in the original paper* cited as the reference in the table. For most of the local learning algorithms, this output was a set of local structures such as Markov Blankets, rather than a whole integrated graph. However, when these algorithms have subsequently been incorporated into software packages, for example, Inter-IAMB in the bnlearn software package (Scutari, 2021) these local structures may be merged, and the output would then be a whole graph, typically a CPDAG.

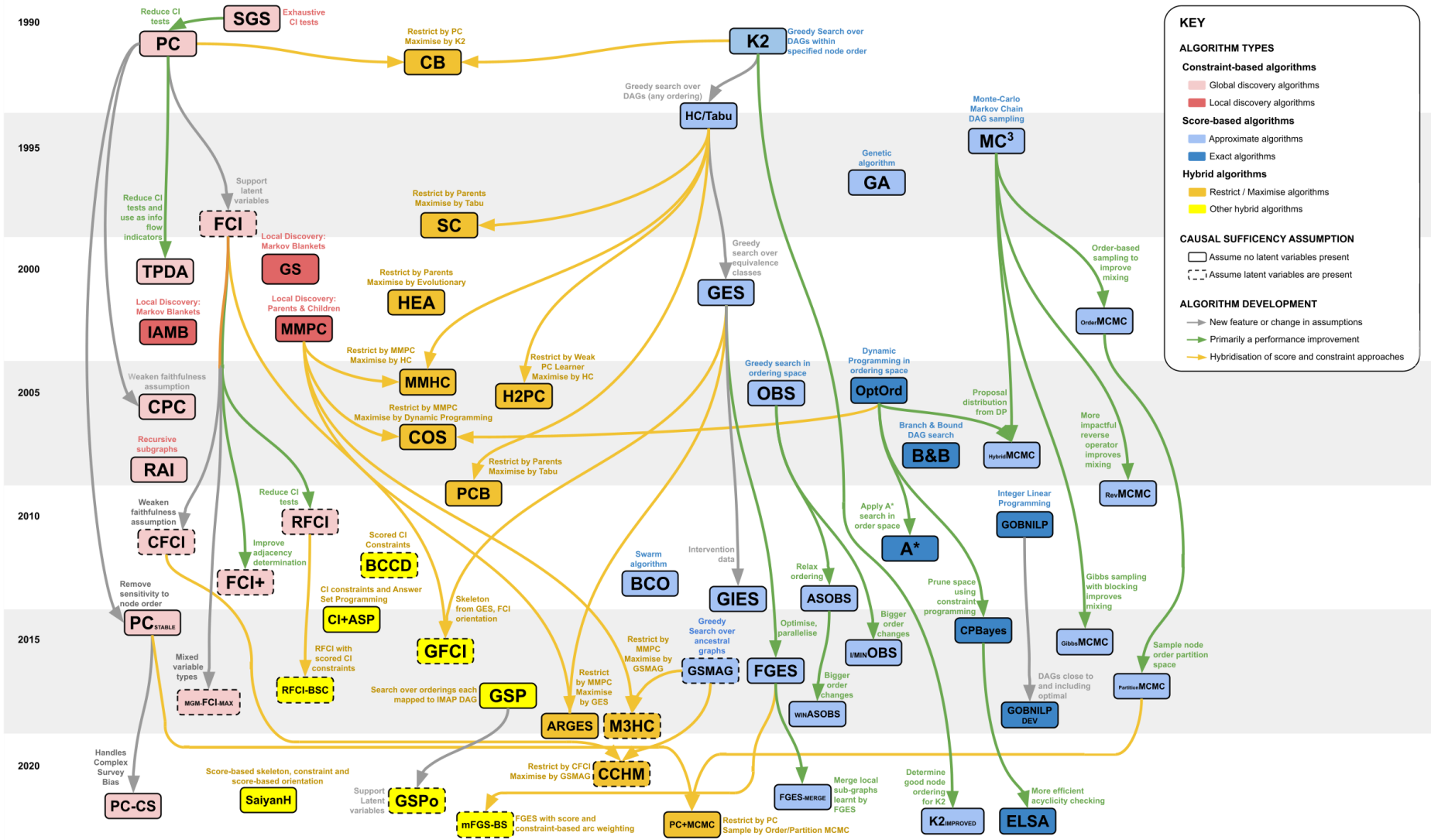

**Figure 6** - The evolution of BN structure learning algorithms across all classes of learning.

**Table 2** - Characteristics of the constraint-based algorithms reviewed, ordered chronologically. Red text represents stronger faithfulness assumptions, and blue text, weaker than usual faithfulness assumptions

| Abbreviation | Algorithm Name / Description | Reference | Global or Local | Type of output | Faithfulness assumptions | Causal sufficiency assumed |
|---|---|---|---|---|---|---|
| SGS | Spirtes-Glymour-Scheines | Spirtes et al., 1990 | Global | CPDAG | complete | yes |
| PC | Peter and Clark | Spirtes and Glymour, 1991 | Global | CPDAG | complete | yes |
| FCI | Fast Causal Inference | Sprites et al., 1999 | Global | PAG | complete | no |
| GS | Grow-Shrink | Margaritis and Thrun, 2000 | Local | CPDAG | complete | yes |
| TPDA | Three Phase Dependency Algorithm | Cheng et al., 2002 | Global | CPDAG | monotonic | yes |
| HITON | Hiton (Greek for blanket or cover) | Aliferis et al., 2003 | Local | Parents and children | complete | yes |
| IAMB | Incremental Association Markov Blanket | Tsamardinos et al., 2003 | Local | Markov Blankets | complete | yes |
| Inter-IAMB | Interleaved-IAMB | Tsamardinos et al., 2003 | Local | Markov Blankets | complete | yes |
| MMPC | Max-Min Parents Children | Tsamardinos et al., 2003 | Local | Parents and children | complete | yes |
| MMMB | Max-Min Markov Blanket | Tsamardinos et al., 2003 | Local | Markov Blankets | complete | yes |
| Fast-IAMB | Fast IAMB | Yaramakala and Margaritis, 2005 | Local | Markov Blankets | complete | yes |
| CPC | Conservative PC | Ramsey et al., 2006 | Global | CPDAG with unfaithful triples marked | adjacency | yes |
| RAI | Recursive Autonomy Identification | Yehezkel and Lerner, 2009 | Global | CPDAG | complete | yes |
| SI-HITON-PC | Semi-interleaved Hiton Parents&Children | Aliferis et al., 2010 | Local | Parents and children | complete | yes |
| RFCI | Really Fast Causal Inference | Colombo et al., 2012 | Global | PAG | complete | no |
| CFCI | Conservative Fast Causal Inference | Colombo et al., 2012 | Global | PAG | adjacency | no |
| FCI+ | Fast Causal Inference+ | Claassen et al., 2013 | Global | PAG | complete | no |
| PC-Stable | Peter and Clark - Stable | Colombo and Maathuis, 2014 | Global | CPDAG with conflicting v-structures marked | complete | yes |
| FCI-Stable | Fast Causal Inference Stable | Colombo and Maathuis, 2014 | Global | PAG | complete | no |
| VCSGS | Very Conservative SGS | Spirtes and Zhang, 2014 | Global | Not implemented | triangle | yes |
| PC-Max | Peter and Clark - MAX | Ramsey, 2016 | Global | CPDAG | complete | yes |
| MGM-FCI-MAX | Mixed Graphical Model - FCI - MAX | Raghu et al., 2018 | Global | PAG | complete | no |
| PC-CS | PC – Complex Surveys | Marella and Vicard, 2022 | Global | CPDAG | complete | no |

### 3.1. Determining Conditional Independence

CI tests check whether nodes $A$ and $B$ are conditionally independent given a *conditioning set* of nodes $\boldsymbol{S} = \{S_1, \ldots, S_q\}$ where $q$ ranges over the number of nodes in the conditioning set. In other words, it decides whether $\boldsymbol{S}$ will be a Sepset for $A$ and $B$ in the learnt graph. Although we generically describe these as tests of conditional independence, the same tests are also used when $\boldsymbol{S} = \emptyset$; that is, testing whether $A$ and $B$ are unconditionally independent. These tests rely on setting an arbitrary threshold used to determine conditional independence, and can only identify conditional independence relationships present in the dataset which may not necessarily reflect those present in the true distribution. Therefore, we must recognise that these CI tests can make "mistakes", and that these errors are more likely to occur with smaller sample sizes. Minimising the effects of these errors is an important consideration when designing constraint-based algorithms because if an edge is mistakenly removed from the graph at an early stage in the discovery process, this is likely to cause the discovery of incorrect edges at a later stage.

The most commonly used CI tests for discrete BNs are the $G^2$ and $\chi^2$ statistical tests and mutual information (MI), whereas for Gaussian BNs, Fisher's z-test is frequently used. CI tests such as $G^2$ and $\chi^2$ assume a null hypothesis that $A$ and $B$ **are** conditionally independent given $\boldsymbol{S}$. The tests produce a test statistic which can then be used to estimate how likely, defined by a *p-value,* that the observed data is, given the null hypothesis. If the p-value is below a predefined *significance level,* $\alpha$, typically chosen as 0.05, the null hypothesis is rejected and it is assumed $A$ and $B$ are conditionally **dependent** given $\boldsymbol{S}$. Conversely, if the p-value is above $\alpha$, the null hypothesis cannot be rejected and we assume $A$ and $B$ are conditionally independent given $\boldsymbol{S}$. The general form of the $G^2$ test statistic is:

$$G^2 = 2 \cdot \sum Observed \cdot \ln\left[\frac{Observed}{Expected}\right]$$

which when applied to test the conditional independence of discrete variables $A$ and $B$ given conditioning set $\boldsymbol{S} = \{S_1, \ldots, S_q\}$ becomes (Spirtes et al., 2000):

$$Observed = N_{abs} \text{ , and } Expected = \frac{N_{bs} \cdot N_{as}}{N_s}, \text{ so } G^2 = 2\sum_{a,b,s} N_{abs} \ln\left[\frac{N_{abs} N_s}{N_{bs} N_{as}}\right]$$

where $a, b$ range over the values of $A, B$ respectively, and $s$ ranges over all the combinations of values of the conditioning set $\boldsymbol{S}$. $N_{abs}$ is the number of data cases with specific values $A = a, B = b, \boldsymbol{S} = \{s_1, \ldots, s_q\}$. $N_{bs}$ is the marginal count over all values of $a$ for data cases with $B = b, \boldsymbol{S} = \{s_1, \ldots, s_q\}$, with $N_{as}$ and $N_s$ being analogous marginal counts over $b$ and $a, b$ respectively. The degrees of freedom, $df$, which is required to determine the p-value from the test statistic is dependent upon the cardinality of the variables and is given by:

$$df = (|A| - 1)(|B| - 1) \prod_{i=1}^{q} |S_i|$$

where $|A|, |B|, |S_i|$ are the number of distinct values that nodes $A, B, S_i$ can take respectively. $df$ is defined here under the assumption that none of the values of $N_{abs}$ is zero, that is, every possible combination of values is present in the data. However, it is likely that some combinations of values will be absent with limited sample sizes, and so Sprites et al. (2000) suggest a heuristic of reducing $df$ by 1 for every combination of values where $N_{abs}$ is zero. The $\chi^2$ CI test is similar, but with the test statistic defined as:

$$\chi^2 = 2 \cdot \sum \frac{(Observed - Expected)^2}{Expected}$$

Another CI test for discrete BNs is *mutual information* (MI) which measures the amount of information shared between two variables (Cheng et al., 1997). The mutual information between two variables $A$ and $B$ is:

$$MI(A, B) = \sum_{a,b} P(a, b) \cdot ln\left[\frac{P(a, b)}{P(a)P(b)}\right]$$

where $P(a, b)$ is shorthand for $P(A = a, B = b)$, and similarly for $P(a)$, and $P(b)$. C*onditional mutual information* is defined as:

$$MI(A, B \mid \boldsymbol{S}) = \sum_{a,b,s} P(a, b \mid \boldsymbol{s}) \cdot ln\left[\frac{P(a, b \mid \boldsymbol{s})}{P(a \mid \boldsymbol{s}) \cdot P(b \mid \boldsymbol{s})}\right]$$

where $P(a, b \mid \boldsymbol{s})$ is shorthand for $P(A = a, B = b \mid \{S_1 = s_1, ..., S_q = s_q\})$, and similarly for $P(a \mid \boldsymbol{s})$ and $P(b \mid \boldsymbol{s})$. A value of 0 for $MI(A, B \mid \boldsymbol{S})$ indicates that there is no information flow between $A$ and $B$ when conditioned on $\boldsymbol{S}$, that is, they are conditionally independent. In practice a threshold value $\epsilon$ is chosen so that if $MI(A, B \mid \boldsymbol{S}) < \epsilon$, conditional independence is assumed. Variable $\epsilon$ may be given an arbitrary small value such as 0.01 (Cheng at al., 1997) or it may be estimated by comparing the predictive accuracy using different values (Cheng and Greiner, 1999). If we rewrite the frequencies used in the definition of the $G^2$ test statistic as probabilities, we see that it only differs from mutual information by a scaling factor:

$$G^2 = 2 \cdot \sum_{a,b,s} N_{abs} \cdot ln\left[\frac{N_{abs} \cdot N_s}{N_{as} \cdot N_{bs}}\right]$$

$$= 2 \cdot \sum_{a,b,s} N \cdot P(a, b, \boldsymbol{s}) \cdot ln\left[\frac{N_{abs}/N_s}{N_{as}/N_s \cdot N_{bs}/N_s}\right]$$

$$= 2 \cdot N \cdot \sum_{a,b,s} P(a, b, \boldsymbol{s}) \cdot ln\left[\frac{P(a, b \mid \boldsymbol{s})}{P(a \mid \boldsymbol{s}) \cdot P(b \mid \boldsymbol{s})}\right] = 2 \cdot N \cdot MI(A, B \mid \boldsymbol{S})$$

In the case of Gaussian BNs, Fisher's Z-test is commonly used to test the null hypothesis that the partial correlation coefficient is zero. Fisher's Z-test uses Fisher's Z-transformation which is defined as:

$$\hat{Z} = \frac{1}{2} \ln \frac{1 + \hat{\rho}_{ab|s}}{1 - \hat{\rho}_{ab|s}},$$

where $\hat{\rho}_{ab|s}$ is the partial correlation coefficient between values $a$ of node $A$ and values $b$ of node $B$, given values $\boldsymbol{s}$ of the conditioning set S. The value $\hat{\rho}_{ab|s}$ can be computed recursively with conditioning sets of increasing size (Anderson, 1962; de la Fuente, 2004). This transformed version of the partial correlation $\hat{Z}$, follows a normal distribution with:

a mean of $\frac{1}{2}\ln\frac{1+\hat{\rho}_{ab|s}}{1-\hat{\rho}_{ab|s}}$ and standard deviation $\sigma$, of $\frac{1}{\sqrt{N-q-3}}$,

where $q$ is the number of variables in the conditioning set. We can, therefore, use a normal distribution Z-score to compute the p-value of obtaining the computed partial correlation coefficient given the null hypothesis of zero partial correlation ($Z_0 = 0$):

$$Z = \frac{\hat{Z}_{tran} - Z_0}{\sigma} = \frac{1}{2}\sqrt{N-q-3}\ln\frac{1+\hat{\rho}_{ab|s}}{1-\hat{\rho}_{ab|s}}$$

The test statistics and associated p-values described in this section are usually used in a binary decision to decide whether variables are conditionally independent or not. However, some algorithms also use them as a measure of the ***degree*** of association, or dependence or independence between variables. For example, a high mutual information value indicates that two variables are strongly associated with each other.

### *3.2. Global Discovery Algorithms*

This group of constraint-based algorithms are known as global discovery algorithms since they attempt to learn the graph structure as a whole rather than first learning the local structure relating to each variable separately as the local constraint-based algorithms do (see subsection 3.3). Both these global and the local constraint-based algorithms make one further assumption known as *causal sufficiency*, which is of importance if we wish to interpret the BNs causally. This assumption means there are no *latent* (unmeasured) variables that would affect the causal relationships. For example, variables that are a common cause of two or more of the measured variables $\boldsymbol{X}$, which are widely known as *latent confounders*.

#### *3.2.1. SGS Algorithm*

The SGS algorithm (Spirtes et. al., 1990) is rather inefficient but is of interest since many constraint-based algorithms build upon its approach. SGS relies on two key theorems derived from the definition of Bayesian Networks (Verma and Pearl, 1990) that apply to faithful and causally sufficient BNs:

1.  if $A \not\perp B \mid \boldsymbol{S}$ for ***every*** subset $\boldsymbol{S} \subseteq \boldsymbol{X} \setminus \{A, B\}$ then $A$ and $B$ are adjacent in the graph ("$\boldsymbol{X} \setminus \{A, B\}$" means set $\boldsymbol{X}$ with elements $A$ and $B$ removed);
2.  if $A$ and $B$, and $B$ and $C$ are adjacent in the graph, but $A$ and $C$ are not adjacent, and if $A \not\perp C \mid \boldsymbol{S} \cup B$ for ***any*** subset $\boldsymbol{S} \subseteq \boldsymbol{X} \setminus \{A, B, C\}$ in the DAG, then $A, B, C$ form a v-structure $A \longrightarrow B \longleftarrow C$

SGS starts from a complete (i.e., there is an edge between every pair of nodes) undirected graph on the node set $\boldsymbol{X}$ and learns the Markov equivalence class in three phases:

1.  *Adjacency phase*: making use of rule 1 above, for each pair of nodes $A, B$ this phase performs a CI test on $A$ and $B$ conditional on every possible subset $\boldsymbol{S}$ of the remaining nodes. If conditional independence occurs for any set $\boldsymbol{S}$, the edge between $A$ and $B$ is removed. This phase produces the graph skeleton.

2. *V-structure phase*: using rule 2 above, for every triple $A, B, C$ in the skeleton where $A, B$ and $B, C$ are adjacent pairs, and $A$ and $C$ are not adjacent, perform CI tests on $A$ and $C$ conditional on every possible subset $\boldsymbol{S}$, of the remaining nodes where $\boldsymbol{S}$ contains $B$. If $A$ and $C$ are conditionally dependent given for every subset $\boldsymbol{S}$, then mark $A - B - C$ as a v-structure $A \longrightarrow B \longleftarrow C$. This phase produces the PDAG.

3. *Orientation propagation phase*: for every undirected edge in the PDAG, check if one of the orientations would:
   a. introduce a cycle into the graph, or
   b. create a new v-structure.

   If so, then that orientation is forbidden and so the opposite orientation can be assumed. These rules are applied repeatedly until no more edges can be orientated, producing the CPDAG.

The first phase in the SGS algorithm is particularly expensive. In the worst case, it requires $n(n-1) \cdot 2^{n-3}$ CI tests, which makes it exponential in $n$ and therefore infeasible for a reasonable number of variables. However, SGS is relatively stable (Spirtes et al., 2000), in that errors made in CI tests tend not to be highly amplified by subsequent steps. A CI mistake in phase 1 may result in an extraneous or missing edge, but this would not affect other decisions made in that phase. However, these adjacency errors and further errors in identifying v-structures may propagate out to cause further orientation errors.

### 3.2.2. The PC algorithm

The adjacency phase in SGS exhaustively tests every possible conditioning set for each pair of nodes. This is computationally expensive and also means that many high order CI tests (CI tests applied to large parent-sets) are performed which are unreliable because the individual elements of the CI test are based on relatively few data instances. To counter these issues, the PC algorithm by Spirtes and Glymour (1991) performs the adjacency phase with conditioning sets of increasing size – checking all pairs of nodes $A, B$ at a particular conditioning set size and removing edges $A - B$ if a Sepset is found *before* moving to higher conditioning sets. Moreover, the PC adjacency phase makes use of the result that the minimum conditioning set that d-separates two nodes must be a subset of the union of the parents of those nodes under the assumptions of faithfulness and causal sufficiency (Verma and Pearl, 1990). Thus, the algorithm need only consider conditioning sets of nodes which are adjacent to $A$ and $B$. This condition has no benefit initially since PC starts from a complete graph, but it reduces the number and order of the CI tests that are performed as the adjacency phase progresses and edges are removed.

To improve computational efficiency, the v-structure phase makes use of the Sepsets identified in the adjacency phase; if the Sepset for $A$ and $C$ identified in the adjacency phase for an unshielded triple $A - B - C$ does not contain $B$, then this identifies it as the v-structure $A \longrightarrow B \longleftarrow C$. The PC algorithm then performs orientation propagation using the "Meek Rules" (Meek, 1995). The complexity of the PC algorithm is bounded by (Spirtes et al., 2000):

$$\frac{n^2 (n-1)^{s_{max}-1}}{(s_{max}-1)!}$$

where $s_{max}$ is the maximum size of any Sepset. This complexity bound is hard to quantify, but PC is polynomial given a limit on node degree (Claassen et. al., 2013). Although far more efficient, the PC algorithm is less stable than SGS. For example, edges mistakenly removed in the adjacency phase can result in other edges being mistakenly retained later on in the adjacency phase.

### 3.2.3. The Conservative PC (CPC) algorithm

The PC and SGS algorithms assume complete faithfulness, and one direction in which constraint-based algorithms have developed is to weaken this assumption. The Conservative PC (CPC) algorithm (Ramsey et al., 2006) does this by considering how faithfulness is assumed in the adjacency and orientation phases of the PC algorithm separately, using the terms *adjacency-faithfulness* and *orientation-faithfulness* respectively. It is shown that if only adjacency-faithfulness is assumed, the v-structure phase can detect and mark unfaithful v-structures. To do this, CPC considers all Sepsets of $A$ and $C$ to determine if $A - B - C$ is a v-structure – marking it as such only if none of the Sepsets contain $B$. Moreover, unless the "vote" is unanimous, the triple is marked as unfaithful. Thus, CPC is more cautious about orientating edges than PC, hence the name "conservative". Simulations on a dataset of sample size 1,000 showed CPC to be only slightly slower than PC, but generating fewer erroneous edge orientations.

### 3.2.4. The Very Conservative SGS (VCSGS) algorithm

Zhang and Spirtes (2008) showed that a restricted assumption of faithfulness could be applied to the adjacency phase too. This weakened faithfulness condition is a combination of the minimality condition described in the Introduction, and *triangle faithfulness* which only assumes faithfulness on fully connected triples. With this weakened faithfulness assumption alone, it is possible to identify all other faithfulness violations. Spirtes et al. (2014) describe a version of SGS, the Very Conservative SGS, which would implement this weaker faithfulness assumption, though it was left as an open question whether it could be made efficient enough to be viable. It does not seem as though this algorithm has been implemented.

### 3.2.5. The PC-Stable Algorithm

Colombo and Maathuis (2014) considered the effect of mistaken CI test decisions arising from limited sample sizes and, in particular, their interaction with the order in which the CI tests are performed. They showed that the output from all three phases of the original PC algorithm (including related algorithms such as FCI and RFCI which we discuss below) is sensitive to the order in which CI tests are performed. The order in which CI tests are performed is generally an artefact of the way the algorithm is implemented; e.g., it may be related to the lexicographic ordering of the node names, or in the order the variables appear in the data. They proposed modifications to each phase of the original PC algorithm (subsection 3.2.2) to remove this order dependence. Figure 8 presents the pseudo-code for the *PC-Stable* algorithm which has the following three phases:

- Adjacency: in the original PC algorithm, mistaken deletions of edges propagate by erroneously reducing the conditioning sets available in subsequent CI tests at a given conditioning set size. This was remedied by only recomputing adjacencies before processing all the CI tests at each conditioning set size, in contrast to the original PC algorithm where edges are removed and adjacencies adjusted as soon as an independence relationship is detected. This is accomplished by taking a copy of the adjacencies at lines 5 and 6 of the pseudo-code and using this stable copy to determine possible conditioning sets ignoring the fact that edges might have been deleted in lines 8 to 10.
- V-structure: the original PC algorithm re-uses the Sepset used to determine that the triple is unshielded, to also decide whether that triple is a v-structure. Given that the original algorithm can use invalid Sepsets in the adjacency phase, this also means its sensitivity to node ordering can adversely affect v-structure orientation. PC-Stable follows the approach adopted by CPC (subsection 3.2.3) by considering all the Sepsets of $A$ and $C$ in triple $A - B - C$ to decide where it is a v-structure. However, PC-Stable takes a less conservative approach than CPC, which they term majority rule, whereby the triple is marked as a v-structure if a majority of the Sepsets do not contain the middle node $B$. Orientation conflicts are identified during this phase and marked by bi-directional edges, as shown in line 13 to 19 in the pseudo-code.

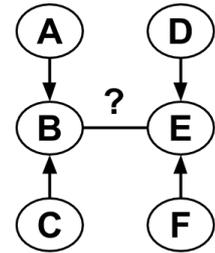

**Figure 7** - Orientation conflict.

- Orientation propagation: mistaken CI tests mean that situations like that shown in Figure 7 can occur; i.e., the two v-structures imply conflicting orientations for edge $B - E$. The original PC algorithm would arbitrarily choose one orientation based on node processing order. PC-Stable instead marks this conflicted edge with a bidirectional arrow.

```
algorithm PC-STABLE is
    input: dataset D
    output: mixed-graph G

    // order-independent adjacency phase

1   G := complete undirected graph   // skeleton, initially complete
2   sepset_size := -1                 // size of CI conditioning sets
3   repeat
4       sepset_size := sepset_size + 1

5       for each node in G do
6           adj[node] := neighbours(node, G)

7       for each edge (X, Y) in G where |adj[X]| > sepset_size do

8           for each possible sepset of X, Y of size sepset_size
9               if X and Y conditionally independent given sepset
10                  delete edge from G

11  until all nodes have less than sepset_size number of neighbours

    // order-independent v-structure orientation phase (majority rule)

12  for each unshielded_triple (X, Z, Y) in G do
13      if majority of sepsets of X and Y do not contain Z
14          if P → X ← Z in G for any P
15              orientate unshielded_triple as X ↔ Z → Y in G
16          else if Z → Y ← Q in G for any Q
17              orientate unshielded_triple as X → Z ↔ Y in G
18          else
19              orientate unshielded_triple as X → Z ← Y in G

    // orientation propagation phase (without conflict identification)

20  repeat
21      for each triple X → Z – Y in G and X, Y not adjacent
22          orientate as X → Z → Y
23      for each pair X – Y in G with a parallel chain X → P → Y
24          orientate as X → Y
25      for each pair X – Y in G with parallel paths X – P → Y and
                X – Q → Y and P, Q not adjacent
26          orientate as X → Y
27  until no more orientations possible
28  return G
```

**Figure 8** - Pseudo-code for PC-Stable algorithm. Program code keywords are coloured blue, comments in grey, key variables in red, and application-specific complex operations or conditions in black. Note, that for clarity, this variant does not identify orientation conflicts in the orientation propagation phase.

The authors compared PC-Stable to PC in a low-dimensional simulation with 50 variables, an average neighbourhood size of 2 or 4 and 1000 rows, and in a high-dimensional simulation with 1000 variables, an average neighbourhood size of 2 and 50 rows. 250 random graphs were generated in each setting. Synthetic Gaussian variable datasets were produced for each graph, and twenty random orderings of variables used with each dataset. The CI test threshold was also varied.

The behaviour of PC and PC-Stable was very similar in the low-dimensional simulation. However, in the high-dimensional one, PC-Stable learnt graphs with lower SHD from the true graph, and with a much smaller variance in SHD across the different dataset orderings. This demonstrated the improved accuracy and stability of PC-Stable over PC. PC-Stable was between three and 13 percent slower than PC due to performing more CI tests. Most recent implementations of algorithms in the PC (and FCI) family employ the order-independence strategies used by PC-Stable.

Marella and Vicard (2022) provide a variant of PC, PC-CS, which addresses selection biases introduced by complex survey designs by using modified independence tests based on resampling techniques. The algorithm was evaluated using synthetic discrete variable datasets generated from random graphs. However, rows with particular values for some variables were preferentially included in the dataset in order to simulate the complex selection biases often found in survey data. The simulation then compared PC-Stable's and PC-CS's ability to learn the random graph. PC-CS produced better SHD scores than PC-Stable, but it should be noted that the simulations had at most 10 variables and so were somewhat limited.

### 3.2.6.  *PC-MAX algorithm*

Whereas PC uses the Sepset identified in the adjacency phase, and CPC and PC-Stable use a voting scheme, to determine whether an unshielded triple is a v-structure, PC-MAX (Ramsey, 2016) uses the Sepset with the highest p-value to determine this. The intuition here is that the Sepset with the highest p-value is the one which most strongly separates the end nodes of the triple, and so should be used to decide whether it is a v-structure or not. Similarly, when two overlapping v-structures would give rise to a bidirectional edge as shown in Figure 7, PC-MAX avoids that conflict by only retaining the v-structure with the highest p-value. PC-MAX also parallelises the adjacency and v-structure phases and adopts the strategies used in PC-Stable to avoid sensitivity to the order of node processing. The authors evaluated performance on Gaussian BNs with PC-MAX obtaining better arc orientation than both PC and PC-Stable on a BN with 1,000 variables. They demonstrated scalability by learning graphs with 20,000 variables and sample size 1,000 on a powerful laptop with four dual-core processors in less than five minutes (Ramsey, 2016), though observed that the score-based Fast Greedy Equivalence Search described in subsection 4.2.2 was faster still and more accurate.

### 3.2.7.  *Three-Phase Dependency Algorithm (TPDA)*

The Three-Phase Dependency Algorithm (TPDA) by Cheng at al. (2002) focuses on reducing the number of statistical tests required, performing at most $O(n^4)$ of them. TPDA adopts an information flow perspective to learn the graph adjacencies and differs from most constraint-based algorithms in that it uses MI tests quantitatively as a measure of information flow along paths, as well as a basis for conditional independence decisions.

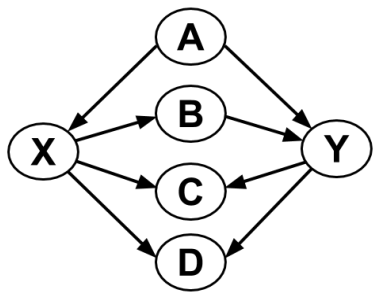

**Figure 9** – A graphical illustration of the process TPDA follows to determine whether to assign an edge between X and Y.

Figure 9, based on an example from Cheng et al. (2002), illustrates a subgraph of a true network to demonstrate the basic principles behind TPDA. In particular, it shows how TPDA determines if a new edge is required between $X$ and $Y$ during its adjacency phase. We consider a point in time in the adjacency phase where TPDA has discovered that $\{A, B, C, D\}$ are the only shared neighbours of $X$ and $Y$. Note at this point, TPDA has not determined the edge orientations. It checks whether $X$ and $Y$ are conditionally independent by testing if $MI(X, Y \mid \boldsymbol{S}) < \epsilon$, where $\epsilon$ represents a threshold negligible information flow. It starts by setting $\boldsymbol{S} = \{A, B, C, D\}$ and progressively removes one node at a time from $\boldsymbol{S}$ so that each time $MI(X, Y \mid \boldsymbol{S})$ is reduced by the greatest amount. It repeats this until either a Sepset is found (in this example, it would find Sepset $\{A, B\}$), or no Sepset is found. The latter situation means that the current graph is not sufficient to explain the information flow between $X$ and $Y$, and hence a direct edge is required between $X$ and $Y$.

In this way, the skeleton of the graph is built up, but with a reduced bound on the number of CI tests. In order for this approach to be sound, a stronger form of faithfulness called *monotone-faithfulness* must be assumed which corresponds to saying that blocking a path between two nodes never increases the information flow between them. In more detail, TPDA builds the skeleton in three phases:

1. *Drafting:* starts with an empty graph and progressively adds undirected edges between pairs of nodes with the highest MI scores, if there is not currently an undirected path between the pair. This creates a maximum spanning tree. That is, where there is one, and only one, path between every pair of variables and the sum of edge scores is a maximum. This tree is used as a good starting point for the next phase.
2. *Thickening*: adds edges between non-adjacent nodes if there is no Sepset in the set of shared neighbours between the two nodes, as described above.
3. *Thinning*: the thickening phase adds edges greedily, and so it can happen that an edge addition can render a previous edge addition superfluous by providing an alternative information flow route. The thinning phase identifies these superfluous edges by looking for direct edges which have parallel indirect routes that can carry the required information flow, and then removes the superfluous direct edge.

TPDA then orientates edges using the v-structure and orientation phases described in the SGS algorithm. Notwithstanding the reduced bound on the number of CI tests required, Cheng at al. (2002) reported similar accuracy and efficiency results to the PC algorithm.

### 3.2.8. *Recursive Autonomy Identification (RAI) algorithm*

Yehezkel and Lerner (2009) also concentrated on reducing the number of CI tests, although they focussed on the costly and unreliable high-order tests. Their *Recursive Autonomy Identification (RAI)* algorithm assumes discrete variables and faithfulness, and starts with a complete undirected graph. Like the PC algorithm, RAI uses CI tests of increasing order. However, edge orientation is undertaken after edge removal at each conditioning set size, and this can allow RAI to identify *autonomous subgraphs.* These can be learnt **independently** of each other through recursive calls of the algorithm.

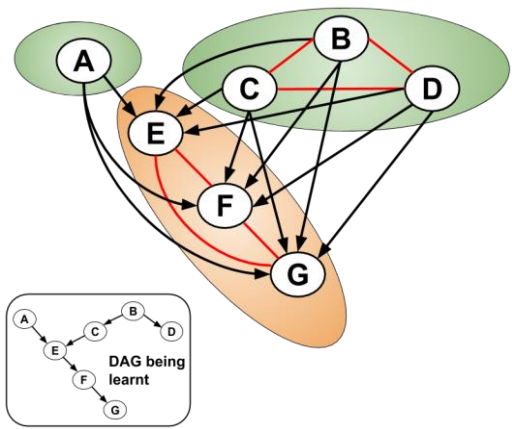

**Figure 10** - Illustration of autonomous subgraphs within the RAI algorithm (based on figure in Yehezkel and Lerner, 2009).

Figure 10, based on the example given in their paper, illustrates these concepts. It shows the state of the graph whilst learning the DAG shown in the inset. In particular, it shows the state after CI tests of order 0 have removed some edges and an orientation step has been performed. At this point, RAI is able to decompose this particular graph into two autonomous ancestor subgraphs marked in green, and a descendant autonomous subgraph marked in orange, which can all then be further refined independently by recursive calls to RAI. The black arrows show the edges which have been orientated after CI tests of order 0, and the red edges are undirected edges within the subgraphs which may be orientated after higher-order CI tests remove more edges. This decomposition allows the overall structure of the graph to appear early on in the learning process, and tends to avoid the higher cost and less reliable high-order CI tests. Whether this decomposition is possible depends upon the independence relationships in the data. If it is not possible, then RAI behaves like the PC algorithm. Nonetheless, the authors reported that RAI demonstrated higher structural and predictive accuracy than contemporaneous algorithms including PC, over a range of commonly evaluated BNs (Yehezkel and Lerner, 2009). They also reported that RAI conducts fewer CI tests and therefore has shorter runtimes.

### 3.3. *Local Discovery Algorithms*

In contrast to the global algorithms described in the previous subsection, the algorithms covered in this subsection learn the local skeleton relating to each variable separately. The local structure learnt can either be the parent and children (i.e., neighbours) of each node, $T$ say, denoted $\boldsymbol{PC}(T)$, or the *Markov Blanket* of $T$, denoted $\boldsymbol{MB}(T)$. The Markov Blanket of node $T$ is defined as the minimal conditioning set for which $T$ is independent of all other nodes besides those in $\boldsymbol{MB}(T)$. Thus, $\boldsymbol{MB}(T)$ shields $T$ from the influence of all other variables. Assuming faithfulness, it can be shown that $\boldsymbol{MB}(T)$ consists of the parents, children, and parents of children (also known as *spouses*) of $T$.

In some contexts, the individual local structure of a particular variable can be useful in its own right. In particular, determining the Markov Blanket of a variable provides a principled causal approach to feature selection, and much of the motivation for, and evaluation of, these local discovery algorithms has been around this use in classification problems (Aliferis et al., 2010). However, within BN structure learning, the local skeletons are learnt for every node and then merged to form the whole skeleton. As we discuss here, this may be done as part of overall constraint-based learning algorithm, with subsequent v-structure and orientation phases producing a CPDAG. Local discovery algorithms may also be part of hybrid approaches which are discussed in section 5.

These local structures should be symmetric. That is, for example, $A \in \boldsymbol{PC}(B) \Leftrightarrow B \in \boldsymbol{PC}(A)$ where $\boldsymbol{PC}(B)$ denotes the parents and children of node $B$. However, errors made by CI tests can mean that local structures may not be symmetric in practice. Algorithms usually resolve these conflicts by applying the "AND-rule", where an edge will only be included in the global skeleton if the two nodes are in each other's parent-and-child sets. More sophisticated symmetry correction approaches can be used however – see, for example, subsection 5.1.5.

### 3.3.1. Markov Blanket algorithms

The Grow-Shrink (GS) algorithm (Margaritis and Thrun, 2000) was the first to exploit the concept of a *Markov Blanket* to reduce the number of CI tests in the adjacency phase. It consists of two steps:

1. *Grow*: for each node $X$ in $\boldsymbol{X}\backslash\{T\}$, GS tests whether $X \perp T \mid \boldsymbol{MB}(T)$. If not, $X$ is immediately added to $\boldsymbol{MB}(T)$ which grows dynamically throughout this step. Nodes are tested for inclusion in $\boldsymbol{MB}(T)$ in decreasing order of the strength of the association between the node $X$ and $T$, which is calculated in a pre-processing step.
2. *Shrink*: the grow step may add unnecessary nodes in the Markov blankets, which this step removes. It checks if $X \perp T \mid \boldsymbol{MB}(T)\backslash\{X\}$ for all $X \in \boldsymbol{MB}(T)$. If yes, $X$ is removed from $\boldsymbol{MB}(T)$.

Having constructed the Markov Blanket of all nodes in $G$, GS performs the following steps:

1. Completes the adjacency determination by removing parents of children of $T$ in each Markov Blanket $\boldsymbol{MB}(T)$. These are identified by having the condition $X \perp T \mid \boldsymbol{S}$ for some $\boldsymbol{S} \subseteq \boldsymbol{MB}(T)\backslash\{X\}$.
2. v-structure and orientation phases similar to SGS and PC.

Margaritis (2003) reported an overall complexity for GS of $O(n^2 + nb^2 2^b)$ CI tests, where $b = max_X(|\boldsymbol{MB}(X)|)$ is the size of the largest Markov Blanket. For dense networks where $b \approx n$, this means the GS algorithm has exponential complexity, although for the more usual sparse networks $b$ can be considered a small constant and in those cases the complexity decreases to $O(n^2)$. Margaritis (2003) reported similar adjacency performance to PC, although GS is said to produce better edge orientation.

The Incremental Association Markov Blanket (IAMB) algorithm optimises Markov Blanket discovery so that it can handle thousands of nodes (Tsamardinos et al., 2003). The authors argue that GS's Markov Blanket grow phase is suboptimal because it is slow to discover spouses in the Markov Blanket $\boldsymbol{MB}(T)$ since these often have weak association with $T$. This in turn leads to more CI tests in the grow and shrink phases. Instead, they propose using conditional mutual information $MI(X, T \mid \boldsymbol{MB}(T))$ to determine the order in which a node $X$ is considered for inclusion into $\boldsymbol{MB}(T)$ during the grow phase. They also propose a variant on IAMB, called *Inter-IAMB*, which interleaves the grow and shrink phases. IAMB and Inter-IAMB were able to handle synthetic networks with up to 1,000 nodes, offering better accuracy in Markov Blanket discovery than GS.

Yaramakala and Margaritis (2005) suggested a further variant, Fast-IAMB. They proposed using the $\chi^2$ test statistic as the metric for deciding which nodes to add to $\boldsymbol{MB}(T)$ during the grow phase. Furthermore, they argued that recomputing the statistic each time a node is added to $\boldsymbol{MB}(T)$ is expensive and so proposed adding groups of nodes to $\boldsymbol{MB}(T)$ before the test statistics are recomputed. They demonstrated similar accuracy in Markov Blanket identification to IAMB and Inter-IAMB, but with savings in execution time of 18-32% over the former, and 28-48% over the latter, together with a reduction in high-order CI tests.

### 3.3.2. Parents-and-children algorithms

The parents and children of node $T$, $PC(T)$ is more directly useful for skeleton learning than $MB(T)$, and can be obtained by removing the spouses from $MB(T)$. However, Max-Min Parents Children (MMPC), HITON-PC and SI-HITON-PC algorithms learn $PC(T)$ directly (Tsamardinos et al., 2003: Aliferis et. al, 2003; and, Aliferis et. al., 2010, respectively). Aliferis et al. (2010) defined a sound generic framework for learning $PC(T)$ into which these three specific algorithms fit, and which consists of:

- a strategy for inclusion of a node $X$ in $PC(T)$, heuristically prioritised, for instance, based on the strength of association between $X$ and $T$;
- an elimination strategy for removal from $PC(T)$, for example, removing $X$ from $PC(T)$ if $X \perp T \mid S$ for some $S \subseteq PC(T) \backslash \{X\}$;
- an approach for interleaving inclusion and elimination. For example, all candidate variables can first be included in $PC(T)$, and then extraneous variables can be eliminated, or variables can be added one at a time to in $PC(T)$, with the elimination step performed each time a new variable is added.

### 3.4. Algorithms assuming the existence of latent variables

The algorithms considered so far have assumed causal sufficiency, which is unreasonable in many real-world situations. We now consider algorithms where this assumption is not made. Explicitly including latent confounders into the DAG might be one approach to avoiding this assumption, but since these confounders are unmeasured and often unknown, this is formidably difficult. It also risks increasing the number of variables so that learning becomes intractable.

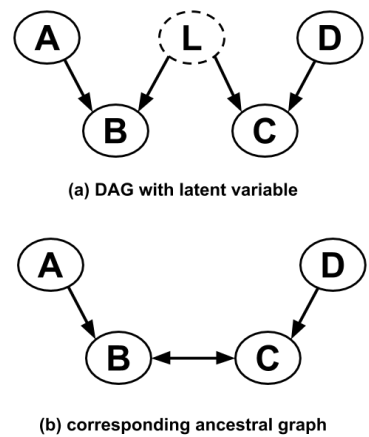

(a) DAG with latent variable

(b) corresponding ancestral graph

**Figure 11** - Latent confounder.

Instead, the most common approach is to learn a graph consisting of only the observed variables, while at the same time taking into account the potential existence of latent variables or confounders that might explain part of the effects or relationships between the observed variables. However, the semantics of DAGs are not detailed enough to represent this information. Figure 11(a) illustrates this issue using a causal graph of four observed variables $\{A, B, C, D\}$, and a latent confounder $L$ which would entail the dependence relationships $A \perp D \mid B$, $A \perp D \mid C$, and $A \not\perp D \mid B, C$. If we attempt to represent this with a DAG of just the four observable variables, then there is no orientation of a directed edge between $B$ and $C$ that could entail these dependence relationships. Fig 10(b) presents an ancestral graph which, unlike DAGs, captures relationships due to latent confounders, and which we describe in the subsection that follows below.

### 3.4.1. Ancestral Graphs

Richardson and Spirtes (2002) introduced a new class of graph called an *ancestral graph*[5] capable of capturing the relationships between observed variables in the presence of both latent confounders and *selection bias*. The latter is the situation where the probability of inclusion of a data instance in the dataset depends upon one or more latent *selection variables*. An example might be where patients in a clinical trial do not complete the trial if they become seriously ill, and so are not present in the dataset. Crucially, DAGs are a special case of an ancestral graph, and ancestral graphs are closed under conditioning and marginalisation. This means that an ancestral graph can be used to represent the probability distribution of a partially observed DAG. Ancestral graphs have three types of edge (Zhang, 2008b):

- directed, e.g. $A \longrightarrow B$: The mapping between edge types in the ancestral graph and relationships in the underlying DAG is given in Richardson and Spirtes (2002) but is somewhat complicated. We first define "$A$ is an *ancestor* of $B$" to mean that there is a directed path from $A$ to $B$ with at least two directed edges. Directed edge $A \longrightarrow B$ in the ancestral graph means that $A$ is an ancestor or parent of $B$ and/or a selection variable, and that $B$ is not an ancestor or parent of $A$ nor of a selection variable ***in the underlying DAG***. Note, for example, that this edge type does not preclude a latent variable being the cause of both $A$ and $B$ as well (i.e., a latent confounder).
- bidirected, e.g. $A \longleftrightarrow B$: indicates that $A$ is not an ancestor or parent of $B$, $B$ is not an ancestor or parent of $A$, and neither are ancestors or parents of a selection variable. This edge type arises in the presence of latent confounders.
- undirected, e.g. $A - B$: $A$ is an ancestor or parent of $B$ or a selection variable *and B* is an ancestor or parent of $A$ or a selection variable.

---

[5] Earlier work by Spirtes et al. 1995 and 2000 represented the effect of latent and selection variables through a similar kind of mixed graph called an *inducing path graph (IPG)*. Ancestral graphs are a sub-class of IPGs which reveal more causal information and are easier to parameterise (Zhang, 2008a).

As the above shows, ancestral graphs primarily provide information about the ancestral and parental relationships in the underlying DAG, hence their name. Figure 11(b) shows the ancestral graph which represents the relationships between the observed variables in Figure 11(a). Richardson and Spirtes (2002) state two key conditions in the definition of an ancestral graph:

- there are no partially directed cycles. A partially directed cycle consists of an *anterior path* from $A$ to $B$ together with an edge $B \rightarrow A$ or $B \longleftrightarrow A$. An anterior path from $A$ to $B$ consists of edges with no arrows pointing towards $A$;
- for any undirected edge $A - B$, $A$ and $B$ should have no incoming arrows.

Many properties of ancestral graphs flow from these two conditions. In particular, that there can be at most one edge between each pair of variables, and that marginalisation and conditioning are closed. It also follows that ancestral graphs encode conditional independence relationships through a graphical criterion called *m-separation* which is analogous to d-separation for DAGs. In an ancestral graph, $\boldsymbol{S}$ *m-separates* $A$ from $B$ if all paths between $A$ and $B$ are blocked by $\boldsymbol{S}$. A path is blocked if at least one node on the path is either:

- a collider, defined in an ancestral graph as having two arrowheads incident to it, and neither it, nor any of its descendants, are in $\boldsymbol{S}$;
- or, is not a collider and is in in $\boldsymbol{S}$.

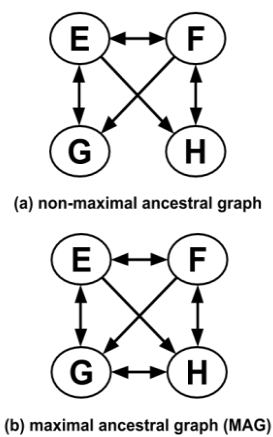

**(a) non-maximal ancestral graph**

**(b) maximal ancestral graph (MAG)**

**Figure 12** - Maximal Ancestral Graphs (Zhang, 2008b).

If two nodes are not adjacent in a DAG, this implies that there is a set of nodes which d-separates them. However, this does not follow for ancestral graphs. Figure 12(a) based on Zhang (2008b) illustrates this, since $G$ and $H$ are not adjacent, but there is no subset of the other nodes that m-separates them.

Therefore, a sub-class of ancestral graphs known as *Maximal Ancestral Graphs (MAG)* is defined which does have the property that non-adjacent nodes can be m-separated. Equivalently, this means that every absent edge in a MAG corresponds to a conditional independence relationship. A MAG can always be constructed from an ancestral graph by adding bi-directional edges such as $G \leftrightarrow H$ in Figure 12(b).

In the same way that an equivalence class of DAGs may be consistent with a given set of independence relationships, the independence relationships with latent and selection variables present may be consistent with multiple MAGs. Analogously to a CPDAG, the equivalence class of MAGs is represented by a *Partial Ancestral Graph (PAG)*. Constraint-based algorithms which take account of latent confounders and selection variables generally produce a PAG. There are three types of endpoint possible at each end of an edge in a PAG:

- an invariant arrowhead, marked as ">", indicating that all MAGs in the equivalence class have an arrowhead at that endpoint;
- an invariant tail, marked as "-", indicating that all MAGs in the equivalence class have a tail at that endpoint;
- a variant endpoint, marked as "o", indicating that some MAGs in the equivalence class have an arrowhead, and others a tail, at that endpoint.

So, for example, an edge $\circ \!\!\rightarrow$ in a PAG indicates that MAGs in the equivalence class may have $\rightarrow$ or $\leftrightarrow$ at that location, and similarly an edge $\circ - \circ$ in the PAG indicates that MAGs in the equivalence class can have a $\rightarrow$, $\leftarrow$, $\leftrightarrow$ or $-$ edge at that location. Note that the semantics of CPDAGs and PAGs are somewhat different. In particular, whereas a $-$ edge in a CPDAG indicates that the edge in equivalent DAGs can be either $\rightarrow$ or $\leftarrow$, a $-$ edge in a PAG indicates that the edge is $-$ in all equivalent MAGs.

### 3.4.2. *Fast Causal Inference (FCI) algorithm*

Spirtes et al. (1993; 2000) described the Fast Causal Inference (FCI) algorithm for structure learning without assuming causal sufficiency, though the causal Markov and causal faithfulness conditions are still assumed. FCI produces a Partially Orientated Inducing Path Graph (POIPG) – an earlier representation which is slightly less informative than a PAG. In broad overview, FCI is similar to PC in that it first determines the adjacencies in the POIPG, and then orientates edges. We recall that the PC adjacency phase is optimised by using conditioning sets of increasing size. The PC adjacency phase also makes use of the fact that, for a DAG, a Sepset must be a subset of the parents of $A$ or $B$, and so it need only consider conditioning sets which are subsets of the neighbours of A and B. The FCI adjacency phase also uses conditioning sets of increasing size. However, Sepsets in a MAG are a subset of $\boldsymbol{D\text{-}Sep}(A, B)$ (Spirtes et al., 2000, page 134) which in general contains nodes which are not adjacent to $A$ or $B$, as well as those that are. This necessitates a more complex strategy for determining adjacencies:

1. Firstly, an initial skeleton is estimated considering conditioning sets that are subsets of the neighbours of $A$ or $B$ . In general, this skeleton will have some extraneous adjacencies.
2. Secondly, a v-structure phase is performed to orientate some edges. The resulting graph allows us to identify nodes that are definitely not in $\boldsymbol{D\text{-}Sep}(A, B)$ and so conversely define a superset of $\boldsymbol{D\text{-}Sep}(A, B)$, denoted $\boldsymbol{Possible\text{-}D\text{-}Sep}(A, B)$.
3. Further edges may then be removed using subsets of $\boldsymbol{Possible\text{-}D\text{-}Sep}(A, B)$ as conditioning sets.

V-structure identification is then repeated on this new skeleton, followed by an orientation phase which is much more complex in POIPGs (and PAGs) than in PDAGs. Zhang (2008b) augmented the process of Spirtes et. al (2000) by defining eleven orientation rules that are said to produce a sound and complete PAG as the sample size $N \to \infty$; i.e., all arrowheads and edge tails are said to be correct and the maximum possible number of them are determined. Colombo and Maathuis (2014) applied the same techniques used in PC-Stable to amend FCI to produce the *FCI-Stable* algorithm, which removes the dependence of the result on node lexicographical ordering. FCI-Stable is often used as the benchmark when assessing learning in the presence of latent and selection variables.

### 3.4.3. Really Fast Causal Inference (RFCI) algorithm

Despite the presence of "Fast" in FCI's name, its adjacency determination is typically far more resource intensive than in the PC algorithm. Really Fast Causal Inference (Colombo et al., 2012) seeks to address this by reverting back to considering only conditioning sets that are the parents of nodes in the adjacency phase as PC does, and having just one adjacency phase instead of two as in FCI. The v-structure phase and one of the eleven orientation rules are also modified to avoid orientation errors that might occur due to the fact that the PC adjacency step is used rather than the more accurate FCI one. The authors showed that for a large class of graphs being learnt, this produced the same PAG that FCI would have. Moreover, when FCI and RFCI do produce different results, RFCI produces PAGs with extra edges, thus slightly weakening the meaning of an edge. On the other hand, the consequent reduction in CI tests, particularly high-order ones, meant that RFCI was around 250 times faster for some synthetic, sparse, high-dimensional graphs (n=500) with latent variables. It should be noted that all structural accuracy evaluations given in Colombo et al. (2012) were done against PAGs produced by FCI rather than against 'ground-truth' graphs.

### 3.4.4. Conservative Fast Causal Inference (CFCI) algorithm

As well as developing the RFCI algorithm, Colombo et al. (2012) also investigated modifying FCI by weakening the faithfulness assumption used to identify v-structures as Ramsey et al. (2006) had done in the Conservative PC algorithm. They identified v-structures as either ambiguous or unambiguous, and only used the latter in subsequent stages. This resulted in fewer arrowheads, smaller possible conditioning sets, and hence extra edges in the resultant PAG compared to FCI. The overall effect was to produce similar numbers of additional edges compared to FCI as RFCI had produced, though with edge orientation closer to FCI.

### 3.4.5. Fast Causal Inference Plus (FCI+) algorithm

Fast Causal Inference Plus (FCI+) centers around another approach to speeding up the adjacency phase of FCI (Claassen et al., 2013). It retains FCI's approach of using conditioning sets for $A, B$ that incorporate ancestors as well as just parents of $A$ and $B$, but focuses on efficiently identifying those cases where ancestors rather than just direct parents are in the Sepset. In doing so, they demonstrate that learning sparse causal graphs can be performed in polynomial time if a limit is placed on the node degree. In particular, in the worst case, FCI+ requires $O(n^{2(d+2)})$ CI tests, where $n$ is the number of observed variables, and $d$ the maximum node degree. This complexity is $O(PC^2)$, that is the square of what the PC algorithm requires. Although detailed performance results are not given, the authors suggested that cases where Sepsets do include non-parents are relatively rare, and so performance may in practice be relatively close to PC.

### *3.4.6. Mixed variable types – MGM-FCI-MAX algorithm*

Raghu et al. (2018) extended FCI to support a mixture of continuous and discrete variable types in their MGM-FCI-MAX variant of FCI. They introduced regression-based tests to detect conditional independence across different variable types. Orientation accuracy is also improved using the Sepsets with the highest p-value to identify v-structures as described in the PC-MAX algorithm (see subsection 3.2.6). This produces higher numbers of CI tests compared to FCI, and so the adjacency and v-structure phases are parallelised to counteract this efficiency drawback. The code parallelisation resulted in modest time savings of the order of 30% using six processor cores instead of one core. The algorithm achieves more substantial savings by another innovation of using a Mixed Graphical Model (MGM) undirected graph as input to the adjacency phases rather than a complete undirected graph as is customary. An MGM is an undirected graph which can represent conditional independence relationships between mixed variable types and was generated by the algorithm described by Lee and Hastie (2015). Combining all these innovations, MGM-FCI-MAX was said to achieve a better balance between precision and recall than CFCI and FCI, as well as significantly reduced runtimes when applied to networks consisting of 500 variables.

## 4. SCORE-BASED LEARNING

Score-based learning represents the other main class of BN structure learning and consists of two elements: a) a search strategy that determines which path to follow in the search space of possible graphs, and b) an objective function that can be used to evaluate each graph explored in the search space of graphs. The overriding challenge for score-based learning is to find high, or ideally the highest, scoring graphs amongst the vast number of possible graphs. As we have seen in the Introduction, a naïve exhaustive search where every possible graph is considered and scored is only feasible in problems with a handful of variables.

We first describe the objective function, which is pertinent to all score-based algorithms, in subsection 4.1, followed by the algorithms themselves. Score-based algorithms are the most diverse type of structure learning algorithms, and there are different ways one might choose to categorise them. Here, we opt to primarily organise them according to those which do not guarantee to return the highest scoring graph, known as *approximate* algorithms and described in subsection 4.2, and those which do offer that guarantee, known as *exact* algorithms and described in subsection 4.3. These different groups of score-based algorithms, and their evolution, are shown in different shades of blue in Figure 6.

The other defining characteristic of score-based algorithms is the search strategy. This is a combination of what search space is used, how the algorithm traverses that search space, and how that search space might be pruned (reduced). Perhaps the simplest score-based algorithm one might imagine is one that starts with an empty graph and greedily adds the arc which most increases the score subject to the restriction that it does not create a cycle in the graph. This process continues until it is no longer possible to find an arc addition that increases the score. The search space in this simple case would be *DAG space* (sometimes referred to as *structure* or *graph space*), and the traversal method is *add arc*. Since the algorithm greedily adds arcs, there is no guarantee it will find the highest scoring graph, and so it is an approximate algorithm.

Approximate algorithms which search DAG space are described in subsection 4.2.1. However, other kinds of search space have also been adopted. For example, subsection 4.2.2 describes approximate algorithms which explore equivalence class space, and subsection 4.2.3 covers those which explore *node* space. Node ordering is a topological ordering of the nodes in the DAG such that a node can only have parents which are higher up the ordering than it.

Note that a node ordering exists for a directed graph if and only if it is acyclic, and that in general a DAG may be consistent with multiple orderings as well as an ordering may be consistent with multiple DAGs.

**Table 3** - Characteristics of score-based algorithms reviewed, ordered chronologically (note that K2 is a type of Bayesian Dirichlet score). "Optimal DAG" in the final column refers to the DAG with the highest score over all possible DAGs with the dataset used to learn the DAG.

| Algorithm | Algorithm Name or Description | Reference | Exact | Search space | Search space traversal | Scores used | Type of output |
|---|---|---|---|---|---|---|---|
| K2 | K2 | Cooper & Herskovits, 1992 | No | DAG | Add arc | K2 | DAG |
| HC-Bouckaert | Hill Climbing by adding arcs | Bouckaert, 1994 | No | DAG | Add arc | K2 & MDL | DAG |
| TABU | Hill Climbing with Tabu list | Bouckaert, 1995 | No | DAG | Add/remove/reverse arc | K2 & BIC | DAG |
| HC | Hill Climbing | Heckerman et al., 1995 | No | DAG | Add/remove/reverse arc | BDe | DAG |
| MC$^3$ | Structure Monte Carlo Markov Chain | Madigan et al., 1995 | No | DAG | Add/remove arc | Posterior probability | DAG distribution |
| GA | Genetic algorithm | Larranaga et al., 1996 | No | Node ordering | Crossover and mutation operators | K2 | DAG |
| K2SN | K2 for Sorting Nodes | de Campos & Puerta, 2001 | No | Node ordering | Randomly generate ordering | K2 | DAG |
| VNS | Variable Neighbourhood Search | de Campos & Puerta, 2001 | No | DAG | Add/remove/reverse up to n arcs | K2 | DAG |
| GES | Greedy Equivalence Search | Chickering , 2002 | No | Equivalence class | Insert/Delete operations (see subsection  4.2.2) | BDeu | CPDAG |
| Order-MCMC | Order Monte-Carlo Markov Chain | Friedman & Koller, 2003 | No | Node ordering | Swap adjacent nodes | Posterior probability | Feature probability |
| OptOrd | Dynamic Programming | Singh and Moore, 2005 | Yes | Node ordering | Add/remove first node in order | BDeu | Optimal DAG |
| OBS | Ordering Based Search | Teyssier and Koller, 2005 | No | Node ordering | Swap adjacent nodes | BDeu | DAG |
| Hybrid MCMC | Hybrid Monte Carlo Markov Chain | Eaton & Murphy, 2007 | No | DAG | Add/remove/reverse arc | BDeu | Feature probability |
| B&B | Branch and Bound | De Campos et al., 2009 | Yes | Directed Graph | Split graph at cycles | AIC or BIC | Optimal DAG |
| GOBNILP | Globally Optimal BN using ILP | Cussens, 2011 | Yes | Directed Graph | Integer Linear Programming | BDe | Optimal DAG |
| GIES | Greedy Interventional Equivalence Search | Hauser and Bühlmann, 2012 | No | Equivalence class | Insert/Delete operations (see subsection 4.2.2) | BIC | CPDAG |
| BCO | Bee Colony Optimisation | Ji et al., 2013 | No | DAG | Add/remove/reverse arc or swap parents | K2 | DAG |
| A* | A* Search | Yuan and Malone, 2013 | Yes | Node ordering | Add/remove first node in order | BIC | Optimal DAG |
| CPBayes | Constraint Programming | Van Beek and Hoffman, 2015 | Yes | Node ordering | Swap adjacent nodes | BDe & BIC | Optimal DAG |
| ASOBS | Acyclic Selection OBS | Scanagatta et al, 2015 | No | Node ordering | Swap adjacent nodes | BIC* | DAG |
| GSMAG | Greedy Search for MAGs | Triantafillou & Tsarmadinos, 2016 | No | MAG | Add/remove/reverse/convert bi/directed edges | BIC for MAG | MAG |
| Partition MCMC | Partition Monte Carlo Markov Chain | Scanagatta et al., 2017 | No | Order partitions | Split/merge partition | BGe | DAG |
| INOBS | Insert Neighbourhood OBS | Lee and van Beek, 2017 | No | Node ordering | Insert/swap adjacent nodes | BDeu & BIC | DAG |
| MINOBS | Memetic Insert Neighbourhood OBS | Lee and van Beek, 2017 | No | Node ordering | Insert/swap adjacent nodes + mutation/crossover | BDeu & BIC | DAG |
| FGES | Fast Greedy Equivalence Search | Ramsey et al., 2017 | No | Equivalence class | Insert/Delete operations (see subsection 4.2.2) | BDeu & BIC | CPDAG |
| WINASOBS | Window Acyclic Selection OBS | Scanagatta et al., 2017 | No | Node ordering | Move blocks of nodes | BIC | DAG |
| GOBNILP-DEV | Globally Optimal BN using ILP | Liao et al., 2019 | Yes | Directed Graph | Integer Linear Programming | BDeu & BIC | Highest scoring DAGs |
| K2-Improved | K2 with improved ordering | Behjati and Beigy, 2020 | No | DAG | Node order, then add arc | BIC | DAG |
| FGES-Merge | Merge sub-graphs learnt with FGES | Bernaola et al., 2020 | No | Equivalence class | Insert/Delete operations (see subsection 4.2.2) | BIC | CPDAG |
| ELSA | Improved acyclicity checks in CPBayes | Troser et al., 2021 | Yes | Node ordering | Swap adjacent nodes | BDeu & BIC | Optimal DAG |
| MAHC | Model Averaging Hill Climbing | Constantinou et al., 2022 | No | DAG | Add/remove/reverse arc | BIC | DAG |

This categorisation by search space is also followed for the exact algorithms. Pruning the search space is particularly important for exact algorithms where the pruning rules must be *sound* so as to guarantee that the pruned space still contains the optimal solution, whereas *heuristic pruning* does not offer this guarantee. Table 3 describes the search space and the search space traversal method used by the score-based algorithms covered in this paper, as well as whether they are approximate or exact algorithms. It also includes the objective function used in the original paper proposing the algorithm. Note that Scutari et al (2019a) argued that the choice of algorithm and score used should be independent, and indeed, many BN tools support using different score functions for each algorithm. Thus, this column does not necessarily indicate a fundamental restriction on the scores that can be used with each algorithm, rather it gives a historical view on preferred scores at the time of their introduction. Finally, Table 3 describes the output graph type each algorithm produces. Approximate algorithms will typically produce a single DAG with a locally optimum score, whereas exact search algorithms will return a DAG with the globally optimum (that is, highest possible) score. Lastly, algorithms searching in equivalence class space will return a CPDAG.

### 4.1. Objective functions

Objective functions fall under two categories: the *Bayesian* scores which generally focus on the goodness of fit and allow the incorporation of prior knowledge, and *information-theoretic* scores which explicitly consider model complexity in addition to the goodness of fit, aiming to avoid model overfitting. Importantly, a score is said to be *decomposable* if the score of a graph can be decomposed into a sum of scores each associated with a node in the graph. Decomposable scores mean that only the scores for nodes affected by a graph change in a search process need to be re-computed, rather than re-computing the score of the whole graph for every single graph modification. As a result, a decomposable score greatly improves computational efficiency and virtually all algorithms employ them.

As noted in the Introduction, all the DAGs in an equivalence class entail the same conditional independence relationships, and therefore there is no reason for preferring one of them above the others on the basis of the observational data alone[6]. Therefore, the objective function is usually specified so it gives the same score to all DAGs in an equivalence class – a property known as *score equivalence*. Most commonly used scores do have this property. However, it is worth noting that approximate and exact algorithms that use score equivalent objective functions often just return a single result DAG. In that case, we should regard the output DAG as being a representative of the equivalence class to which it belongs. Indeed, the particular DAG within an equivalence class that the algorithm returns is usually just an artefact of the dataset (Constantinou et al., 2021b). It may depend on the lexicographical ordering of the variable names, or the order in which the variables in the dataset are encountered.

### 4.1.1. Bayesian scores

Bayesian scoring functions return a relative posterior probability for a graph conditioned on the data, taking into account prior beliefs about the graphical structure and/or dependence relationship parameters. The approach provides a theoretical underpinning to assign a posterior probability to each possible structure, something that constraint-based approaches do not offer. This in turn allows Bayesian Model Averaging (BMA) where, for instance, the posterior probability of a given feature such as a specific arc can be averaged across a set of likely structures.

---

[6] Note that if we have data that includes the effects of interventions then we may be able to orientate some of the undirected edges in the equivalence class, and come closer to fully specifying the causal DAG.

Most commonly, one assumes that all graph structures are equally probable a priori. For discrete data, we generally assume a Dirichlet prior for the parameters which gives rise to the well-established general Bayesian Dirichlet (BD) score which, in its general form, is not *score equivalent* (Heckerman et al., 1995). Formally, the general BD score is defined as:

$$S_{BD}(G, D) = \log P(G) + \sum_{i=1}^{n} \sum_{j=1}^{q_i} \left[ \log \frac{\Gamma(N'_{ij})}{\Gamma(N_{ij} + N'_{ij})} + \sum_{k=1}^{r_i} \log \frac{\Gamma(N_{ijk} + N'_{ijk})}{\Gamma(N'_{ijk})} \right]$$

where $\Gamma$ is the Gamma function, $i$ is the index over the $n$ variables, $j$ is the index over the $q_i$ combinations of values of the parents of the node $X_i$, and $k$ is the index over the $r_i$ possible values (states) of node $X_i$. Further, $N_{ijk}$ is the number of instances in the data $D$ where node $X_i$ has the $k^{th}$ value, and its parents have the $j^{th}$ combination of values, and $N_{ij} = \sum_{k=1}^{r_i} N_{ijk}$ representing the total number of instances in the data $D$ where the parents of node $X_i$ have the $j^{th}$ combination of values. Lastly, $N'_{ijk}$ and $N'_{ij} = \sum_{k=1}^{r_i} N'_{ijk}$ are defined analogously based on prior beliefs of these values. $P(G)$ is the prior probability of a particular graph structure which is generally assumed to be the same for all graphs and so can be ignored.

A drawback of the general BD score is that it requires the user to specify the values of $N'_{ijk}$ individually, which renders it impractical. The K2 score is the BD score where $N'_{ijk} = 1$, (Cooper and Herskovits, 1992) and simplifies the general BD score to:

$$S_{K2}(G, D) = \log P(G) + \sum_{i=1}^{n} \sum_{j=1}^{q_i} \left[ \log \frac{(r_i - 1)!}{(N_{ij} + r_i - 1)!} + \sum_{k=1}^{r_i} \log(N_{ijk}!) \right]$$

The K2 score also is not *score equivalent.* Heckerman et al. (1995) introduced the *score equivalent* BDe score, defined as

$$S_{BDe}(G, D) = \log P(G) + \sum_{i=1}^{n} \sum_{j=1}^{q_i} \left[ \log \frac{\Gamma(N' \sum_{k=1}^{r_i} \theta'_{ijk})}{\Gamma(N_{ij} + N' \sum_{k=1}^{r_i} \theta'_{ijk})} + \sum_{k=1}^{r_i} \log \frac{\Gamma(N_{ijk} + N'\theta'_{ijk})}{\Gamma(N'\theta'_{ijk})} \right]$$

Here, $\theta'_{ijk}$ is the prior conditional probability of node $X_i$ having the $k^{th}$ value given the parents have the $j^{th}$ combination of values in the *prior distribution.* $N'$ is the *equivalent sample size* (ESS, also sometimes known as the *imaginary sample size*, ISS) and expresses our confidence in the prior parameters.

The most commonly used Bayesian score is the BDeu score (Buntine, 1991; Heckerman et al., 1995) which is a special case of BDe where the prior parameters are set to $\theta'_{ijk} = 1/r_i q_i$ for all $i, j, k$ leading to the following definition:

$$S_{BDeu}(G, D) = \log P(G) + \sum_{i=1}^{n} \sum_{j=1}^{q_i} \left[ \log \frac{\Gamma(\frac{N'}{q_i})}{\Gamma\left(N_{ij} + \frac{N'}{q_i}\right)} + \sum_{k=1}^{r_i} \log \frac{\Gamma\left(N_{ijk} + \frac{N'}{r_i q_i}\right)}{\Gamma\left(\frac{N'}{r_i q_i}\right)} \right]$$

Cooper and Yoo (1999) define a variant of BDeu which is suitable for a mix of observational and interventional data where the terms that express the likelihood of the data given a particular structure are left out for nodes that are intervened on. They showed that using this approach, a combination of observational and experimental data was the most effective at identifying causally-related nodes.

BDeu is score equivalent but requires the user to choose a suitable value for ESS ($i.e., N$). Unfortunately, BDeu, and hence the graphs learnt using it, are sensitive to the value of ESS chosen, and it is difficult to decide what value to use for ESS. As might be expected, large values of ESS will tend to regularise the parameter values (Heckerman et al., 1995). What is rather more surprising is the effect of ESS on the learnt graph structure. Steck and Jaakkola (2002) found that as ESS tends to zero, arc deletion is favoured producing sparser graphs. Similarly, Silander et al. (2006) observed that the number of arcs rose as ESS was increased. Ueno (2010) provided a detailed asymptotic analysis of BDeu supported by empirical experiments. This work showed that different elements of BDeu responded differently to ESS, with the complex behaviour heavily influenced by sample size and the skewness of the parameters. This work also showed that the K2 score approximates the BIC asymptotically as the sample size tends to infinity. The author recommended that ESS be set to 1 for small sample sizes.

Correia et al. (2019) introduced the concept of a *robustness interval* defined as the ESS range over which all the graphs generated are members of the same equivalence class. They found that this range increased with sample size, but that large amounts of data were required to achieve a reasonably wide robustness interval for ESS of [0.1, 4.0]. All 15 real-world datasets examined did not have sufficient data to achieve this robustness interval, leading them to conclude that "*almost every real-world dataset might be too sparse for BDeu*". The robust interval calculated did not include the widely adopted value of ESS = 1 in 11/15 datasets.

Scutari (2016) introduced a new BD score, BDs, aiming to produce better results with sparse datasets where some possible combinations of parental values are not present in the dataset. BDs has the same algebraic form as BDeu given above, the difference being the way $q_i$, the number of parental value combinations, is calculated. As an illustration, suppose node C has parents A and B, and A can take three possible values, and B two possible values, giving 6 possible combinations of parental values, but suppose only 4 of these combinations are actually present in the data. BDs will use $q_i = 4$, and BDeu will use $q_i = 6$. This paper also showed that the uniform structural prior usually used with BDeu favoured the inclusion of arcs, and suggested a new structural prior named *marginal uniform* which weighted arc addition and deletion equally. The combination of BDs and the marginal uniform prior outperformed the traditional BDeu and structural uniform prior combination in terms of structural accuracy and the likelihood of the observed data given the learnt structure in all sixty combinations of BN and sample size tested. The improved structural accuracy resulting from using the combination of BDs and the marginal uniform prior instead of BDeu was more pronounced at lower sample sizes. It should be noted that a disadvantage of BDs is that it is not score equivalent when there are missing parental value combinations, a situation likely to occur in all but very large sample sizes.

The analogous score to the Bayesian Discrete equivalent scores for continuous variables is the Bayesian Gaussian equivalent score (BGe) defined in Geiger and Heckerman (2002) and subsequently corrected in Kuipers et. al. (2014). The prior beliefs of the parameter values are encapsulated as the parameters of a Normal-Wishart distribution in an analogous fashion to the Dirichlet prior for discrete Bayesian scores.

### *4.1.2. Information-theoretic scores*

Information-theoretic scores aim to avoid over-fitting by balancing the goodness of fit with model dimensionality given the available data. The most commonly used scores include the *Bayesian Information Criterion* (BIC) which is also known as the *Minimum Description Length* (MDL) (Suzuki, 1993, 1999), the *Akaike Information Criterion* (AIC) (Akaike, 1974), the *Mutual Information Test* (MIT) (de Campos, 2006), the *Normalised Maximum Likelihood* (NML) (Rissanen, 1996), the *factorized Normalised Maximum Likelihood* (fNML) ) (Silander et al., 2008), and the *quotient Normalised Maximum Likelihood* (qNML) (Silander et al., 2018). The general form of these scores can be expressed as:

$$S(G, D) = \log\left[\hat{p}(D|G)\right] - \Delta(D, G)$$

where $\log\left[\hat{p}(D|G)\right]$ denotes the goodness of fit as measured by the log likelihood of the data given the graph, in the case where the distribution parameters, $\boldsymbol{\Theta}$, take their Maximum Likelihood Estimation (MLE) values, and $\Delta(D, G)$ is a function which penalises graph complexity. The detailed expression of $\log\left[\hat{p}(D|G)\right]$ for discrete variables is

$$\log[\hat{p}(D|G)] = \sum_{i=1}^{n} \sum_{j=1}^{q_i} \sum_{k=1}^{r_i} N_{ijk} \log\frac{N_{ijk}}{N_{ij}} = S_{LL}(G, D)$$

Setting $\Delta(D, G) = 0$ removes the dimensionality penalty and makes the score equivalent to the *Log-likelihood* score $S_{LL}(G, D)$. Since each arc addition increases $S_{LL}(G, D)$, this score will favour denser graphs.

In the AIC score, the complexity penalty is just the number of free parameters in the model, $F$, defined as:

$$F = \sum_{i=1}^{n} (r_i - 1)q_i,$$

so that

$$S_{AIC}(G, D) = S_{LL}(G, D) - F$$

The AIC score represents a rather soft penalty in terms of the number of free parameters. As a result, the AIC score tends to favour networks with a higher number of free parameters compared to BIC which is represented by

$$S_{BIC}(G, D) = S_{LL}(G, D) - \frac{\log N}{2} \cdot F$$

where $N$ is the sample size. Note that in BIC, and even more so AIC, the relative influence of the complexity penalty decreases as $N$ grows, implying that increasing sample size will eventually allow $LL$ to dominate the score. The BIC score is widely popular and has been found to be able to learn the true network faster than other scoring functions such as AIC, BDeu and fNML (Liu et al., 2012).

While both the AIC and BIC scores can recover the underlying network when the sample size is sufficiently high, they are suboptimal with limited sample sizes. To that end, Silander et al. (2010) proposed the *factorized Normalized Maximum Likelihood (fNML)* score. fNML is based on the Normalised Maximum Likelihood (NML) distribution which gives the probability of every possible dataset of sample size $N$ for a specific graph $G$. An NML-based score is not decomposable, so Silander et al. (2008) define a decomposable variant:

$$S_{fNML}(G, D) = S_{LL}(G, D) - \sum_{i=1}^{n} \sum_{j=1}^{q_i} \zeta_{N_{ij}}^{r_i},$$

where $\zeta_{N_{ij}}^{r_i}$ is the *stochastic complexity* which reflects the amount of information required to encode all possible combinations of $N_{ij}$ values of a multinomial variable with $r_i$ different possible values, where $\zeta_N^r$ is defined:

$$\zeta_N^r = \sum_{k_1 + k_2 + \cdots + k_r = N} \frac{N!}{k_1! \, k_2! \ldots k_r!} \prod_{j=1}^{r} \left( \frac{k_j}{N} \right)^{k_j},$$

This stochastic complexity can be computed in linear time using a recursive formula (Kontkanen and Myllymäki, 2007):

$$\zeta_N^r = \zeta_N^{r-1} + \frac{N}{r-2} \cdot \zeta_N^{r-2}$$

fNML was shown to perform well on small datasets (Silander et al., 2010; Liu et al., 2012). fNML is not score equivalent, and Silander et al. (2018) proposed another variant of a NML-based score, quotient Normalised Maximum Likelihood (qNML) which is score equivalent.

Finally, the MIT score was proposed by de Campos (2006) and is expressed as

$$S_{MIT}(G, D) = \sum_{\substack{i=1 \\ \boldsymbol{Pa}(X_i) \neq \emptyset}}^{n} \left( 2N \cdot MI(X_i, \boldsymbol{Pa}(X_i)) - \sum_{j=1}^{|\boldsymbol{Pa}(X_i)|} \epsilon_{\alpha, l_{ij}} \right)$$

where $MI(X_i, \pi_i)$ is the *mutual information* between variable $X_i$ and its parents $\boldsymbol{Pa}(X_i)$. $\epsilon_{\alpha, l_{ij}}$ is a threshold value for the mutual information between a parent and the variable $X_i$ below which we assume independence between that parent and the variable conditional on the remaining parents. $\epsilon_{\alpha, l_{ij}}$ depends on the statistical significance level $\alpha$ chosen and $l_{ij}$ which is the number of degrees of freedom based on the number of states of the parents. Thus, this score might be regarded as a "hybrid" score since it involves considerations of conditional independence. Furthermore, the summation of $2N \cdot MI(X_i, \boldsymbol{Pa}(X_i))$ results in an expression proportional to the log likelihood, so the MIT score is another example of a penalised log likelihood score. Using a simple hill-climbing score-based algorithm (see subsection 4.2.1), de Campos (2006) showed that, according to their empirical experiments, MIT achieves better structural accuracy and data fitting than K2, BIC and BDeu scores. Notice that, MIT score is decomposable but not score equivalent.

Compared with Bayesian scoring functions, information-theoretic scoring functions (excluding MIT score which requires the significance level α) are objective and feature no prior parameters, which avoids the sensitivity problems. Therefore, when users have little background knowledge about the target network, information-theoretic scoring functions may be preferred.

### 4.2. *Approximate score-based algorithms*

This section describes algorithms which do not guarantee to return the highest possible scoring graph. Instead, they tend to return a graph with a locally maximum score, although it is still possible that they will return a graph with the globally maximum score. It should also be noted that some approximate algorithms do offer a guarantee to return the optimal graph with $probability \rightarrow 1$ as the sample size $N \rightarrow \infty$, which is known as *(classical) consistency* - or as the algorithm being *asymptotically correct*. Furthermore, some algorithms also offer *high-dimensional consistency* which is where they will return the optimal graph with $probability \rightarrow 1$, as both the sample size and number of variables grow $N \rightarrow \infty, n \rightarrow \infty$.

#### 4.2.1. Approximate search of DAG space

One of the earliest BN structure learning algorithms was the *K2* algorithm by Cooper and Herskovits (1992), which assumes that a node ordering is already known. The algorithm works down the ordered list of nodes and greedily adds arcs from the candidate parents higher up the list to increase the K2 score maximally. Note that K2 does not consider all possible parent sets for each node and therefore cannot guarantee to find the highest scoring DAG for a particular node ordering.

```
algorithm HC is
    input: dataset D
    output: DAG G

1   G := empty DAG      // DAG, initially empty

2   repeat

        // possible changes mustn't create a cycle, and can delete
        // or reverse arcs currently in G, or add an arc to G
        // variable delta holds score change for each possible arc change

3       for each possible arc change in G do
4           if delta[change] needs calculating or recalculating
5               delta[change] := delta(G, change, D)

6       if max(delta[change]) > 0
7           change := change corresponding to max(delta[change])
8           G := G + change

9   until max(delta[change]) ≤ 0

10  return G
```

**Figure 13** - Pseudo-code for hill-climbing (HC) algorithm. Program code keywords are coloured blue, comments in grey, key variables in red, and application-specific complex operations or conditions in black.

Bouckaert (1994) removed the restriction of having a predefined node ordering and describes a general *hill-climbing* (HC) greedy search algorithm over the space of DAGs. This is perhaps the simplest and the most commonly used search strategy. Pseudo-code for HC is shown in Figure 13. At each iteration, HC explores all the neighbouring DAGs $G'$ of the current DAG $G$ which can be formed by adding an arc to $G$, or (in later variants) removing or reversing an arc in $G$. The change in score corresponding to each $G'$ is stored in the delta variable in the pseudo-code in Figure 13, and the graph modification with the largest delta applied. If no neighbouring DAGs increase the score, then we have reached a local, or occasionally a global, maximum and the DAG is returned as the result. The starting point for HC search can be any DAG such as a random one, one produced by another structure learning algorithm, or even one based on expert knowledge. However, it typically starts from the empty graph. HC is a very efficient algorithm, however it may 'get stuck' on a poor local maximum score.

Several techniques are adopted to escape local maxima. Heckerman et al. (1995) used *local restarts* where random perturbations are made to the DAG at the local maximum, and hill-climbing restarted from the perturbed DAG. Bouckaert (1995) employs a *tabu list* to prevent returning to DAGs recently visited, and permits some changes to the graph where the score is allowed to decrease. Figure 14 shows the pseudo-code for the Tabu algorithm, the key differences from HC being a tabu_list structure which maintains a list of the most recently visited DAGs on line 9, and the fact that possible changes to a graph cannot result in a DAG in tabu_list on line 4. This approach encourages moving into new regions that may contain an improved local maximum. The tabu approach can make runtime less predictable and may be more susceptible to noise than plain hill climbing (Constantinou et al., 2021b). Despite its simplicity, HC remains a very competitive algorithm (Scutari et al., 2019a).

```
     algorithm TABU is
         input: dataset D
         output: DAG G

   1     G := empty DAG        // DAG, initially empty
   2     tabu_list := []       // fixed length list of last DAGs visited

   3     repeat

             // possible changes as for HC except also the change
             // cannot result in a DAG currently in tabu_list

   4         for each possible arc change in G do
   5             if delta[change] needs calculating or recalculating
   6                 delta[change] := delta(G, change, D)

   7         change := change corresponding to max(delta[change])
   8         G := G + change
   9         add G to tabu_list

   10    until stop_condition  // e.g. limit on number of score decreases

   11    return G
```

**Figure 14** - Pseudo-code for Tabu algorithm. Program code keywords are coloured blue, comments in grey, key variables in red, and application-specific complex operations or conditions in black.

De Campos and Puerta (2001) used *variable neighbourhood search* (VNS) which widens the local neighbourhood explored by considering graphs resulting from changing several arcs in the current graph. Thus, each iteration of the graph may contain multiple differences from the previous iteration, whereas classic hill-climbing makes one change at a time. De Campos and Puerta (2001) found that this approach achieved better results than plain hill-climbing and hill-climbing with random restarts, although the results were based on a small set of experiments learning the Alarm network (Beinlich et al., 1989). Model Averaging Hill Climbing, MAHC (Constantinou et. al., 2022) is a recent algorithm which also considers the scores of graphs beyond one move ahead. However, it does so in a way that averages scores across these more distant graphs, and makes a single change to move to the neighbouring graph where the mean score of *its* neighbours is the highest.

Recognising the importance of providing a good node ordering to K2, Behjati and Beigy (2020) focus on determining such an ordering. To do this, their improved K2 algorithm first determines the highest scoring set of parents for each node separately and constructs the directed graph using these parent sets. This directed graph is generally cyclic, and if so, it is decomposed into strongly connected components (SCCs). The graph of these SCCs is itself a DAG of SCCs which defines an ordering of SCCs. This approach is used recursively on SCCs until a node ordering is produced which is then used with the original K2 algorithm. The authors evaluated this improved K2 algorithm against the original K2, HC, GES (see subsection 4.2.2), and an exact score-based algorithm, GOBNILP (subsection 4.3.2). The evaluation learnt well-known networks from the bnlearn repository (Scutari, 2021) with between 5 and 441 variables. GOBNILP is guaranteed to return the highest scoring graph, but the enhanced K2 algorithm nearly always produced the highest scoring graphs amongst the other algorithms and was generally the fastest algorithm. The authors did not report the structural accuracy of the learnt graphs, however.

The approximate score-based algorithms considered so far have focussed on learning a single high-scoring graph. This can be a reasonable approach for small networks with large amounts of data where the highest scoring DAG may be much more likely than any other model (Heckerman et al., 1997). However, it is less appropriate for complex models with small amounts of data. Friedman and Koller (2003) argued that for the gene expression data they studied, there might be many models with similar high scores, and that any single model selected might be very sensitive to the actual data instances used for learning. In this circumstance, an approach which generates a sample of plausible DAGs, and which reflects the posterior probability distribution across all possible DAGs, may be more appropriate. This might also show how "peaked" the distribution is, and so offer some insight into how confident one might be in any particular DAG.

Markov Chain Monte Carlo (MCMC) is a well-established technique for sampling from complex high-dimensional probability distributions, such as the posterior distribution of DAGs. In the context of structure learning, each state in the Markov chain represents a different model, such as a DAG or node ordering. The Metropolis-Hastings (MH) algorithm is the most common MCMC variant used. MH has a proposal distribution which defines the probability of specific state changes such as a particular edge addition or node order swap at each step in the chain. A change is randomly selected according to the proposal distribution and then accepted or not using a stochastic acceptance condition. The proposal distribution and acceptance condition are chosen so that models with high posterior probabilities are preferred. Provided certain conditions are met, the states generated by the Markov chain stabilise to a stationary distribution of models which represents a sample of the true posterior distribution. These conditions include that the chain must be *irreducible* such that every model is reachable from every other model in a finite number of steps. Madigan et al. (1995) used MH to sample DAGs in the MC$^3$ algorithm. The proposal distribution used gives the same non-zero probability for each possible single edge addition or deletion, and one of these changes is randomly selected. The acceptance condition always accepts a change that increases the posterior probability of the DAG, but may reject the change if it decreases the posterior probability. The more the posterior probability decreases, the more likely the change is to be rejected.

A concern with MC$^3$ is slow convergence to the stationary distribution, a problem known as slow mixing. Eaton and Murphy (2007) therefore proposed a *Hybrid MCMC*[7] algorithm that uses an exact score-based algorithm, Dynamic Programming (see subsection 4.3.1), to develop a global proposal distribution. They showed that the Hybrid MCMC method converges faster than both the MC$^3$ and Order-MCMC methods (see subsection 4.2.3). Grzegorczyk and Husmeier (2008) also improved the convergence of DAG sampling using the *REV-MCMC* algorithm that adds a new edge reversal proposal that re-samples all possible parents of the two endpoints of the reversed edge.

Goudie and Mukherjee (2016) used a special case of Metropolis-Hastings known as Gibbs Sampling, where proposed changes are always accepted. Gibbs sampling can lead to slow mixing which they counteracted with a broader proposal distribution that considers all possible changes made to the parents of a block of nodes. Experiments showed that this Gibbs sampling mixed better than MC$^3$ and REV-MCMC, as well as producing more accurate DAGs than both of them, and the constraint-based PC algorithm. Goudie and Mukherjee (2016) evaluated the DAG with the highest posterior probability, the Maximum A Posteriori (MAP) DAG, as being the single 'best' DAG from the sample. However, they also evaluated a single DAG constructed using arcs which have a marginal posterior edge probability above 0.5 across the sample, an example of *model averaging* whereby a single DAG is constructed from a collection of DAGs.

---

[7] Note that, following normal usage, we use "hybrid" in this paper to mean a combination of constraint-based and score-based approaches, whereas, although this algorithm is named "Hybrid MCMC", it uses a combination of exact and approximate score-based approaches.

The Birth-and-Death algorithm (Jennings and Corcoran, 2018) treats structure learning as a continuous time Markov chain process where the waiting times (intervals) between changes to the DAG follow exponential distributions. At any given point in the learning process there is a 'birth' rate for each possible edge addition, and a 'death' rate for deleting edges. The proportion of time each edge exists represents the posterior probability of that edge. The authors demonstrated superior mixing to Metropolis-Hastings approaches but did not report any accuracy evaluations.

Many approaches inspired by biological systems have also been proposed and include Genetic Algorithms (Larranaga, 1996b), Ant Colony (de Campos et al., 2002), and Bacterial Foraging (Yang et al., 2016). These explore multiple DAGs and therefore are similar to the sampling approaches in that respect. They employ methods adopted from biology, for example, genetic mixing, mutation, swarming, foraging and selection of the fittest, to generate and select a high scoring graph. Genetic mixing, in other words breeding, is typically implemented by a merge operation where edges are taken from two graphs to create a new graph. Mutation is implemented by making random changes to the graph edges. To take one biological inspired algorithm in a little more detail, Bee Colony Optimisation (Ji et al., 2013) models the roles of real bees in a hive to search DAG space:

- *employed bees* perform local search in a DAG neighbourhood moving to the <u>first</u> neighbour with an increased score using add, delete, and reverse arc, and swap parents operators;
- *onlooker bees* perform more knowledgeable searches by moving to the <u>best</u> scoring neighbour which can be viewed as a form of hill climbing;
- if a bee gets stuck on a local maximum it becomes a *scout bee* and moves to a randomly generated DAG to begin searching in a new region, which is analogous to a random restart in hill climbing;
- pheromone is deposited on the best solution at each iteration to attract bees to high scoring regions.

### 4.2.2. Approximate search of equivalence class space

Chickering (2002) introduced the *Greedy Equivalence Search* (GES) algorithm which searches across the space of Markov equivalence classes rather than DAGs. A Markov equivalence class may contain several DAGs and hence the search space of Markov equivalence graphs is always smaller than the corresponding DAG space. Although Gillispie and Perlman (2002) suggest that most equivalence classes contain few DAGs, this is still sufficient to speed up search considerably.

GES has an insert and then a delete phase. In the first phase, each insert operation performs a change equivalent to determining all the DAGs consistent with the current CPDAG (equivalence class), adding whichever arc increases their individual score the most, and then selecting the CPDAG corresponding to highest scoring DAG of all. This logic means that each insert operation on the CPDAG represents the addition of either a directed or an undirected edge, and may be accompanied by several other edge (re)orientations. Note that GES implements this process more efficiently than this explanatory description would suggest. The insert phase continues until there is no further insert operation which would increase the score. The delete phase then proceeds in an analogous fashion until a final maximum scoring CPDAG is produced. Importantly, and surprisingly for an approximate algorithm, GES offers a guarantee of classical consistency. That is, it is guaranteed to produce a CPDAG which perfectly matches the conditional independence relationships in the data as $N \rightarrow \infty$. Ramsey et al. (2017) described *fast GES* (FGES) which optimises GES by parallelising operations and caching scores where possible.

Greedy Interventional Equivalence Search (GIES) by Hauser and Bühlmann (2012) is a generalisation of GES which supports learning from datasets[8] which all have the same set of variables and an assumed common underlying causal model, but where interventions have been performed on different sets of variables in each dataset. One of the datasets may be observational, that is, have no intervention variables. Like GES, GIES searches equivalence space in an addition and then deletion phase, but the equivalence graphs (which they term *interventional essential graphs*) are all consistent with the intervention targets across all datasets. The authors evaluate GIES against GES, learning random Gaussian graphs with number of variables, $n = \{10, 20, 30, 40\}$, and a sample size between 50 and 10,000, and using a modified SHD suitable for interventional settings (Kalisch and Bühlmann, 2007). The evaluation uses one, two or four intervention nodes, and varies the number of intervention datasets, each having a different randomly chosen set of intervention nodes. The authors find that GIES orientates more edges and has increasingly better accuracy than GES as the number of different intervention datasets or number of intervention targets grows.

Bernaola et. al. (2020) introduce FGES-Merge which is focussed on learning very large networks, with tens of thousands of variables, some with very large degree, typical of those encountered when modelling gene regulation networks. FGES-Merge uses FGES to learn sub-graphs around each node and then merges these to create the whole graph. The nodes chosen to be in each sub-graph learnt separately by FGES are those which have the largest BIC scores when treated as parents of the particular node. The number of nodes in each sub-graph is limited to a specified maximum, and so some edges for high degree nodes may be omitted, but they may be re-introduced from neighbouring nodes when the sub-graphs are merged. The union of the sub-graphs forms the final graph, with cycles and weaker arcs being eliminated. FGES-Merge was found to be more accurate than all but one of the other BN structure learning algorithms evaluated in the DREAM5 gene network modelling challenge (Marbach et. al., 2012).

Madigan et al., (1996) noted that MCMC sampling over DAG space may be inefficient since DAGs in the same equivalence class have the same posterior probability, and instead proposed using Metropolis-Hastings sampling over equivalence classes. In this case, the proposal distribution must include one and two edge changes so that the chain is irreducible and can therefore sample the whole equivalence class space.

### 4.2.3. *Approximate search of node-ordering space*

Searching over a node ordering space offers some potential benefits (Teyssier and Koller, 2005):

---

[8] In their paper, Hauser and Bühlmann (2012) describe the problem that GIES is tackling as structure learning from a *single* dataset, but where different intervention targets can be specified for different sets of rows. In this review, we prefer to describe this as learning from *multiple* datasets where each dataset has the same variables and assumed causal model, but with different intervention targets for each dataset. We do this so that the problem description is as common as possible across different algorithms.

- ordering-based space has complexity $2^{O(n \log n)}$ which is considerably smaller than $2^{\Omega(n^2)}$ for DAG space;
- each change to the ordering represents a larger change to the current hypothesis than those typically performed in DAG space, and so may be better at avoiding local maxima;
- since node ordering imposes acyclicity, the algorithms can select the parents of one node independently of any other node.

Larranaga et al. (1996a) used a genetic algorithm where the individual's chromosomes represented the node ordering. The fitness of an individual is assessed using the K2 algorithm and score (see subsection 4.2.1) to determine a high scoring DAG consistent with the ordering. An initial population of individuals is created which is then iterated through generations as follows:

1. pairs of high scoring parents are selected, and their children are created through a *crossover operator* which forms a new ordering from parts of each parent's ordering;
2. children have a small probability of the node ordering *mutating* so that new regions are explored;
3. low scoring individuals are pruned from the population to return it to its original size.

The K2SN algorithm by deCampos and Puerta (2001) uses a very simple strategy to explore node-ordering space. It randomly generates orderings and then uses the K2 algorithm to find a high scoring DAG consistent with that ordering. The *Ordering Based Search* (OBS) algorithm employs a more systematic search of ordering space (Teyssier and Koller, 2005) and moves through ordering space by swapping adjacent nodes in the ordering. OBS uses an exhaustive approach to select the highest scoring DAG consistent with each ordering. At each iteration, all $n-1$ node swaps are scored, and the ordering that has the highest scoring DAG is adopted for the next iteration. This process is repeated until a local maximum is reached. Note that the swap adjacent operator only changes the possible parents of two nodes and so the score of the new ordering can be computed cheaply. OBS used a simple sound pruning rule based on the observation that if $\boldsymbol{Pa}(X_i)$ and $\boldsymbol{Pa}'(X_i)$ are two possible sets of parents of node $X_i$ and $\boldsymbol{Pa}(X_i) \subset \boldsymbol{Pa}'(X_i)$ and $score(X_i|\boldsymbol{Pa}(X_i)) \geq score(X_i|\boldsymbol{Pa}'(X_i))$, then $\boldsymbol{Pa}'(X_i)$ cannot possibly be the best set of parents of $X_i$ and can be removed (pruned) from further consideration in any ordering.

The pruning rule used in OBS is applicable to any decomposable score, but has the drawback that it is necessary to compute the score for the superset $\boldsymbol{Pa}'(X_i)$ before it can be discounted. Other rules, typically specific to a particular score such as BIC or BDeu, are more powerful in that conditions applying for a set of parents can be used to prune supersets of parents without having to score these supersets. This makes the algorithm more efficient, and

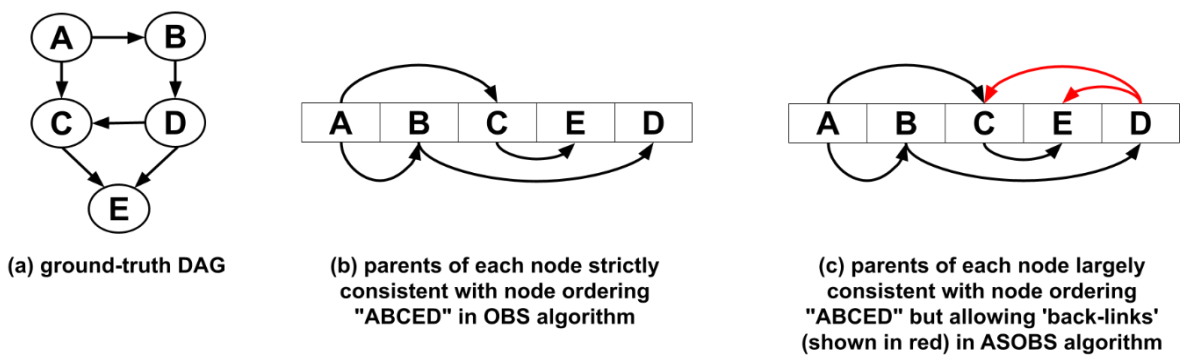

(a) ground-truth DAG

(b) parents of each node strictly consistent with node ordering "ABCED" in OBS algorithm

(c) parents of each node largely consistent with node ordering "ABCED" but allowing 'back-links' (shown in red) in ASOBS algorithm

**Figure 15** – Example illustrating the relaxed node ordering consistency used by the ASOBS algorithm

these scores often remove large portions of the search space. The ASOBS algorithm (Scanagatta et. al., 2015) uses the BIC* score which approximates BIC to accomplish this. Like OBS, ASOBS uses a primary search over node orderings using the swap adjacent operator. However, when generating a DAG from the ordering, ASOBS relaxes the restriction that a node can only have parents earlier in the ordering, whilst continuing to ensure the DAG is acyclic. Figure 15 illustrates a situation that might occur when learning the DAG shown in Figure 15(a). OBS would select parent sets for node ordering "ABCED" that are strictly consistent with the ordering as shown in Figure 15(b). In contrast, ASOBS allows some 'back-links', shown in red in Figure 15(c), where nodes "C" and "E" have parent "D" from lower down the ordering. ASOBS will find a DAG with a score at least as high as OBS would find for any given ordering. The authors used ASOBS to model networks with more than 1,000 nodes and removed the restriction on the number of parents for a node.

The fact that the swap adjacent operator changes few parents makes it relatively inexpensive, but it does mean that there may be many relatively weak local maxima close to one another. The INOBS algorithm introduces an insert operator which causes larger steps in changing the order of a node in the node ordering input to address this (Lee and van Beek, 2017). The authors also investigated a variant of INOBS, called IINOBS, which employs *iterated local search*. This is an extension to hill-climbing that adds a perturbation operator of swapping nodes in the ordering so that when a local maximum is reached, hill-climbing is restarted in a relatively close neighbourhood. The perturbation operator chosen should not be so weak that restarted hill-climbing just finds the same local maximum, and nor should it be so strong that hill-climbing is starting all over again in a completely new region. Iterated local search is based on the intuition that local maxima may occur in clusters. Similarly, Lee and van Beek (2017) combined INOBS with genetic algorithm techniques to create the *memetic* algorithm, MINOBS. Hill-climbing search is performed on an initial population of orderings to get a population of locally maxima orderings. Crossover, mutation and population pruning genetic techniques then operate on this population of locally optimum orderings to produce a new population to perform hill-climbing on.

WINASOBS (Scanagatta et. al., 2017) employs a yet more impactful *window* operator which changes the position of a *group* of nodes in the node ordering, and uses the same relaxation as ASOBS when generating a DAG from the ordering. The authors evaluated WINASOBS against IINOBS and MINOBS when learning from 24 real-world datasets with between 16 and 1556 variables and synthetic networks with up to 10,000 variables. The learnt graphs were evaluated only on the basis of their BIC score. WINASOBS produced higher scoring graphs than the other algorithms except for being on par with MINOBS when execution time was limited to one hour.

Sampling in node-ordering space can be implemented using MCMC approaches. This seeks to counter the slow convergence to a stationary distribution, also known as *slow mixing*, that can be encountered when sampling DAG space, for example, in the $MC^3$ algorithm. Friedman and Koller (2003) sampled ordering space in the *Order-MCMC* algorithm. They derived a closed-form expression for the probability of specific graph features, such as an individual edge, occurring with a particular ordering. They then used MCMC to sample over the space of orderings and hence obtain the overall probability of a particular feature occurring. A DAG can then be constructed using only edges with a probability above a specified threshold in another example of model averaging. Niinimaki et al. (2011) proposed the Partial Order-MCMC method, where nodes are grouped into ordered buckets which demonstrated better mixing than Order-MCMC. A disadvantage of order-based MCMC approaches arises because a given DAG may be consistent with many node orders and different DAGs will be consistent with different numbers of node orders. It is therefore difficult to define priors on node orders which do not bias the posterior distribution of DAGs. This problem is more pronounced with small sample sizes where the priors have more influence.

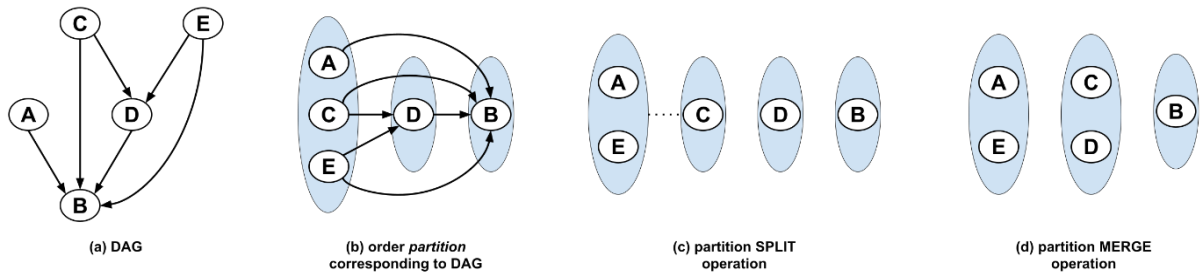

**Figure 16** - Partitions used in Partition-MCMC (adapted from figure in Kuipers and Moffa, 2017). Note that other authors refer to this partitioning of the nodes as a *(topological) partial ordering*.

Kuipers and Moffa (2017) introduce Partition-MCMC to address the bias issue discussed in the previous paragraph. It searches in the space of *partitions*, illustrated as the set of light blue ellipses in Figure 16(b), which is the partial topological ordering of the nodes of the DAG shown in Figure 16(a). The algorithm assigns a score to each partition which is the sum of scores over all DAGs consistent with that partition, thus taking into account the number of DAGs in the partition and therefore removing the bias discussed previously. Partition-MCMC samples the partitions using operators which split or merge an element within the partition as illustrated in Figure 16(c) and (d). They also assess the more impactful reverse operator proposed by Grzegorczyk and Husmeier (2008).

### 4.2.4. *Approximate search of ancestral graph space*

The majority of algorithms that assume the presence of latent variables are constraint-based, some of which were discussed in subsection 3.4. However, Triantafillou and Tsamardinos (2016) proposed the Global Search for Maximal Ancestral Graphs (GSMAG) algorithm which searches in the space of ancestral graphs containing directed or bidirected edges. That is, it allows latent, but not selection variables. It explores the search space using greedy search in which directed and bidirected edges are added, removed, reversed or converted between each other. A BIC score is defined based on the probability distribution factorisation for Gaussian MAGs described in Richardson (2009), which allows the score to be decomposed into a likelihood for each *c-component* in the MAG given the data. C-components are the fragments connected by bidirected edges that result from removing all directed edges in a MAG. Thus, only scores for the c-components affected by any edge change need to be recomputed. The authors evaluated GSMAG against FCI and CFCI constraint-based algorithms on synthetic random Gaussian networks of up to 50 variables, 10% of which are latent variables. They found that GSMAG had better recall (see subsection 6.1 for explanation of recall and precision) than FCI and CFCI (slightly), but worse precision than CFCI. It was also considerably slower than both FCI and CFCI.

### 4.3. *Exact score-based algorithms*

In contrast to the algorithms considered so far, *exact* algorithms guarantee to return the highest scoring DAG. Note that whilst some approximate algorithms can guarantee to return the highest scoring graph as $N \to \infty$ (Chickering, 2002; Chickering and Meek, 2002), exact algorithms guarantee to return the highest scoring graph for the input dataset, however small the sample size might be. They usually make use of sound pruning rules to avoid searching over the whole search space. Whilst these algorithms produce the DAG with the highest score, this may not be the graph which best matches the underlying ground-truth for reasons such as:

- limitations in the data learnt from, particularly sample size and any form of noise;
- biases introduced by the score used, for example, towards sparser or denser graphs.

Note that some authors refer to exact algorithms as global algorithms since they guarantee to return the graph with the globally maximum score. However, in this paper, we apply the term "global" to constraint-based algorithms that consider the global structure, and so we prefer "exact" to "global" here. Exact algorithms typically treat structure learning as a constrained combinatorial optimisation problem which involves determining the optimally scoring combination of parents for each node subject to the constraint that the graph is acyclic.

### 4.3.1. Exact search of node-ordering space

Dynamic programming was an early technique used in exact algorithms (Koivisto and Sood, 2004; Ott et al., 2004) searching in node-ordering space. Dynamic programming is an algorithm paradigm which solves small sub-problems first and uses these results to solve larger problems built on them. Singh and Moore (2005) and Silander and Myllymaki (2006) applied this paradigm to structure learning using the insight that every DAG must have at least one *sink node* (a node with no children, also referred to as a leaf node). So, any DAG with nodes $X$ can be constructed from a sink node $X_{sink}$ and a sub-DAG with nodes $X - \{X_{sink}\}$. The maximum graph score can thus be expressed as a recurrence relation:

$$score_{max}(X) = score_{max}(X - \{X_{sink}\}) + score_{max}(X_{sink}|Pa(X_{sink}))$$

Note that which nodes in the sub-DAG are chosen to be the parents of the sink node, $Pa(X_{sink})$, does not affect the score of the sub-DAG itself, and so this element of the score can be maximised independently. Also, since the end of a node ordering is always a sink node, recursively finding sink nodes represents a way of traversing node ordering space.

Figure 17 based on Singh and Moore (2005) shows how dynamic programming exploits this recurrence relationship. Each box in this *ordering lattice* represents one of the possible sub-DAGs in a small network with nodes {1, 2, 3, 4}. The optimal DAG is determined by a depth first search of this lattice. Starting at the top with the DAG containing all nodes, the search moves down the first blue arrow to sub-DAG {1, 2, 3} with node 4 left behind as the sink node. The highest scoring parents for node 4 in sub-DAG {1, 2, 3} are determined, in this illustration assumed to be {1, 2} with a score of 4. The search continues down the blue arrows, determining the highest scoring parents for each local sink node until we reach the bottom sub-DAG on the blue path, {1}. This blue search path represents node ordering {1, 2, 3, 4}, and the DAG corresponding to the path is shown to the left of the lattice, where the concentric dashed ellipses represent the sub-DAGs encountered, and arcs showing the highest-scoring parents determined for each sink node. Having reached the bottom, the search backtracks to sub-DAG {1, 2} to then score the purple path. At this point, all search paths below {1, 2} have been followed, so a maximum score can be assigned to sub-DAG {1, 2} and the paths below {1, 2} never need to be revisited, illustrating the 'self-pruning' nature of this approach. The search is guaranteed to find the highest scoring DAG, illustrated in this example by the red search path.

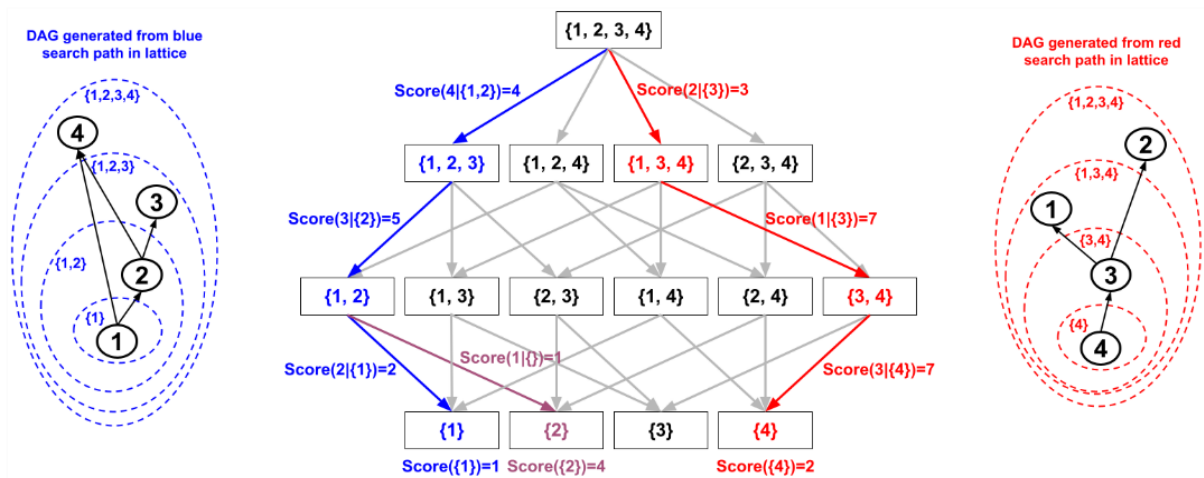

**Figure 17** - Lattice of all sub-DAGs for a hypothetical four node network based on Singh and Moore (2005), illustrating the dynamic programming approach using sink nodes.

Yuan and Malone (2011) noted that dynamic programming is inefficient because it needed to fully evaluate the exponential solution space and that, by ignoring acyclic constraints, Branch and Bound (covered below in subsection 4.3.2) made the search space unnecessarily large. They proposed that navigating the lattice in Figure 17 be performed by the general-purpose graph algorithm A* which finds optimal weighted paths. As with the dynamic programming technique explained above, the algorithm maintains a current best score down to the current sub-DAG reached in the lattice, but it also estimates the best score obtainable down the currently unexplored paths below the sub-DAG. At any one time, A* explores the paths which have the highest estimated score from top to bottom, a so-called 'best-first' approach. As long as the estimated score is *admissible* (i.e., never underestimates the score obtainable), A* guarantees to find the highest-scoring path and the hence highest-scoring DAG. Their approach proved to be several times faster than dynamic programming and much faster than Branch & Bound which we cover later in subsection 4.3.2.

Another recent exact score-based algorithm called *CPBayes* (van Beek and Hoffman, 2015) adopts the *constraint programming* paradigm. In this paradigm, constraints are defined across the variables and a domain of possible values maintained for each variable. Note that these are not conditional independence constraints, and so this approach should not be confused with the constraint-based structure learning described in section 2. As the algorithm explores possible solutions by changing one variable, the constraints mean that the possible values of other variables are altered; a process known as *constraint propagation*. A simple application of constraint programming to the Sudoko game may serve to illustrate the concept. The domains for each square are the numbers 1 to 9, but as a number is chosen for one square, this reduces the domains of possible values for other squares, according to the constraints of the game that each number can only occur in one row, column or 3x3 block.

For application to BN structure learning, CPBayes defines three classes of variables describing: the node ordering; the parent set of each node; and depth, defined as the longest path of any source vertex to the node. Several types of constraints are defined, and the ones that relate the three classes of variables include acyclicity constraints, *symmetry breaking* constraints which avoid redundant solutions which belong to the same equivalence class or node ordering, for example, and *dominance* constraints which apply cost-based pruning. CPBayes is a depth-first branch-and-bound algorithm that explores the node-ordering space by swapping nodes in the order, with the constraints and score bounds used to reduce the search space. The authors argued that the inclusion of the depth variable together with the extensive set of constraints reduces the search space compared to other exact algorithms, though without quantifying this. Results obtained at the time showed that runtimes were comparable to the Integer Linear Programming approach in GOBNILP (refer to subsection 4.3.2).

Troser et al. (2021) introduce the ELSA algorithm which enhances CPBayes by including linear programming techniques similar to GOBNILP to provide more efficient acyclicity checking. It uses a more specialised greedy, and therefore efficient, algorithm than GOBNILP to solve the linear programming problem. ELSA is able to find the optimal graph within a time limit of 10 hours (for datasets with between 61 and 111 variables) in considerably more cases than either GOBNILP and CPBayes. Note that this time limit applies to the graph search only; it does not include the pre-computation of parent set scores.

Tan et. al. (2022) propose two variable partitioning approaches which they demonstrate can improve learning times in algorithms such as A*, often by orders of magnitude. The first heuristic is *ancestral partitioning* which assumes a partial ordering as illustrated in Figure 16 to greatly prune the number of nodes in the order graph. Note that, with this heuristic, the structure learning algorithm can only guarantee to return the optimal scoring graph if the partial ordering that is assumed is consistent with the ordering of the true graph. The second heuristic, *heuristic partitioning,* splits the variables into partitions in such a way that more paths in the order graph can be ruled out during the search phase. This heuristic does not invalidate the guarantee of returning the highest scoring graph.

### 4.3.2.  Exact search of DAG space

The Branch & Bound algorithm uses another recursive approach but starting from a larger problem which it decomposes (de Campos et al., 2009). The algorithm first creates a cache of the highest scoring parents for each node. As the algorithm proceeds, it maintains a queue of candidate graphs ordered by score, and a record of the highest scoring DAG found so far. Initially, the queue is populated with the graph where each variable is assigned its optimally scoring parents without regard to acyclicity. The algorithm proceeds by considering the top scoring graph on the queue. If it is a DAG, it checks its score and updates the best scoring DAG if needed. If the graph contains a cycle, it breaks the cycle at each arc in the cycle creating several sub-graphs which it places back on the queue. The algorithm continues until the queue is empty by which time the globally optimal DAG will have been identified.

Most exact search algorithms, including Branch & Bound, maintain a cache of possible parent sets for each node together with their associated score. If done for every parent set this cache would have $2^{n(n-1)}$ entries which quickly becomes prohibitive. Research into pruning (Cussens, 2012; Suzuki, 2017; de Campos et al., 2018, Correia et al., 2020) which can reduce the size of this cache whilst maintaining the guarantee of optimality has been important to the development of exact search. Most of the later pruning approaches are specific to a particular score (e.g. BIC or BDeu) and can have a dramatic effect on the cache size and search space considered. De Campos and Ji (2010) reported cache size being reduced by between approximately $10^2$ and $10^7$ times depending upon the network and specific score considered. In the most dramatic cases, this resulted in a reduction of the search space by hundreds of orders of magnitude. Guo and Constantinou (2020) show that pruning candidate parent sets simply by removing those with low scores can offer considerable runtime saving, particularly for larger networks. This type of pruning means that previously exact algorithms no longer guarantee to return the graph with the highest possible score, that is, they become approximate algorithms. Nonetheless, this often does not reduce the accuracy of the learnt graph by very much.

Integer Linear Programming (ILP) approaches treat structure learning as a constrained integer programming problem (Jaakola et al., 2010; Bartlett and Cussens, 2017). Figure 18 illustrates some of the concepts of integer linear programming with reference to a simple optimisation problem. Suppose we wish to maximise $y$ where the following constraints apply: $3y \leq 2x + 6$, $3y \leq 15 - 5x$, $x \geq 0.5$, and $y \geq 0.5$. This is a *linear programming (LP)* problem, and the shaded space in the figure shows the feasible solutions which meet the constraints. With three variables the feasible solutions would be bounded by a polyhedron, and with $n$ variables they are bounded by an *n-polytope*. The LP solutions are always at vertices of the polytope, in this case at $x = 9/7$, $y = 20/7$. However, if we restrict the solutions to integer values then the problem becomes an *integer linear programming (ILP)* problem. The feasible integer solutions are shown as the black circles in the figure.

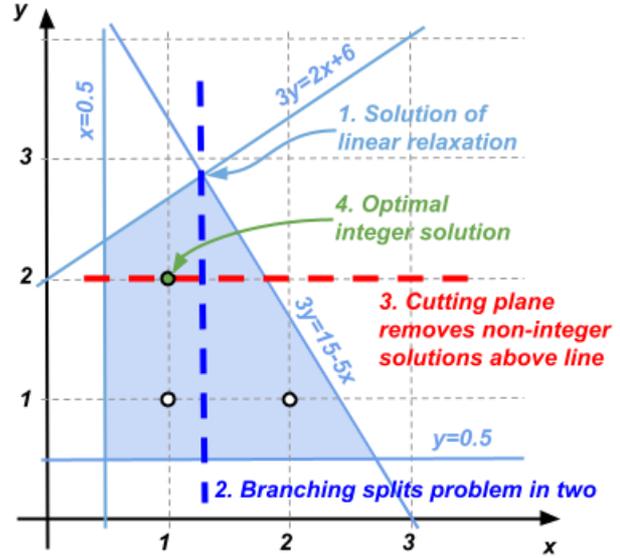

The ILP problem is tackled by first using the well-established *simplex* algorithm (Dantzig, 2016) to solve the LP problem, which is referred to as solving the *linear relaxation* of the ILP program. The problem is then split into two by looking at the two solution spaces either side of one of the non-integer solution values; for example, either side of the dashed blue line in the figure. This branching is repeated, forming a *branch-and-bound* search of the solution space. The search can be made more efficient by including extra constraints at each step so that the search becomes *branch-and-cut*. These extra constraints cut off part of the polytope between the LP solution and the integer solutions and are therefore known as *cutting planes*, illustrated by the dashed red line in the figure.

**Figure 18** - Simple Integer Linear Programming Example

The GOBNILP algorithm (Cussens, 2011) represents a DAG by binary integer variables, called *family variables* denoted $I(\boldsymbol{Pa}(X) \to X)$, each family variable representing a possible parent set $\boldsymbol{Pa}(X)$ for node $X$ where $\boldsymbol{Pa}(X) \subseteq \boldsymbol{X} \backslash \{X\}$. $I(\boldsymbol{Pa}(X) \to X) = 1$ for a particular value of $\boldsymbol{Pa}(X)$ indicates that $X$ has that set of parents in the DAG. Learning the optimal DAG can be cast as an ILP problem in the family variable space where we wish to maximise the DAG score, given by:

$$\sum_{X, \boldsymbol{Pa}(X)} score(X | \boldsymbol{Pa}(X)) \cdot I(\boldsymbol{Pa}(X) \to X)$$

*Convexity constraints* are imposed to ensure that each variable can only have one of the possible parent sets:

$$\forall X: \sum_{\boldsymbol{Pa}(X)} I(\boldsymbol{Pa}(X) \to X) = 1,$$

and *cluster constraints* to enforce acyclicity, where any subset $\boldsymbol{X}'$ of all the nodes $\boldsymbol{X}$ must contain at least one node that has no parent in the subset:

$$\forall \boldsymbol{X}': \sum_{X \in \boldsymbol{X}'} \sum_{\boldsymbol{Pa}(X): \, \boldsymbol{Pa}(X) \, \cap \, \boldsymbol{X}' = \emptyset} I(\boldsymbol{Pa}(X) \to X) \geq 1$$

GOBNILP employs an off-the-shelf optimisation program such as SCIP (Berthold, 2012) to solve this ILP problem. GOBNILP uses whether a particular family variable is 0 or 1, that is, whether a node has a particular parent set, as its branching strategy. There are a huge number of cluster constraints possible for reasonably sized networks, and so these are only applied where necessary as the search progresses. Even so, it is usually necessary to place a limit on the maximum number of parents for any node in most applications.

The exact algorithms discussed in these subsections also naturally support BMA where the probability of specific features such as arcs, ancestral relations and Markov Blankets can be calculated by summing their posterior probability over all graph structures. This capability is explored by Koivisto and Sood (2004), Tian and He (2009) and Pensar et al. (2020). The latter two search over DAG structures and so avoid the bias that can arise because different numbers of DAGs are associated with each node-ordering, for example. Recently, Liao et al. (2019) proposed an adaption of the GOBNILP algorithm called GOBNILP-DEV which rather than returning a single highest scoring DAG, returns all the DAGs which have $score(G)$ meeting the condition that:

$$(1 - \varepsilon) \cdot score(G_{OPT}) \leq score(G) \leq score(G_{OPT})$$

where $G_{OPT}$ is the DAG with the highest possible score; i.e., all DAGs within a fraction $\varepsilon$ of optimal. This provides a principled algorithm for obtaining a set of plausible graphs.

### 4.3.3. Exact search of equivalence class space

Chen et al. (2016a) proposed an exact algorithm for searching equivalence class space by defining an Equivalence Class Tree (EC Tree) where each node represents a CPDAG, and which has a unique path to each CPDAG. The algorithm uses A* search to explore the EC Tree efficiently. Chen et al. (2016a) compared this algorithm with an earlier dynamic programming based approach proposed by Chen and Tian (2014) aimed to find the k-best equivalence classes. They found the EC Tree search to be always faster than the dynamic programming approach, and occasionally orders of magnitude faster. Interestingly, they also found that the highest scoring equivalence classes represented very different numbers of DAGs in the eleven networks studied which had between 14 and 23 variables. Whilst some networks had values around the 3.7 DAGs per CPDAG (Gillispie and Perlman, 2002) often quoted in the literature, others had hundreds or even thousands of DAGS in the highest scoring equivalence classes.

## 5. HYBRID LEARNING AND OTHER APPROACHES

Hybrid algorithms combine constraint-based and score-based approaches in an attempt to offer the best characteristics of each. Perhaps the most common way of combining the approaches is to use a constraint-based approach to restrict the search space in which a subsequent score-based approach finds a graph with a local or globally maximum score. We refer to these as Restrict/Maximise hybrid algorithms and discuss them in subsection 5.1. A diverse set of other hybrid approaches is described in subsection 5.2. We then further group the algorithms according to which space they search in, to provide some commonality with the score-based section of this paper. Figure 6 shows the evolution of hybrid algorithms which are shown in yellow colours, and in particular how developments in score and constraint-based algorithms have informed that evolution. Table 4 presents the key characteristics of the hybrid algorithms reviewed here.

### 5.1. Restrict/Maximise algorithms

### 5.1.1.  Restrict/Maximise in DAG Space

Early hybrid algorithms tended to have a constraint (restrict) and score-based (maximise) step in each iteration. For example, each iteration of the Constraint-Bayesian (CB) algorithm (Singh and Valtorta, 1993) had a restrict step in which the PC algorithm learnt a CPDAG, but only using conditioning sets up to a specified size for that iteration. The maximise step then firstly orientates undirected edges so that the product of the K2 score associated with the two endpoints is maximised, producing a DAG. The maximise step then uses the score-based K2 algorithm to construct the optimum-scoring DAG consistent with that DAG's ordering. The restrict and maximise steps are then repeated at increasing conditioning set size until the resulting DAG's score no longer improves.

Similarly, each iteration of the Sparse Candidate (SC) algorithm (Friedman et al., 1999) had a restrict step which used Mutual Information Independence tests to determine candidate parents for each node, followed by a maximise step using the Tabu algorithm constrained by those parent sets. The parent sets of the DAG produced in one iteration were always included as candidate parents for the next iteration, ensuring that each iteration would find a DAG with at least as high a score as the previous iteration.

**Table 4** - Characteristics of hybrid algorithms reviewed, ordered chronologically.

| Algorithm Abbreviation | Algorithm Name or Description | Reference | Algorithm Group | Search space | Search space traversal | Causal sufficiency assumed | Type of output |
|---|---|---|---|---|---|---|---|
| CB | Constraint Bayesian | Singh and Valtorta, 1993 | Restrict/Maximise | DAG | Add/remove/reverse arc | yes | DAG |
| SC | Sparse Candidate | Friedman et al., 1999 | Restrict/Maximise | DAG | Add/remove/reverse arc | yes | DAG |
| HEA | Hybrid Evolutionary Algorithm | Wong and Leung, 2004 | Restrict/Mamimise | DAG | Merge, mutate, select DAG | yes | DAG |
| MMHC | Max-min Parents and Children and Hill Climbing | Tsamardinos, 2006 | Restrict/Maximise | DAG | Add/remove/reverse arc | yes | DAG |
| COS | Constrained Optimal Search | Perrier et al., 2008 | Restrict/Maximise | DAG | Add/remove sink node | yes | DAG |
| PCB | Partial Correlation Bayesian | Yang et al. 2011 | Restrict/Maximise | DAG | Add/remove/reverse arc | yes | DAG |
| BCCD | Bayesian Constraint-based Causal Discovery | Claassen and Heskes, 2012 | Other | DAG | n/a | no | PAG |
| H2PC | Hybrid HPC (HPC is Hybrid Parents and Children) | Gasse et al., 2014 | Restrict/Maximise | DAG | Add/remove/reverse arc | yes | DAG |
| ASP | Weighted CI constraints optimised with Answer Set Programming | Hyttinen et al., 2014 | Other | CI constraints | n/a | no | Cyclic mixed graph |
| COmbINE | Causal Discovery from Overlapping Interventions | Triantafillou and Tsarmadinos, 2015 | Other | CI constraints | n/a | no | PAG |
| GFCI | Greedy Fast Causal Inference | Ogarrio et al., 2016 | Other | Equivalence class | Insert/Delete operations (see subsection 4.2.24.2.2) | no | PAG |
| GSP | Greedy Sparsest Permutation | Solus et al., 2017 | Other | Node ordering | Reverse covered arc in associated DAG | yes | DAG |
| ARGES | Adaptively Restricted Greedy Equivalence Search | Nandy et al., 2018 | Restrict/Maximise | Equivalence class | Add/remove edge + orient edges | yes | CPDAG |
| M³HC | MAG Max-Min Parents and Children and Hill Climbing | Tsirlis et al., 2018 | Restrict/Maximise | MAG | Add/remove+ orient directed/undirected edges | no | MAG |
| GSPo | Greedy Sparsest Poset | Bernstein et al., 2020 | Other | Node Ordering | Directed/bidirected switch in associated DMAG | no | MAG |
| SaiyanH | Saiyan Hybrid | Constantinou, 2020 | Other | DAG | Add/remove/reverse arc | yes | DAG |
| RFCI-BSC | RFCI with Bayesian Scoring of Constraints | Jabbari et al., 2017 | Other | MAG | n/a | no | PAG |
| CCHM | Conservative rule and Causal effect Hill-climbing for MAG | Chobtham and Constantinou, A 2020 | Restrict/Maximise | MAG | Add/remove + orient directed | no | MAG |
| PC + MCMC | PC restricts search space used by Order or Partition MCMC | Kuipers et al., 2020 | Restrict/Maximise | DAG | Swap adjacent nodes or split/merge partitions | yes | DAG |
| mFGS-BS | Majority rule with FGS and Bayesian Scoring | Chobtham et al., 2022 | Other | Weighted arcs | n/a | no | PAG |

Wong and Leung (2004) also interleaved score and constraint-based operations, although in the context of an evolutionary algorithm. Their Hybrid Evolutionary Algorithm (HEA) begins by using low-order CI tests to evaluate the possible parents of each node, maintaining a record of the p-value indicating the likelihood of each CI relationship, and then creates a population of random DAGs. At each iteration, new DAGs are generated through genetic mutation and merge operations (see subsection 4.2.1), and then the population reduced back to its original size by removing the lowest scoring DAGs. Each individual DAG is assigned its own, dynamic, conditional independence significance level which is used in conjunction with the CI test p-values to restrict parent sets in individual creation, mutation and merge operations. This individualised significance level helps maintain population diversity.

However, most Restrict/Maximise approaches do not interleave restrict and maximise operations. Rather, they use a constraint-based algorithm to define a restricted search space, and then a score-based algorithm operates within that restricted space. Max-Min Hill Climbing (MMHC) proposed by Tsamardinos (2006) is a widely-used example of this. In the restrict phase, MMHC uses the MMPC local constraint-based algorithm (see subsection 3.3.2) to construct the graph skeleton. The subsequent maximise phase uses Tabu hill-climbing (see subsection 4.2.1) to learn the output DAG, but is constrained to only use edges in the graph skeleton produced in the restrict phase. The author conducted a detailed evaluation of MMHC against leading constraint and score algorithms at the time with MMHC producing more accurate graphs than GES, PC and TPDA, and demonstrating the ability to learn networks with 1,000 variables. Gasse et al. (2014) proposed Hybrid HPC (H2PC) which uses a Hybrid Parents and Children (HPC) algorithm to create a skeleton in the restrict phase with a focus on avoiding false missing edges. HPC uses an ensemble of weak parent-and-children algorithms to achieve this, and the maximise phase of H2PC uses Tabu hill-climbing. H2PC produced graphs with better structural accuracy and data fitting than MMHC across 10 networks with up to 1,836 variables. However, H2PC was considerably slower than MMHC, being around 10 times slower at large sample sizes.

Whilst most hybrid algorithms use an approximate score-based approach, Perrier et al. (2008) proposed Constrained Optimal Search (COS) which used an exact score-based dynamic programming approach (see subsection 4.3.1) in the maximise phase operating within a reduced search space defined by the skeleton returned by the restrict phase. Like MMHC, COS uses the MMPC algorithm to generate the skeleton, but uses a deliberatively high CI significance level so that the skeleton is denser than usual to increase the chances of the restricted search space including the globally optimal graph. COS was compared with MMHC and produced more accurate structures and data fitting scores, although comparisons were limited to $n = 20$ due to runtime constraints of dynamic programming at that time. It is noteworthy that the authors found that using the output from either MMHC or COS as the initial graph for a further score-based hill-climbing phase improved graph quality for both algorithms.

The papers introducing the hybrid algorithms discussed so far focussed on discrete variables, but Yang at al. (2011) described the Partial Correlation Bayesian (PCB) algorithm which considers continuous variables. The restrict phase uses partial correlation CI tests to determine the graph skeleton, followed by a hill-climbing maximise phase. The authors showed that their approach is applicable whenever a continuous variable is a linear function of its parents, not just the usual special case when the variables follow a Gaussian distribution. PCB produced more accurate graphs than Sparse Candidate, MMHC, PC and TPDA.

### 5.1.2. *Restrict/Maximise in equivalence class space*

Nandy et al. (2018) proposed Adaptively Restricted GES (ARGES) for continuous variables which uses MMPC for the restrict step and then a modified GES (see subsection 4.2.2) for the maximise step. As is usual in restrict/maximise algorithms, they generally restrict GES to the skeleton produced in the restrict phase. However, they relax this restriction slightly to allow it to add shielding edges on v-structures whilst the CPDAG is being learnt. These extra shielding edges automatically disappear as the algorithm progresses, and so do not appear in the final CPDAG, hence why the algorithm is known as "adaptively restricted". The authors provide theoretical arguments showing that temporarily allowing these extra edges means that the ARGES algorithm can offer both classical and high-dimensional consistency.

### 5.1.3. *Restrict/Sampling in ordering space*

Kuipers et al. (2022) describe a hybrid MCMC algorithm which creates a restricted search space using the PC constraint-based algorithm (see subsection 3.2.2) followed by MCMC sampling in the node ordering space (see the Order-MCMC algorithm in subsection 4.2.3), or sampling in the partition space (see the Partition-MCMC algorithm in subsection 4.2.3). The sampling is done in a space that initially corresponds to that identified by the PC algorithm, but each node is allowed an additional parent outside the initial space, so that the restriction is relaxed. The authors claim a large improvement in efficiency with the complexity to find the Maximum A-Posteriori (MAP) DAG reduced from $n^k$ to $2^k$ where $n$ is the number of variables, and $k$ the maximum in-degree, making the algorithm suitable for high-dimensional problems. The authors find that accuracy is improved by allowing the sampling phase to add another parent node to the parent sets found during the constraint phase. Using PC in combination with Order-MCMC, the authors demonstrate considerably better SHD scores than GES or PC on random graphs with $n = \{20, 80, 140, 200\}$ and sample size $N = \{2n, 10n\}$.

Viinikka et. al. (2020) build upon the approach described in Kuipers and Moffa (2017) and Kuipers et. al. (2022) but reduce time and memory requirements to support much higher maximum in-degrees. They also improve the selection of candidate parent sets which they formulate as an optimisation problem which they solve exactly for smaller networks and heuristically for larger networks. The authors evaluated accuracy in predicting pairwise ancestral relationships and found improved accuracy over Partition-MCMC.

### 5.1.4. *Restrict/Maximise in ancestral graph space*

The MAG Max-Min Hill Climbing (M³HC) algorithm proposed by Tsirlis et al. (2018) uses MMPC in its restrict phase, and GSMAG (see subsection 4.2.4) in the maximise phase so that causal sufficiency is not assumed. MMPC produces a superset of the true adjacencies in the presence of latent variables and so is a suitable candidate for the restrict phase in causally insufficient settings. The empirical results of M³HC outperform GSMAG, FCI, CFCI and GFCI on standard BNs with up to 1041 variables (Tsirlis et al., 2018).

CCHM (Chobtham and Constantinou, 2020) follows the approach used by the CFCI constraint-based algorithm (see subsection 3.4.4) to learn the skeleton and classify unshielded triples of nodes as either definitely a v-structure, definitely not a v-structure, or an ambiguous triple. It then uses a greedy hill-climbing search to further orientate edges in the MAG to maximise the BIC score for MAGs which is employed by M³HC and GSMAG. Since this score is score equivalent, some edges remain un-orientated, and so CCHM applies Pearl's do-calculus to orientate the remaining edges. Chobtham and Constantinou (2020) found that CCHM is generally more accurate than M³HC, FCI, CFCI and GFCI on both random and well-known BNs. Both M³HC and CCHM were generally slower than FCI and GFCI while M³HC was faster than GSMAG, and CCHM is faster than CFCI. Note that GSMAG, M³HC and CCHM currently assume linear GBNs and so the data must be continuous values.

### 5.1.5. Symmetry correction

As noted in subsection 3.3, skeletons produced by local discovery constraint algorithms, such as MMPC, are subject to errors arising from asymmetries in local structures where one node is in another's local structure but not vice versa, making it unclear whether there should be an edge between the two nodes. In this situation, Zhao and Ho (2019) proposed Symmetry Correction which involves relearning the combined local structure of the two nodes using a score-based algorithm, and deciding whether to include the questionable edge dependent upon whether the local score-based search generates a graph including that edge. Incorporating Symmetry Correction, itself a hybrid approach, can improve the results of both constraint-based local discovery algorithms, and hybrid algorithms which use them. The paper evaluated the technique across 16 BNs with the number of variables up to $n = 724$, discovering that local structure asymmetries were commonly produced by MMPC, SI-HITON-PC, IAMB and GS algorithms. In the majority of cases, Symmetry Correction produced structurally more accurate and better fitting graphs, particularly when used in the restrict phase of a hybrid algorithm.

### 5.2. Other hybrid approaches

### 5.2.1. Other hybrid approaches which search in DAG space

Structure learning algorithms often produce graphs with several isolated components with no edges between them. Constantinou (2020) argued that this is undesirable in real-world settings since it prevents the propagation of evidence between the variables in the different components if the learnt graph is subsequently parameterised and used for inference. The author therefore proposed the SaiyanH hybrid algorithm which guarantees to produce a DAG with a connected skeleton not containing independent components. SaiyanH begins by creating a connected undirected graph containing edges between pairs of nodes which have the strongest relationships between them according to a novel associational score. A second phase orientates all the edges in three steps: firstly, using a sequence of CI tests; secondly, using a score-based orientation heuristic; and thirdly, seeing which orientation maximises the effect of interventions. The resultant DAG is used as the initial graph for a third Tabu hill-climbing phase which is constrained to not violate independence relationships discovered in the first two phases, nor generate isolated components. Hence, the final DAG skeleton is guaranteed to be connected. SaiyanH ranked 4[th] when evaluated for structural accuracy against 12 other leading constraint, score and hybrid algorithms when learning six BNs with up to $n = 109$ variables, whilst always achieving its aim of producing a connected skeleton.

### 5.2.2. Other hybrid approaches which search in node-ordering space

The Greedy Sparsest Permutation (GSP) algorithm (Solus et al., 2017) is an approximate algorithm building upon the exact Sparsest Permutation (SP) algorithm (Raskutti and Uhler, 2013, 2018) which was only viable up to 10 variables. Both are hybrid algorithms which search node ordering (which the authors referred to as a permutation) space and use a constraint-based approach within the overall algorithm to generate a minimal I-MAP DAG associated with each node ordering encountered. This constraint-based approach generates a DAG, denoted $G_{\prec}$, associated with node ordering $\prec = (\prec_1, \prec_2, \dots, \prec_i, \dots, \prec_n)$, using the following rule to generate the edges in $G_{\prec}$:

$$j < k \text{ and } X_{\prec_j} \not\perp X_{\prec_k} \mid \left\{ X_{\prec_1}, \dots, X_{\prec_{k-1}} \right\} \setminus \left\{ X_{\prec_j} \right\} \iff arc\ X_{\prec_j} \to X_{\prec_k}\ in\ G_{\prec}$$

where $X_{\prec_i}$ is the node at position $i$ in the node ordering. That is to say, it generates an arc $A \to B$ in the associated DAG if $B$ is lower down the ordering than $A$, and if $A$ is dependent on $B$ conditional on any subset of the nodes higher up the ordering than $B$. Pearl (1998) showed that this creates a minimal I-MAP DAG for that node ordering.

Solus et al. (2017) proposed the Triangle Sparsest Permutation (TSP) algorithm which traverses ordering space in a depth-first search, by flipping a covered arc in the associated DAG and then moving to the new ordering associated with that DAG. A covered arc is one where the two endpoint nodes have the same set of parents (ignoring the endpoint node that is the parent of the other endpoint). Solus et al. (2017) demonstrated that TSP is asymptotically consistent, the first ordering space algorithm to offer this guarantee. GSP limits the depth of the search but restarts the search, and so offers shorter runtimes than TSP. Comparisons showed GSP was on par with GES and PC in high sample size settings ($n = 10, N = 10,000$) and produced more accurate graphs in low sample size settings ($n = 100, N = 300$).

Bernstein et al. (2020) applied a similar approach to GSP but their Greedy Sparsest Poset (GSPo) algorithm targets causally insufficient problems. A *poset* is a partial node ordering associated with a directed MAG (DMAG), a MAG that has only directed and bidirected but no undirected edges, and so supports latent variables but not selection variables. Analogously to GSP, they provided a mapping from a poset to a minimal I-MAP DMAG and traverse poset space by making changes in the associated DMAG, moving to the poset associated with that new DMAG. In this case, the allowed change in the DMAG is changing a single bidirected edge into a directed one, or vice versa. Each of these moves in poset space results in a DMAG with the same or fewer edges. The authors conjectured, and supported with empirical evidence, that this algorithm produces a DMAG that is Markov equivalent to the true graph as $N \to \infty$. Results on synthetic Gaussian networks with between 10 and 50 variables, three of which are latent, showed better structural accuracy than FCI and FCI+. Runtime is sensitive to the initial poset provided to GSPo, although provided a good starting poset is used (for instance using GSP to produce an initial DAG), GSPo is faster than FCI in all cases and slower than FCI+ when there are more than $30-40$ variables.

### 5.2.3. Other hybrid approaches which search in equivalence class space

Ogarrio et al. (2016) also aimed to produce a hybrid algorithm that provides asymptotic guarantees of correctness in the presence of latent variables. They observed that constraint-based algorithms for causally insufficient settings such as FCI, RFCI and FCI+ are asymptotically correct, but low sample size performance is poorer, especially returning graphs with too many bidirected edges. The score-based GES and FGES are asymptotically correct in causally sufficient situations, although they produce extra adjacencies and incorrect orientations when there are latent variables present. Thus, they proposed the Greedy Fast Causal Inference (GFCI) algorithm which first uses GES to produce a CPDAG. GFCI then employs CI tests to remove extraneous adjacencies in this skeleton, followed by modified FCI orientation rules to produce a PAG. The authors evaluated GFCI on synthetic Gaussian BNs with number of variables, $n = 100, 1,000$ with either 5% or 20% of those being latent variables. GFCI was shown to generally have better recall and precision on adjacencies and arrow end marks than FCI, RFCI and FCI+, and in the cases where it was worse, it was only slightly worse. GFCI has worse recall of bidirected edges but much better precision of them, supporting the theoretical argument advanced that FCI, RFCI and FCI+ tend to produce too many adjacencies. GFCI was around 23% slower than RFCI and faster than FCI+ (comparisons were not provided for FCI).

### 5.2.4. Other hybrid approaches which search in ancestral graph space

The Bayesian Constraint-based Causal Discovery (BCCD) algorithm (Claassen and Heskes, 2012) is a hybrid algorithm which does not assume causal sufficiency and produces a PAG. It assigns a Bayesian score to CI constraints to reflect the reliability of each constraint, rather than the binary true/false judgement made by most constraint-based approaches. The score is used to rank CI constraints, helping to prevent unreliable decisions being propagated, and providing a principled means to resolve orientation conflicts.

BCCD follows the approach of PC and FCI and starts with a complete undirected graph and then uses conditioning sets of increasing size in the adjacency phase. The probability of CI constraints is incremented during this adjacency phase and if the probability exceeds a threshold the relevant edge is removed from the evolving skeleton. The ranked CI constraints are used to orientate unshielded triples and then further orientation rules applied using the CI constraint ranking to resolve conflicts. Claassen and Heskes (2012) evaluate BCCD on small random graphs with six or twelve variables and find that it is slightly more accurate than FCI and CFCI.

Hyttinen et. al. (2014) use a similar approach that associates a cost with each independence and dependence constraint. However, the algorithm is targeted at learning a causal model from multiple datasets with different but overlapping sets of variables, with interventions on different sets of variables, in a causally insufficient setting. The problem is tackled as a constrained optimisation problem which is solved using an off-the-shelf Answer Set Programming approach. Performance is evaluated on small randomly generated graphs with six continuous variables and demonstrated better accuracy than PC and CPC for causally sufficient tests, and FCI and CFCI for causally insufficient tests.

The COmbINE algorithm (Triantafillou and Tsamardinos, 2015) also assigns probabilities to CI statements and is applied to the same causally insufficient setting with multiple interventional datasets with overlapping variables. COmbINE uses FCI to learn the PAG for each interventional dataset. It merges these using an open-source Boolean satisifiability application, MINISAT (Sorensson and Een, 2005), to produce a summary graph showing the edges and orientations that are invariant across the individual PAGs.

Jabarri et al. (2017) proposed a hybrid variant of the RFCI constraint-based algorithm which supports latent and selection variables, known as RFCI-BSC. It assigns a novel Bayesian Scoring of Constraints (BSC) score to each CI test which reflects the probability that the variables are indeed conditionally independent. The RFCI algorithm is then modified so that it stochastically decides whether each CI is true or not according to its BSC score. The algorithm is not deterministic because it uses a different random seed each time it runs.

The overall RFCI-BSC algorithm repeats this stochastic learning process to produce a series of PAGs. It then re-uses the BSC score concept to generate an overall BSC score for each PAG based on the BSC scores for the CI relationships used to generate that PAG. This BSC score is reflective of the posterior probability of that PAG. Finally, model averaging produces a single, non-deterministic, output PAG by considering the probability of each edge across all the PAGs. Jabarri et al. (2017) found that the structural accuracy of RFCI-BSC was generally better than RFCI, with up to $n = 70$ variables and relatively small sample sizes of $N = \{200, 2000\}$, though it had worse adjacency accuracy with the largest number of variables.

Chobtham et al. (2022) describe the mFGS-BS hybrid algorithm which learns a PAG from one observational dataset and one or more interventional datasets. The datasets must all have the same variables, but causal sufficiency is not assumed. FGES is used to learn a CPDAG from each dataset, and probabilities are then assigned to individual arcs based on their frequency in the different CPDAGs, a majority-voting constraint-based approach to arc orientation in unshielded triples, and from do-calculus considerations around intervened variables. A final arc orientation phase removes cycles and resolves any orientation conflicts to produce a single PAG. The authors find that mFGS-BS outperforms COmbINE, as well as GFCI and RFCI-BSC which are baselines that do not account for interventions.

### 5.3. Other Structure Learning Approaches

We discuss some algorithms in this sub-section which take a different approach to the algorithms already discussed.

#### 5.3.1. Functional Causal Models

Functional Causal Models (FCMs) can be used to model causal systems where it is assumed that a variable's value can be expressed as some deterministic function of its parents plus a noise term that is independent of all the causes, that is:

$$X = f(\boldsymbol{Pa}(X), \varepsilon)$$

where $f$ is a deterministic function, and $\varepsilon$ is a noise term which is independent of $\boldsymbol{Pa}(X)$ and all the other variables' noise terms. This approach is mostly used with continuous variables, though it can be used to model, for example, noisy-OR relationships between discrete variables. Given some assumptions about the form of $f$ and $\varepsilon$, it is possible to learn the complete causal structure from observational data alone, including the case of identifying the arc orientation with just two variables.

One set of assumptions is the Linear Non-Gaussian Acyclic Model (LiNGAM) (Shimizu et al., 2006) which assumes $f$ is linear, $\varepsilon$ is independent but non-Gaussian noise, and that the system is causally sufficient. With these assumptions, the authors show that structure learning can be undertaken by performing independent component analysis (ICA) on the data using well-established approaches (Hyvärinen and Oja, 2000). Hoyer et al. (2008b) extend LiNGAM to support latent variables.

Zhang and Hyvärinen (2009) propose a more general type of FCM called the Post Non-Linear (PNL) Causal Model which has the form:

$$X = f_{mes}(f_{nl}(\boldsymbol{Pa}(X)) + \varepsilon)$$

where $f_{nl}$ defines the variable's value as a non-linear function of its causes, $f_{mes}$ is a non-linear function that can represent measurement error, and $\varepsilon$ is the independent noise term. This more general formulation of FCMs incorporates LiNGAM and Additive Noise Models (Hoyer et al., 2008a, Gretton et al., 2009) as special cases. Zhang and Hyvärinen (2009) show that arc orientations are identifiable in the PNL model with five exceptions. The most important exception is when $f_{mes}$ is the identity function, $f_{nl}$ is linear, and $\varepsilon$ is Gaussian. This is the linear Gaussian setting which is the usual assumption made by algorithms in the rest of this paper when modelling systems with continuous variables, and hence why, in that case, it is only possible to learn up to an equivalence class from observational data. We have only presented a brief overview of FCMs here and would encourage interested readers to read recent reviews of the area such as the one by Glymour et al. (2019).

#### 5.3.2. Continuous Optimisation

Another recent development in structure learning is the continuous optimisation approach. In the *combinatoric* approaches discussed so far, a DAG can be represented as an *adjacency matrix* where a 1 at a particular position $(i, j)$ indicates the presence of an arc from node $X_i$ to node $X_j$, whereas a 0 would indicate the absence of an arc. In contrast, continuous optimisation treats the adjacency matrix as a real-valued matrix. As with non-continuous score-based methods the goal of continuous optimisation is to maximise how well the DAG fits the data, but the requirement for acyclicity is expressed as an equality constraint on real values making the method an *equality constrained problem* (Vowels et al., 2021). One of the key advantages of this approach is that it is much closer to approaches used in mainstream machine learning and allows the use of powerful off-the-shelf optimisation approaches.

The Non-Combinatoric Optimization via Trace Augmented Lagrangian Structure (NOTEARS) (Zheng et al., 2018) algorithm was perhaps the first to use continuous optimisation for structure learning. NOTEARs represents the DAG as a $n \times n$ weighted adjacency matrix $W$ with elements $w_{ij}$ so that the value of any variable $X_i$ is given by:

$$X_i = \sum_{j \neq i} w_{ij} \cdot X_j + z_i$$

where $X_j$ iterates over the other variables, and $z_i$ is noise for the variable which can be non-Gaussian. If $w_{ij}$ is zero this indicates that there is no arc from $X_j$ to $X_i$. Therefore, matrix $W$ encapsulates both the DAG structure and the strength of relationships associated with each arc. This representation naturally fits where the variables take continuous values, but can be extended to discrete ordinal valued variables by using a logistic regression approach to obtain ordinal discrete values from continuous values. The objective function, $F(W)$, to be minimised is given by:

$$F(W) = \frac{1}{2N} \|\boldsymbol{X} - \boldsymbol{X}W\|^2 + \lambda \|W\|_1$$

where $\boldsymbol{X}$ is the complete dataset over $n$ variables and $N$ instances, and therefore $\boldsymbol{X}W$ is the predicted data values derived from the weighted adjacency matrix and values of parentless variables. Thus, the first term in $F(W)$ is the least square error of the predicted values minus actual data values and therefore a measure of data fitting. The second term is an L1 regulariser with $\|W\|_1$ being the sum of the absolute weight values. Its inclusion therefore has the effect of penalising complex DAGs.

The key contribution of Zheng et al. (2018) is to formulate the acyclicity requirement as the following continuous function constraint $h(W)$:

$$h(W) = tr(e^{W \circ W}) - n$$

where $\bigcirc$ is Hadamard matrix multiplication, $e^{matrix}$ is matrix exponentiation which may be expressed as a power series of matrix products, and $tr(matrix)$ is the matrix trace operation which is the sum of the diagonal elements. The derivation of this expression relies on the fact that the trace of a normal binary adjacency matrix raised to power $q$ equals the number of cycles of length $q$ in a directed graph, and this must be zero for all $q$ for a DAG. The constraint $h(W)$ has desirable properties such as being zero for a DAG, with lower values being closer to a DAG, and it is continuous and differentiable. This means that the problem can be solved by off-the-shelf optimisers, with NOTEARS using an augmented Lagrangian approach. Note that the problem is *non-convex* meaning that it has local minima so that NOTEARS is an approximate algorithm.

The authors compare NOTEARS with PC, LiNGAM and FGES in learning both Erdos-Rényi (Erdos and Rényi, 1960) and scale-free[9] random graphs (Barabási and Albert, 1999) which the authors argue can be more representative of real-world networks. They use $n \in$ $\{10, 20, 50, 100\}$ variables and $N = 20$ as a low sample size, and $N = 1000$ as a high sample size, and simulate Gaussian, Exponential and Gumbel noise. FGES and NOTEARS had comparable accuracy on sparser graphs, but NOTEARS was more accurate on denser networks across all noise types. We recommend that interested readers consult the review by Vowels et al. (2021) which covers continuous optimisation methods much more broadly.

---

[9] Scale-free graphs have a power law probability distribution for the node degree, $d$, of the form $P(d) \propto d^{-\gamma}$ where $\gamma$ is typically between 2 and 3. The resulting network has a small number of nodes with many tens or more of neighbours which are known as *hubs*. Biological networks such as gene regulation or metabolic networks are often scale-free, for example.

## 6. PRACTICAL CONSIDERATIONS

This section deals with some of the practical considerations when using BN structure learning algorithms in real world applications. The first subsection discusses some of the methods used to evaluate BN structure learning algorithms, and the second subsection describes the performance of these algorithms in terms of the quality of the networks they produce and their computational efficiency, and how their real performance compares to the theoretical or synthetic performance. The third subsection discusses how various forms of noise in the data can affect the quality of the learnt graph including some techniques which can mitigate those effects. Subsection 6.4 considers how expert knowledge can be incorporated into the learning process to improve the learnt networks, subsection 6.5 provides details of open-source packages and datasets relevant to structure learning, and the final subsection provides some guidelines for choosing and using algorithms.

### 6.1. Algorithm Evaluation

Evaluating structure learning algorithms can be a straightforward or a complicated process, depending on the selected evaluation approach. Indeed, it is fair to say there is no agreed process to determine the effectiveness of these algorithms (Korb and Nicholson, 2011), partly because of the different types and aims of the algorithms. The relevant literature consists of various evaluation criteria and, whilst many are similar, others represent entirely different concepts. In the absence of an agreed evaluation method, it is difficult to reach a consensus on the effectiveness of an algorithm. As a result, it is not infrequent to observe conflicting conclusions about which algorithm is 'best'.

Two main factors that tend to determine the evaluation criteria involve a) the learning class of the algorithm, each of which is described in Sections 2 to 4, and b) the data generation process. As shown in Table 5, there are two main types of evaluation, each of which is largely determined by the two above factors. Graphical evaluations correspond to scoring metrics that measure the differences between the learnt and ground truth graphs, whereas inference-based evaluations generally focus on how well the learnt distributions fit the observed distributions.

When the algorithms are applied to real data, the evaluation is generally not based on graph comparisons since the true graph is unknown. However, graphical-based evaluations are occasionally used when a knowledge-based graph is produced that can be compared to a learnt graph, as in (Kitson and Constantinou, 2021). As a result, most real-world applications of BN structure learning are evaluated in terms of inference.

Constraint-based learning tends not to be assessed with inference-based scores, at least in the case of synthetic experiments, because this learning class focuses on causal discovery which can only be effectively measured by means of graphical structure. On the other hand, inference-based evaluation is predominantly based on functions that only score-based algorithms employ. Therefore, while it makes sense to judge score-based algorithms in terms of the highest scoring graph achieved, it might be less appropriate to judge constraint-based algorithms by the same standards since they are based on a learning process that does not aim to maximise the global score of the learnt graph.

**Table 5** - The most common type of evaluation for different combinations of learning class and data generation process.

| Learning class | Data | |
|---|---|---|
| | **Synthetic** | **Real** |
| **Constraint-based** | Graphical | Inference |
| **Score-based** | Graphical, Inference | Inference |
| **Hybrid** | Graphical, Inference | Inference |

Metrics that focus on measuring the relationship between two graphical structures can be viewed as variants, often with modifications, of a confusion matrix that consists of:

a) True Positives (TP), corresponding to the number of true edges/arcs present in the learnt graph; i.e., number of corrects edges/arcs discovered,
b) False Positives (FP), corresponding to the number of false edges/arcs present in the learnt graph; i.e., number of incorrect edges/arcs discovered,
c) True Negatives (TN), corresponding to the number of true absent edges in the learnt graph; i.e., number of correct independence relationships discovered, and
d) False negatives (FN), corresponding to the number of false absent edges in the learnt graph; i.e., number of incorrect independence relationships discovered.

Two of the most commonly used metrics are *Precision* (P) and *Recall* (R), defined as

$$P = \frac{TP}{TP + FP} \quad \text{and} \quad R = \frac{TP}{TP + FN}$$

respectively. Specifically, Precision represents the rate of correct edges discovered across all edges discovered, whereas Recall represents the rate of edges discovered across all true edges that could have been discovered. Independently, however, these metrics can be misleading in judging the performance of an algorithm, which is why the F1 score is often preferred since it provides the harmonic mean of Precision and Recall. The F1 score is defined as:

$$\text{F1} = 2\frac{P.R}{P + R}$$

A frequent alternative, or an additional, metric is the Structural Hamming Distance (SHD) proposed by Tsamardinos et al. (2006). The SHD score represents the number of edge insertions, deletions and arc reversals needed to convert the learnt graph into the true graph, and is defined as

$$SHD = FN + FP$$

Tsamardinos et al. (2006) originally proposed using SHD to compare CPDAGs (representing equivalence classes) and this is the setting in which it is usually used, though it may also be applied to comparing DAGs. Variants of SHD are often applied and focus on assigning different penalty weights for edge insertion, deletions and arc reversals. The most frequent modification involves assigning a lower weight to arc reversals on the basis that an arc reversal corresponds to the discovery of an edge, albeit one with an incorrect direction (de Jongh and Druzdzel, 2009; Constantinou, 2019a).

Other structural metrics have focused on assessing the learnt graph in terms of causal effects, such as the Structural Intervention Distance (SID) by Peters and Buhlmann (2015) which measures the closeness of two DAGs in terms of their corresponding causal inference statements. More specifically, it is the number of ordered pairs of nodes $(X_i, X_j)$ where an intervention on $X_i$ produces a different intervention distribution in node $X_j$ in the two graphs $G_1, G_2$ being compared. The SID of DAG $G_2$ from DAG $G_1$, denoted $SID(G_1, G_2)$, is computed as the number of pairs of nodes $(A, B)$ where:

$$B \in \boldsymbol{Pa}^{G_2}(A) \text{ and } B \in \boldsymbol{De}^{G_1}(A)$$

plus the number of pairs of nodes $(A, B)$ where:

$B \notin \boldsymbol{Pa}^{G_2}(A)$ and $\boldsymbol{Pa}^{G_2}(A)$ meets one or both of the following conditions:
- some $Z \in \boldsymbol{Pa}^{G_2}$ is a descendant of any $W \neq A$ on a directed path from $A$ to $B$ in $G_1$
- $\boldsymbol{Pa}^{G_2}(A)$ does not d-separate $A$ and $B$ in $G_1$

where $A$ and $B$ represent nodes, $\boldsymbol{Pa}^{G_2}(A)$ are the parents of node $A$ in graph $G_2$ and $\boldsymbol{De}^{G_1}(A)$ are the descendants of node $A$ in graph $G_1$. Peters and Buhlmann (2015) also proposed a variant of SID which can be used to compare CPDAGs which only considers those pairs of nodes where the intervention is identifiable in the CPDAG. They also note that the SID metric only takes account of interventions on single nodes and that, in general, $SID(G_1, G_2) \neq SID(G_2, G_1)$.

Lastly, the Balanced Scoring Function (BSF) proposed by Constantinou (2019a) takes into consideration the complete confusion matrix to eliminate score imbalance by adjusting the reward function relative to the difficulty of discovering an edge, or the absence of an edge, proportional to their occurrence rate in the true graph. The BSF is defined as:

$$\text{BSF} = \left( \frac{\text{TP}}{|E|} + \frac{\text{TN}}{|M|} - \frac{\text{FP}}{|M|} - \frac{\text{FN}}{|E|} \right) \Big/ 2$$

where $|E|$ and $|M|$ represent the number of edges present and the number of edges absent (compared to the complete graph) in the true graph respectively, and

$$|M| = \frac{n \times (n-1)}{2} - |E|$$

On the other hand, the BD/e/u, Log-Likelihood, and BIC scores described in subsections 3.1.1 and 3.1.2 are the approaches most commonly used as an alternative to metrics of graphical discrepancy. Specifically, they are used to judge algorithms in terms of the highest scoring graph discovered with reference to the input data, according to the preferred scoring function. Although often less desirable than structural metrics, inference-based approaches can be extended to include other types of evaluation, such as predictive accuracy as determined by the Area Under the Curve (AUC) of the Receiving Operating Characteristic (ROC) (Fawcett, 2004).

In general, the BIC score is the most widely used metric across the various inference-based approaches, especially in real-world applications of BN structure learning. A problem with BIC, however, and which applies to most inference-based scores, is that it is score-equivalent; implying that it will generate the same score for multiple DAGs that are part of the same Markov equivalence class (refer to Fig 5). While this is not an issue when comparing CPDAGs, it is an issue when comparing DAGs which are generally required when applying structure learning to real problems.

An important limitation of inference-based scores is that a higher scoring graph, or higher predictive accuracy, do not necessarily reflect a more accurate causal structure (Constantinou et al., 2021b). For example, the highest BIC scoring graph across all possible graphs is often not the ground truth graph that generated the data. This limitation is exaggerated with limited and noisy data that distort the scores, and relaxed with big and clean data that increase our confidence in the scores generated. Despite their imperfections, inference-based evaluations are considered reasonably effective and represent an important metric for structure learning.

Lastly, a further limitation of most structural and inference metrics is that they scale with the number of variables in the graph, and in the case of inference metrics, with the sample size also. This makes comparisons between learning performance on different networks and datasets problematic. Several authors have attempted to address this by using scaled variants of the metrics. For example, Scutari et al. (2019a) used SHD divided by the number of arcs in the true graph in their comparative review of algorithms, and Malone et al. (2015) employed the Log-Likelihood divided by the product of the number of variables and sample size in their review of the generalisability of score-based algorithms. Note that BSF is one structural metric that has the advantage of not scaling with the number of variables.

*6.2. Algorithm Performance*

In this subsection, we consider the quality of the graphs that the algorithms produce, as well as their computational efficiency. Table 6 summarises some of the noteworthy papers which provide insights into algorithmic performance, and the type and scale of the data and networks to which they have been applied. Most papers which introduce a new structure learning algorithm tend to only evaluate it against previous algorithms of the same type, making it hard to get a picture of how they perform against the broad range of algorithms available. For this reason, we focus on comparative studies which cover a decent range of different types of algorithms, either studied as an end-goal in itself, or as part of study into some aspect of structure learning such as the effect of the objective function (Scutari, 2016) or noisy data (Constantinou et al., 2021b). Table 6 does include some papers which introduced a new algorithm: MMHC (Tsarmadinos et al., 2006); PC-Max (Ramsey, 2016) and FGES (Ramsey et al., 2017) where this has extended the scale of problems that have been tackled.

**Table 6** - Sources and characteristics of performance information used.

| Principal area of investigation | DATASET CHARACTERISTICS | | | | | | | | | ALGORITHMS EVALUATED | | | | | | | TYPE OF EVALUATION | | | |
|---|---|---|---|---|---|---|---|---|---|---|---|---|---|---|---|---|---|---|---|---|
| | Random Networks | Standard Networks | Real Datasets | Discrete variable | Continuous variables | Mixed variables | Number of variables | Sample Size | Average degree | Number of algorithms | PC Family | FCI Family | HC / Tabu | GES/FGES/FGS | Exact score-based | Hybrid algorithms | Structural | Inferential - fitting | Inferential - causal | Efficiency |
| Introduce MMHC Algorithm (Tsarmadinos et al., 2006) | | 10 | | Y | | | 20 - 801 | 500 - 20000 | 1.17-2.08 | 7 | Y | | | Y | | Y | Y | Y | | Y |
| Generalisation of score-based algorithms (Malone et al., 2015) | | 13 | 29 | Y | | | 17 - 60 | 31 - 20000 | n/a | 5 | | | Y | Y | Y | Y | Y | Y | | |
| Scoring Function and Structural Priors (Scutari, 2016) | | 10 | | Y | | | 8 - 442 | ~10 - $2.2\times10^6$ | 1.00 - 1.93 | 1 | | | Y | | | | Y | Y | | |
| Introduce PC-Max Algorithm (Ramsey, 2016) | 20 | | | | Y | | 1000 - 20000 | 1000 | 2.00 - 4.00 | 4 | Y | | | Y | | | Y | | | Y |
| Algorithms supporting latent variables and cycles (Singh et al., 2017) | 1 | | 2 | | Y | | 10 – 100 | 11 - 10000 | n/a | 10 | Y | Y | Y | | | Y | Y | Y | Y | |
| Scalability learning continuous variable networks (Ramsey and Andrews, 2017) | 270 | | | | Y | | 50 - 500 | 100 - 1000 | 2.00 - 6.00 | 18 | Y | | Y | Y | | Y | Y | | | Y |
| Introduce FGES Algorithm (Ramsey et al., 2017) | 443 | | | Y | Y | | 1000 - $10^6$ | 1000 | 1.00 - 2.00 | 1 | | | | Y | | | Y | | | Y |
| Investigate learning mixed variable type networks (Raghu et al., 2018) | 13 | | | | Y | Y | 50 - 100 | 100 - 5000 | ~3 – ~5 | 5 | Y | | Y | | | | Y | | | Y |
| Learning globally optimal graphs with exact score-based algorithms (Liao et al., 2019) | | 6 | 20 | Y | | | 10 – 57 | 32 – 58265 | n/a | 3 | | | | | Y | | | Y | | Y |
| Comparing score, constraint and hybrid algorithms (Scutari et al., 2019a) | | 14 | | Y | Y | | 20 – 442 | 23 – $3.9\times10^5$ | 1.18 – 2.06 | 9 | Y | | Y | Y | | | Y | | | Y |
| Evaluating hill-climbing algorithm on large sample sizes (Scutari et al., 2019b) | 1 | | 5 | | | Y | 19 – 37 | $10^6$ – $5.4\times10^7$ | n/a | 1 | | | Y | | | | Y | | | Y |
| The impact of noisy data on different algorithms (Constantinou et al., 2021b) | | 6 | | Y | | | 8 – 109 | 100 – $10^6$ | 2.00 – 3.58 | 15 | Y | Y | Y | Y | Y | Y | Y | | | Y |

### 6.2.1. Accuracy comparisons between algorithms

The wide range of algorithmic approaches discussed in Sections 2, 3 and 4 and the varied ways in which performance has been evaluated shown in Table 6, make it difficult to make definitive statements about the optimal approach to use in a particular context. As Table 6 suggests, the algorithms selected for comparison vary considerably between studies, though we note that most include an algorithm from the PC family, hill-climbing or Tabu, one from the GES family, and a hybrid algorithm, usually MMHC. These may thus be regarded as a 'benchmark' set of algorithms to which performance can be usefully compared. It seems rare for comparative studies to include sampling or genetic algorithms, presumably because they tend to be non-deterministic and produce a slightly different result each time they run, which makes it difficult to judge how these different kinds of approaches rank against the more established algorithms.

Comparisons between algorithms usually consider datasets with either discrete or continuous variables, though Scutari et. al. (2019a) considered both discrete and continuous variable networks, and Raghu et al. (2018) is one of the few studies that examines algorithms capable of learning networks containing a mixture of discrete and continuous variables. With discrete data, Scutari et al. (2019a) found that Tabu was the most accurate algorithm (14/20 cases) especially for large sample sizes (10/10 cases where sample size was equal to, or greater than, the number free parameters), and Constantinou et al. (2021b) similarly found that Tabu and hill-climbing ranked highest, overall. However, Scutari (2019a) found that Tabu was less accurate than the constraint and hybrid algorithms studied when learning Gaussian BNs, tending to produce too dense a graph.

Within score-based algorithms, Scutari (2019a) found that Tabu is more accurate than FGES (lower SHD in 18/20 cases), a finding echoed by Constantinou et al. (2021b). The study by Scutari (2019a) included a variant of the Order-MCMC sampling algorithm referred to as *Simulated Annealing* which fared poorly in terms of accuracy being the least accurate in 11/20 cases. Constantinou et al. (2021b) found that Tabu and hill-climbing ranked more highly in graphical accuracy than the exact GOBNILP algorithm, suggesting that the latter's theoretical advantages may be diminished with realistic sample sizes and/or noisy data, and that the highest scoring graph will not necessarily be closer to the ground truth graph than some other lower scoring graph. However, Malone et al. (2015) found evidence that exact algorithms (A* search and Integer Linear Programming) generalise better to unseen data than approximate algorithms do.

Scutari et al. (2019a) found that there was "*no systematic difference in accuracy*" when comparing constraint-based and hybrid algorithms, though they and Constantinou et al. (2021b) found that H2PC tended to perform better than other non-score-based algorithms. Many of the papers comparing constraint-based algorithms have considered only Gaussian networks. Singh et al. (2017) noted that all the constraint-based algorithms they investigated tended to learn a similar skeleton. Ramsey (2016), Ramsey and Andrews (2017) and Singh et al. (2017) all found that constraint-based algorithms obtained better edge precision than score-based approaches, but worse edge recall. The weaker faithfulness assumptions made by the CPC algorithm resulted in better arrowhead precision than the PC algorithm achieved, but arrowhead recall was worse (Raghu et al., 2018; Ramsey and Andrews, 2017). Raghu et al. (2018) and Ramsey (2016) found that algorithms from the PC and FCI family performed better than score-based algorithms on real data sets suggesting that they may perform better in the presence of latent variables and other forms of noise. On the other hand, Constantinou et al. (2021b) found that the algorithms that accounted for latent variables (FCI, GFCI and RFCI-BSC) did not offer improved graphical accuracy relative to other algorithms, in experiments which incorporated latent variables.

### 6.2.2. *Efficiency comparisons between algorithms*

As with accuracy comparisons, the literature provides a complex picture as to which algorithms are fastest. Tabu is found to be faster than both constraint-based and hybrid algorithms by Scutari et al. (2019a) when learning discrete networks, whereas Constantinou et al. (2021b) found some hybrid (MMHC) and constraint (Grow-Shrink) algorithms have runtimes closer to Tabu than other constraint-based and hybrid algorithms. As might be expected, exact score-based algorithms are considerably slower than approximate score-based algorithms (Constantinou et al., 2021b) and are generally limited to problems with less than 100 variables. In general, approximate score-based algorithms can tackle much higher dimensional problems, with Ramsey et al. (2017) using the FGES algorithm to learn both discrete and Gaussian BNs with 30,000 variables in a few minutes with parallel processing on a powerful laptop, and learning a Gaussian network with one million variables on a supercomputer. Scutari et al. (2019a) found that FGES was always faster than Tabu in the 20 discrete BN cases considered, whereas Constantinou et al. (2021b) found that FGES was considerably slower than Tabu and hill-climbing over the cases they considered.

Considering Gaussian networks only, the study by Ramsey and Andrews (2017) provides evidence about the importance of the implementation of an algorithm, such as which programming language is used, showing an order of magnitude speed difference between two implementations of the CPC algorithm. Tsarmadinos (2006) reported clear efficiency advantages of the MMHC algorithm being 41.35 and 10.09 times faster than the author's implementation of PC and Tabu respectively, whereas subsequent studies such as by Scutari et al. (2019a) have not borne out these advantages.

### 6.2.3. *Other factors affecting performance*

Many other factors, beside the choice of algorithm, affect the accuracy and speed of the structure learning process. One factor where there seems to be consistency across the different studies is that increasing sample size improves the accuracy of the learnt graph. For example, Malone et al. (2015) find that increasing sample size both improves the predictive likelihood of test datasets and reduces the variability of the predicted likelihood between different test datasets for both Tabu and exact score-based algorithms. Similarly, Liao et al. (2019) used the exact GOBNILP-DEV algorithm to show that the number of graphs within a certain factor of the global optimally scoring graph reduces as sample size increases; i.e., the highest scoring graphs become more differentiated as the sample size increases. Scutari et al. (2019b) investigated learning with very large sample sizes where a 24 variable CLGBN is learnt accurately with sample sizes of 5 million cases and above, though it is not clear how accuracy behaves with sample size in the general case.

The objective function chosen for score-based algorithms, as well as the CI test used in constraint-based algorithms, can have a large effect on the accuracy of the learnt graphs. Scutari (2016) showed that the choice of scoring function and structural priors within the hill-climbing algorithm affects the accuracy of the learnt graph as described in subsection 4.1.1. Indeed, Scutari et al. (2019a) argued that the choice of objective function and CI tests are confounding factors when comparing algorithms. Hence, they used equivalent objective functions and CI tests across all the algorithms they compared in their study. Raghu at al. (2018) showed that the choice of CI test had a large effect on structural accuracy when learning mixed variable type networks, and an even more dramatic effect on algorithm runtime with over three orders of magnitude difference in runtime of the CPC algorithm between using the fastest and slowest CI test. The choice of hyperparameters used, such as the ESS for Bayesian scoring functions (see subsection 4.1.1) and the significance level chosen for CI tests (subsection 3.1) affect performance too.

A further factor affecting performance is the dimensionality and quality of the data used. As Table 6 indicates, many studies make use of randomly generated networks or standard networks to provide a reference graph to which the learnt graph can be compared. In these cases, data is randomly generated synthetically to be consistent with the global probability distribution implied by the network parameters (e.g., the entries in the CPTs in discrete networks). Less commonly, real-world datasets are used but these generally suffer from having no reference ground-truth graph with which to compare the learnt graph. Structural learning has been investigated with a handful of variables right through to one million variables, with higher dimensionality studies generally using synthetically generated graphs and data. The sparsity of graphs used in structure learning tends to be much more consistent, with most studies investigating graphs with an average degree between one and six. The preponderance of synthetic graphs and synthetic data in evaluation studies raises the concern that they may not reflect real world performance. This was the motivation behind Constantinou et al.'s (2021b) evaluation of algorithms in the presence of synthetically generated noise, which did indeed find that graph accuracy in the presence of different forms of synthetic noise could decrease by up to 37%.

Lastly, comparative performance is strongly affected by the metrics chosen to evaluate the learnt graph. This includes the choice of the broad class of metrics used which typically include structural comparisons with a reference graph or inferential metrics based upon the likelihood of the data being generated by the learnt BN. In the latter approach, the same data is generally used to evaluate the graph that was used to learn the graph. This does not capture how well the learnt graph might generalise to new data. Malone et al. (2015) are relatively unique in having tried to assess the generalisability of the learnt graph by reporting metrics on subsets of test data. Also, notable from Table 6, is the lack of evaluation of the causal inferential properties of the learnt graph, such as how well it can predict the effect of interventions, or perform counterfactual inferences. This is surprising given that presumably one often wishes to learn causal BNs from data for these kinds of purposes. It should also be noted that, even within the same broad type of metric, different metrics can alter the ranking of algorithms. For example, Constantinou et al. (2021b) report different rankings according to whether the SHD or F1 metric is used.

### 6.3. Noise

The structural learning algorithms considered so far implicitly assume that the input data is a perfect sample from the underlying true distribution. However, there are often multiple types of noise in real-world observed data sets. For example, instrumental error or a survey respondents' unwillingness to respond to a question may mean that some of the values in a data instance may be missing, which we refer to as *missing data*. Even if the value is recorded in the data set, it may not be exactly same as the true value which is known as *measurement error*. This section will describe these two main types of noise in observed data and introduce algorithms that aim to handle these forms of noise in structure learning with discrete or continuous variables.

#### 6.3.1. Missing Data

Missing data is a common and serious problem in many real-world scientific research areas such as medical research and gene analysis. Rubin (1976) categorised missing data into three types: Missing Completely At Random (MCAR), Missing At Random (MAR) and Missing Not At Random (MNAR). In the MCAR case, the missingness of data is a purely random process and is not dependent on any other substantive variables. This type of missing data pattern is often caused by instrumental failure and normally would not bias the learnt graph. In the MAR case, the probability of a particular data value being missing is dependent on observed values. For example, in an investigation between age and frequency of smoking, data is MAR if younger respondents are more likely to not disclose their smoking frequency. Finally, data is MNAR if it is neither MCAR nor MAR. In this case, the probability of being missing may be related to missing values of other observed variables or even unobserved variables. In the above example, data is MNAR if the age variable also contains missing values.

One of the earliest algorithms which specifically deals with MAR missing data for structure learning is the structural EM algorithm (Friedman, 1997). Structural EM is an iterative process making the MAR assumption, which consists of two steps: an Expectation (E) step and a Maximisation (M) step. In the E step, Structural EM infers the missing values to produce a complete data set based on the current learned graph. Then, in the M step, Structural EM applies a standard structure learning algorithm using the inferred complete dataset to update the learned graph. Although Structural EM has the advantage of being able to work with any standard structure learning algorithm, and with both discrete and continuous variables, it is computationally inefficient due to the inference process in the E step.

More recently, several algorithms based on constraint-based structure learning have been proposed for handling missing data for continuous variables. Strobl et al. (2018) treated missing data as a type of selection bias and justified using test-wise deletion of the missing data in CI tests. Test-wise deletion is an approach which ignores data cases with missing values among the variables involved in a CI test. They showed that this a sound approach for handling missing data in the FCI and RFCI algorithms.

Gain and Shpitser (2018) proposed a modified version of PC called CBR-PC which uses the Inverse Probability Weighting (IPW) method (Horvitz and Thompson, 1952) on test-wise deleted data to construct an IPW-based CI test. IPW is an approach which alleviates distributional bias in data by reweighting each data case. By replacing the original CI test with IPW-based CI tests, CBR-PC maintains the consistency offered by PC but with missing data. However, IPW-based CI tests rely on knowing the causal relationships between observed variables and missing data, which is unlikely to be known in real-world problems. Tu et al. (2019) tackled this issue by firstly detecting the causal relationship between variables and missing data with a constraint-based process, and then used IPW-based CI tests in the PC algorithm.

Liu and Constantinou (2022) propose a modified version of HC called HC-aIPW which applies the test-wise deletion and IPW method to the score-based HC algorithm to deal with missing data in discrete variables. They utilised pairwise deletion and the IPW method with the HC algorithm, thus extending the approaches applied to constraint-based algorithms described in the previous paragraph to a greedy search score-based algorithm. The experimental results in their paper show that the HC-aIPW algorithm outperforms the commonly used Structural EM algorithm both when data are missing at random, and missing not at random.

### 6.3.2. Measurement Error

Measurement error is the discrepancy between the measured value of a variable and its true value, which can be treated as a disturbance from its underlying distribution. For continuous variables, the simplest way to model measurement error on measured variable $Y_m$ is to add a noise term $\epsilon_Y$ on its underlying error-free variable $Y$, i.e., $Y_m = Y + \epsilon_Y$. Under the presence of measurement error, the conditional independence relation detected from the measured variables may be different from the relation derived from their error-free variables. Consider the BN presented in Figure 19 and suppose we can acquire the true values of variables $X$ and $Z$, but only the measured values of variable $Y$ with measurement error $\epsilon_Y$. When $\epsilon_Y \neq 0$, and assuming faithfulness, the underlying independence relation $X \perp$ $Z \mid Y$ does not hold on the measured variables since the measured variable $Y_m$ does not d-separate $X$ and $Z$. The following three approaches apply to measurement error in continuous variables.

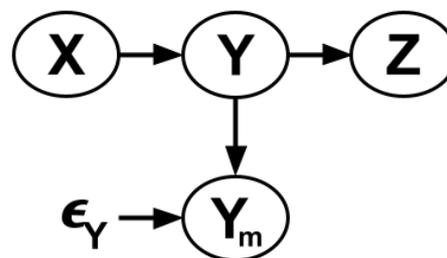

**Figure 19** - Example of measurement error.

Scheines and Ramsey (2016) studied the effect of Gaussian measurement error on the score-based FGES algorithm (Ramsey et al., 2017). They generated synthetic data based on Linear Gaussian models with additional random Gaussian noise to simulate the measurement error. Their results indicate that minor levels of measurement error can considerably worsen the learning accuracy of FGES on small sample sizes. However, when the sample size is 5,000, the accuracy of FGES remains relatively high, even when 80% of the variance of each variable is due to the measurement error.

Zhang et al. (2018) investigated the identifiability conditions for Linear Non-Gaussian Models in the presence of measurement error. Their theoretical result shows that under certain assumptions, the causal DAG remains fully identifiable learnt from noisy data by utilising overcomplete Independent Component Analysis (ICA) to learn the adjacency matrix. The assumptions used for their result include the causal Markov assumption, faithfulness assumption, non-linear noise assumption and two other assumptions that imply a sparse graphical structure.

Blom et al. (2018) proposed a method to detect the upper bound of the variance of random measurement error in observed variables for linear Gaussian models. Their method is based on the *tetrad constraints* (Silva et al., 2006) which can identify sets of four variables that are mutually d-separated by a common latent variable. After obtaining the upper bound, the conditional independence can be corrected based on the minimal and maximal partial correlation that lie within that bound. The measurement error upper bound was wrongly computed in just three cases and could not be detected in 39 cases in 200 simulation experiments.

Liu et al. (2020) studied the measurement error in discrete BNs and proposed a method called Spurious Edge Detection (SED) to remove spurious edges from the learned graph caused by measurement error. As illustrated in Figure 19, conditional independence relationships between error-free variables may not exist for measured variables in the presence of measurement error. Therefore, structure learning algorithms may learn spurious edges representing dependence relationships that do not exist in the true distribution. The SED algorithm firstly detects the candidate spurious edges that could form a 3-vertex clique, then assesses each candidate edge for removal based on an EM process. Their experimental results show that post-processing by the SED algorithm is able to generally improve the F1 and SHD performance of four different structure learning algorithms (HC, GOBNILP, PC-stable and H2PC) in the presence of measurement error.

### 6.4. Knowledge

The fusion of expert knowledge into the structure learning process provides a promising approach for improving performance and counteracting the effects of noise discussed in the previous two subsections. We discuss those approaches which seek to influence the learning process, known as *soft constraints*, and those which enforce requirements that the learnt structure must conform to, known as *hard constraints*. In both these cases, the knowledge is provided by experts to the algorithm before it starts the learning process. We also discuss *active learning* where the algorithm interacts with a human expert during the learning process.

### 6.4.1. Soft Constraints

Bayesian objective functions (see subsection 4.1.1) for score-based algorithms explicitly include terms for both the prior beliefs of the BN parameters and for a prior belief of each possible graph structure. These priors provide a Bayesian approach to incorporating expert knowledge into the learning process as a soft constraint. However, the vast numbers of possible graphs and resulting configurations of parameters make it impracticable to specify priors for individual parameters or graphs. As discussed in subsection 4.1.1, a standard conjugate prior (e.g. Dirichlet prior for discrete variables) is assumed for parameter priors, typically with a single hyperparameter value ESS. Likewise, it is not practical to provide a prior for each possible DAG, and therefore several approaches have been proposed where the human expert can provide priors on selected features, such as the presence or absence of a particular arc.

Castelo and Siebes (2000) provided a mechanism for an expert to assign a prior to the presence or absence of an arc between any number of pairs of variables. The remaining priors for the pair are assigned uniformly. For example, if an expert specifies a prior of 0.8 for arc $A \rightarrow B$, then arc $B \rightarrow A$ and no arc between $A$ and $B$ are both assigned a prior of 0.1. This scheme defines priors over the space of directed graphs that allow cycles. This represents a bigger space than DAG space, and so the approach redistributes probability mass from cyclic graphs to DAGs to correct for this. The approach was validated by showing that these priors could recover the true graph from synthetic noisy data. Eggeling et. al. (2019) evaluate more general structure priors which influence the overall characteristics of the learnt graph. This includes priors which are uniform across all DAGs, those which model each edge as having a defined uniform probability of existing, a prior which balances the probabilities for nodes with different in-degrees, and one which limits the maximum in-degree. The authors evaluate these priors with the GOBNILP algorithm for moderately sized networks, and Tabu for larger networks. The results show that priors which favour sparser graphs produce more accurate graphs especially at low sample sizes.

As well as constraints on particular arcs, it also possible to place constraints on the node ordering within the learnt structure. We have already encountered algorithms such as K2 (in subsection 4.2.1) where an ordering of all the nodes is specified. It is also possible to specify ordering constraints on a partial set of variables referred to as *ancestral constraints*, or since they are often derived from considerations of time and causality, *temporal constraints*. For example, Borboudakis and Tsamardinos (2013) allowed experts to specify a probability distribution over the possible ancestral relationships between a pair of variables ($A$ is an ancestor or parent of $B$, $B$ is an ancestor or parent of $A$, they have a common ancestor or parent, or they have no ancestral relationship). They combined the resulting joint prior probability distribution of ancestral relationships with an approximation of the number of DAGs consistent with each configuration of ancestral relationships to define a prior on DAGs. The authors incorporated this prior into a standard hill-climbing algorithm (see subsection 4.2.1) and also augmented the standard add/remove/reverse arc operations with a new *swap-equivalent* operator which swaps the current DAG for one in the same equivalence class with the highest ancestral prior. Experiments showed that modest amounts of prior ancestral knowledge, involving 12 out of 37 nodes, and using the new swap-equivalent operator could reduce the SHD error by around 15 to 20%.

Amirkhani et al. (2016) modified the objective function to include an explicit extra component representing the opinion of several, possibly conflicting, expert views on the presence and direction of arcs. The accuracy of each expert can also be specified. An Expectation-Maximisation variant of hill-climbing is described with the Expectation step adding, removing or reversing an arc to increase the score as usual, and the Maximisation step modifying the expert accuracy parameters to maximise the score of the current graph. The approach is evaluated using three synthetic sets of experts with their opinions generated from ground-truth graphs with three levels of accuracy: weak, mediocre and good. Using the 'good standard', experts halved SHD on the Alarm network (Beinlich et al., 1989) commonly used in evaluations, although improvement was limited when the weak set of experts were used.

Specification of the initial graph used in score-based algorithms can also be considered a form of soft-constraint since it may well influence the final learnt graph whilst not representing a hard constraint. Surprisingly, there seems to have been relatively little research undertaken into the influence of the initial graph in general, though Constantinou et al. (2021a) do investigate it. This is despite several hybrid algorithms using one algorithm to create a good starting point for a subsequent algorithm. Examples of such algorithms include SaiyanH and GSPo described in subsections 5.2.1 and 5.2.2, respectively.

### 6.4.2. Hard Constraints

Early work with hard constraints involved specifying features (arcs or edges) that were either required or prohibited in the learnt graph, or specifying a topological ordering that the learnt graph must be consistent with. De Campos and Castellano (2007) modified the score-based hill-climbing algorithm and the constraint-based PC algorithm to produce graphs consistent with hard constraints. Simulations involved specifying between 10 and 40% of possible required or prohibited features. In general, increasing the number of constraints improved all structural accuracies in graphs learnt by the modified hill-climbing algorithm. However, increasing the number of constraints, whilst improving most structural metrics, sometimes increased the number of extraneous arcs in the learnt graph with the PC algorithm. It was suggested this may have been because more constraints led to less CI tests being performed and hence less arcs being removed.

The Branch & Bound algorithm (de Campos et al., 2009) discussed in subsection 4.2.1 supports hard constraints to represent expert knowledge. These include specifying arcs which must appear in the result, and ones which are prohibited, as well as being able to limit the number of parents on an individual node basis. More unusually, it also supports *parameter constraints* which place restrictions on how parental values influence child values. These restrictions affect structural learning by changing the best scores achievable by different sets of parents.

More recent work has supported ancestral relationships as hard constraints. Borboudakis and Tsarmadinos (2012) apply ancestral constraints to PDAGs and PAGs produced by, for example, the PC and FCI algorithms respectively. They developed a set of algorithms which can determine variant endpoints in PDAGs and PAGs, or identify when the ancestral constraints are not consistent with the learnt graph. They found that even small numbers (around 10) of ancestral constraints can orientate around 30% of variant endpoints. This orientation effect is generally larger in PDAGs than PAGs, since constraints propagate more in the absence of confounding variables.

Li and van Beek (2018) described the MINOBSx algorithm which supports arc and edge, as well as ancestral, constraints which they claimed to be the first approximate score-based algorithm to do so. It is based on the approximate MINOBS algorithm which searches in ordering-based space (see subsection 4.2.3). They modified the parent set pruning rules and the approach used to determine a high-scoring DAG for each order, in order to take account of ancestral constraints. Li and van Beek (2018) compared MINOBSx with a MCMC approach which also supports constraints called CaMML (Korb and Nicholson, 2011), when learning well-known BNs of up to 48 variables. Li and van Beek (2018) found MINOBSx to be more robust than CaMML as it was always able to satisfy all the knowledge constraints specified. They noted that ancestral constraints reduced the number of arcs erroneously missing or mis-orientated in the learnt graph. However, ancestral constraints tended to encourage additional extraneous edges in the learnt graph. Ancestral constraints were more effective than simple arc/edge constraints in obtaining correct causal paths as measured by the SID metric (discussed in subsection 6.1).

Chen et al. (2016b) applied ancestral constraints to their exact score-based algorithm which searches an EC Tree (Chen et al., 2016a) described in subsection 4.3.3. To do so, they pruned CPDAGs prohibited by the ancestral constraints from the EC Tree and converted ancestral constraints into decomposable edge constraints which can then be used by the A* search heuristic. They found that their constrained A* search was typically orders of magnitude faster than the GOBNILP integer linear programming approach described in subsection 0, when it applied the same ancestral constraints to sub-graphs of well-known BNs with up to 20 variables.

Wang et al. (2021) have also implemented ancestral constraints in an exact score-based algorithm named ACOG from the space it searches, the Ancestral Constraint Ordering Graph (ACOG). The ACOG space is a modified version of the order graph shown in subsection 4.3.1, but where nodes in the graph have multiple DAGs associated with them to account for the effect of ancestral constraints, and pruning is applied using a novel revenue function that accounts for the ancestral constraints.

Constantinou et al. (2021a) provided a review of the effect of different forms of soft and hard constraint knowledge on the graphs learnt by Hill climbing, Tabu and SaiyanH algorithms. They found that specifying required arcs had the greatest effect followed by specifying an initial graph with edges in common with the true graph. Conversely, prohibiting arcs seemed to have little effect. This may be because graphs tend to have many more edges absent than edges present, and this reduces the impact of prohibiting, compared to enforcing, a given number of edges.

### 6.4.3. Active Learning

Active learning identifies relationships between the variables which the algorithm finds difficult to adjudicate on the basis of the data alone, and so where it may be advantageous to ask for human input during the learning process. This may require less human input than the normal approach of inputting expert knowledge before the algorithm begins, since it avoids supplying knowledge which merely duplicates something that the data also clearly implies.

Active learning was applied to an MCMC algorithm by Cano et al. (2011). They assumed a node ordering was known, and used a MCMC approach to generate a probability distribution over all possible parent sets for each variable. The algorithm asks the human expert to specify whether arcs having a probability close to 0.5 actually existed or not in their opinion. Two variations were explored, one where the expert was consulted after the MCMC learning algorithm had completed, and another where the expert was consulted during the MCMC learning process. Simulations were conducted on standard BNs containing between 23 and 56 variables with sample sizes $N = \{50, 100, 500, 1000\}$. Cano et al. (2011) found that expert specification of uncertain edges improved structural accuracy metrics, and to a lesser extent but perhaps more surprisingly, improved data fitting. The algorithm identified relatively few edges as uncertain, so that typically between 10 and 16 queries were directed at the expert after processing was completed with $N = 50$, and around 5 with $N = 1,000$. In these cases, expert specification of uncertain edges reduced SHD by around a quarter. Interacting with the expert during the MCMC learning process made similar structural improvements, but required fewer judgements from the expert.

Masegosa and Moral (2013) proposed a new hybrid restrict/maximise algorithm designed to support active learning which has three phases. The first phase constructs a probability distribution of plausible skeletons built from Markov Blankets, and the second phase builds a distribution of plausible DAGs using hill-climbing constrained by sampling from the first phase skeletons according to the relative probability of each skeleton. The distribution of plausible DAGs is improved in a third phase of unconstrained hill-climbing. A human expert can be asked about the existence of variables in Markov Blankets, or edges during hill-climbing if the algorithm judges that the answer would provide an information gain above a specified threshold. Masegosa and Moral (2013) evaluated their approach on standard networks with the number of variables varied between 23 and 56 and sample size $N = \{100, 500, 1000, 5000\}$. SHD improved by around 10% when knowledge was used. The number of queries asked of the expert ranged from 13-15 with $N = 100$, to 1-4 with $N = 5000$. Results were slightly better if knowledge was requested in both the Markov Blanket and DAG learning phases rather than just in the DAG learning phases.

### 6.5. Structure Learning Software Packages and Datasets

This subsection lists some of the open-source software packages that are freely available for BN structure learning and datasets which are often used to evaluate structure learning algorithms. The lists are not meant to be exhaustive but focus on the algorithms described in this paper, and datasets commonly used to evaluate them.

**Table 7** – Open-source software providing algorithms described in this paper.

| Name | Reference | Algorithms | Description or focus | Programming Language |
|---|---|---|---|---|
| BayesSuite | Michiels et. al., 2021 | **Constraint**: PC, GS, IAMB, Fast-IAMB and Inter-IAMB, MMPC<br>**Score**: HC, Tabu, Hiton-PC, FGES-Merge<br>**Hybrid**: MMHC | Includes visualisation and inference capabilities. Focussed on massive BNs particularly in neuroscience | Javascript and Python |
| Bayesys | Constantinou, 2019b | **Score**: HC, Tabu, MAHC<br>**Hybrid**: SaiyanH | Averaging version of score-based algorithms and incorporation of prior knowledge | Java |
| BiDAG | Suter et. al., 2021 | **Score**: Order-MCMC, Partition-MCMC<br>**Hybrid**: Hybrid MCMC (Kuipers et. al., 2022) | MCMC algorithms for sampling and learning the MAP DAG especially in large networks. | R |
| Bnlearn | Scutari, 2010 | **Constraint**: PC-Stable, GS, IAMB, Fast-IAMB, Inter-IAMB, MMPC, HITON-PC<br>**Score**: HC, Tabu<br>**Hybrid**: MMHC, H2PC | Well-established package often used as benchmark algorithms. | R |
| Bnstruct | Franzin et. al., 2017 | **Constraint**: MMPC<br>**Score**: HC, Dynamic Programming<br>**Hybrid**: MMHC | Package focuses on handling missing data including missing data imputation. | R |
| CausalExplorer | Aliferis et al., 2003 | **Constraint**: GS, IAMB, Inter-IAMB, TPDA, MMPC, MMMB, HITON, SI-HITON-PC, PC | Local and global constraint-based algorithms with a focus on bioinformatics | Matlab |
| Gobnilp | Cussens, 2020 | **Score**: GOBNILP | Often used as benchmark exact score-based algorithm | C or Python |
| Pcalg | Kalisch et. al., 2012 | **Constraint**: PC, CPC, PC-Stable, FCI, CFCI, FCI-Stable, RFCI, FCI+<br>**Score**: GES, GIES<br>**Hybrid**: ARGES | Includes algorithms which support latent variables and interventional data | R |
| Tetrad (including causal-learn) | Ramsey et. al., 2018 | **Constraint**: FCI, FCI-MAX, PC, PC-Stable, PC-MAX, CPC, RFCI<br>**Score**: GES, FGES, A*, DP<br>**Hybrid**: GFCI, RFCI-BSC | Large range of constraint-based algorithms including supporting latent variables and time-series. | Java (or Python) |

Table 7 provides details of some software packages that may be of interest and which provide algorithms described in this paper. The location where the software can currently be found is included in the References section at the end of this paper. Only the algorithms described in this paper are included, but note that these packages may include other algorithms. The programming language is given as this may be relevant for readers wishing to invoke them from their own software or learn the details of the algorithm from the program code. Note that we only list the 'primary' programming language, but some packages may provide 'wrappers' allowing easy access from other languages. Moreover, lower-level functions that need to be very performant are often written in languages such as C or C++.

**Table 8** – Repositories of networks and datasets commonly used to evaluate structure learning algorithms.

| Name | Reference | Variable Types | Number of networks | Number of variables | Reference graphs available | Notes |
|---|---|---|---|---|---|---|
| Bayesys | Constantinou et. al., 2020 | Discrete | 7 | 8 - 109 | yes | Includes datasets with synthetic noise |
| Bnlearn | Scutari, 2021 | Continuous, discrete and mixed | 27 | 5 – 1,041 | yes | Contains networks most often used in BNSL |
| DREAM5 | Marbach et. al., 2012 | Continuous | 4 | 1,643 – 5,667 | yes | Real gene regulatory data often used in high-dimensional BNSL |
| UCI ML | Dua and Graff, 2019 | Continuous, discrete and mixed | 622 | 3 – 3,231,961 | no | Real-world datasets used across machine learning |
| DEBD | Van Haaren and Davis, 2012 | Binary | 20 | 16 – 1,556 | no | Real-world datasets originally collected for Markov Network learning |

Table 8 lists some repositories of networks and datasets commonly used to evaluate structure learning algorithms. Perhaps the most common evaluation approach is to generate synthetic data from a known BN, attempt to learn the DAG from that data, and then compare the learnt DAG with the DAG used to generate the data. This approach is appropriate for the entries in Table 8 where reference graphs are available. Another approach to obtaining a reference graph is to generate random graphs, for example Erdos-Rényi graphs (Erdos and Rényi, 1960) or scale-free networks (Barabási and Albert, 1999), typically with some specified characteristic such as expected node degree. An alternative to generating a dataset from a reference graph is to learn from a 'real-world' dataset where the underlying reference graph is unknown. In that case, the learnt graph is typically evaluated using model selection functions that take into consideration both data fitting and model dimensionality; for example, the BIC score. Another possibility when there is no reference graph available, is to compare the predictive abilities with other machine learning approaches.

### 6.6. Guidelines on choosing and using structure learning algorithms.

This subsection is aimed mostly at practitioners who want to learn the structure of BNs but are not necessarily familiar with the mechanics of learning algorithms. Thus, this section focuses on the capabilities of the algorithms available in the packages listed in Table 7, which tend to be well-maintained and documented. In our view, there is little consensus in the literature on what might be the best algorithm in any particular context, so the focus here is on providing some guidelines to choosing and using algorithms, rather than attempting to provide a definitive guide. Note that some of the content in this subsection depends upon what software packages currently provide and is subject to change.

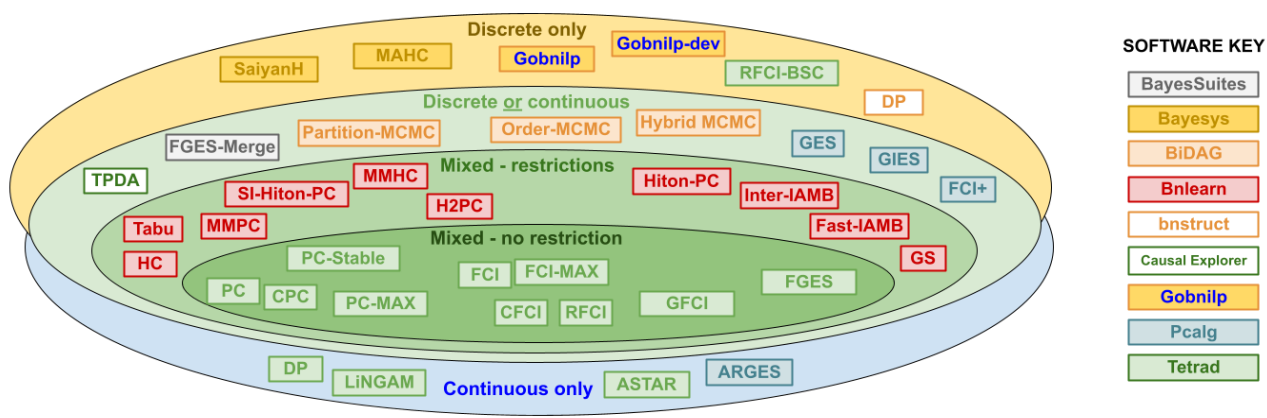

**Figure 20** - Supported variable types for algorithms in open-source software. Note that where an algorithm is implemented by more than one package, we show the package with the most flexible variable type support.

One factor which might affect algorithm choice is the data types of the variables. Figure 20 shows whether the algorithms offered by the software in Table 7 support continuous or discrete variables or both. Note that whether a particular algorithm supports a variable type is often not a fundamental property of the algorithm, but rather whether the software implementation supports CI tests or scores that support that variable type. We classify algorithms that support both continuous and discrete variables in three categories of increasing flexibility. Firstly, those algorithm implementations that support continuous or discrete variables, but not both in the same network, are shown in the lightest green area in Figure 20. Secondly, those algorithm implementations that support mixtures of continuous and discrete variables in the same network, but with the restriction that continuous variables cannot be parents of discrete variables. This kind of restricted mixed network in known as a *Conditional Linear Gaussian* network, and the algorithms which support it are shown in the mid-green area in the figure. Most flexibly of all, there are algorithms which support mixtures without that restriction. To our knowledge, only the Tetrad package offers this for some of its algorithms through the use of the Conditional Gaussian (Andrews et al., 2018) and Degenerate Gaussian (Andrews et al., 2019) scores or CI likelihood tests (Andrews et al., 2018).

With score-based approaches a further choice is whether to opt for an approximate or exact algorithm. Exact algorithms can be feasible up to around one hundred variables and would therefore seem to be preferred in low to medium dimensional settings. However, they may not offer improved accuracy where latent confounders, selection bias or measurement error is present, a finding supported by some of the results in Constantinou et al. (2021b). It is also often necessary to place a limit on in-degree for exact algorithms to be tractable, and this may not be appropriate where dense or scale-free graphs are expected.

Another consideration is whether the system may have latent confounders or selection variables present. The Tetrad software package focuses on this area and provides the FCI, CGCI, RFCI and FCI-Max algorithms that account for latent confounders and selection variables. FCI+ provided by the Pcalg software supports latent confounders too. Despite the theoretical advantages of these algorithms in causally insufficient settings, the study by Constantinou et al. (2021b) found that algorithms which did not account for latent confounders fared as well as those which did in simulated settings with latent confounders.

Another decision point is whether to choose an algorithm which returns a single graph, usually referred to as model selection, or one which returns multiple graphs. Even where the algorithm produces a single graph this generally represents multiple DAGs (or MAGs) when they are learning from observational data. This is explicit for constraint-based approaches which return a CPDAG (or PAG). However, most hybrid or score-based approaches return a DAG, but since they invariably use equivalent scores, this graph might be best regarded as an example from the equivalence class. We note that even in causally sufficient settings, interpretation of CPDAGs may be problematic as many constraint-based algorithms mark arcs as ambiguous if they encounter conflicts in the orientation phase, for example. Interpretation of PAGs is rather more difficult given that edge types can have several meanings. Generally speaking, causal inference applications (Noguiera et al., 2022) require a BN model with a well-defined DAG to be able to do causal inference. Incorporation of human knowledge either before the learning process, during the learning process (active learning) or by orientating undirected edges in the CPDAG can help towards obtaining a DAG for inference purposes. Inclusion of interventional data is another approach which can help orientate arcs, and we note that GIES (Hauser and Bühlmann, 2012) is provided in the Tetrad package.

In some cases, a practitioner's aim may be to identify the major causal effects rather than try to learn a complete causal graph. Here, approaches such as those provided by the BiDAG package which produce a sample of the most probable graphs may be useful. Model averaging over the sample graphs allow the most probable examples of features such as arcs or Markov Blankets to be identified. The approach can also be useful in understanding the uncertainty associated with particular graphs or features.

One of the most common and perhaps most robust method for evaluating the result of a learning algorithm is to compare the learnt graph structurally with the true causal graph. However, this is generally only possible in simulation studies, as a reference causal structure is not usually available in practical problems. If a reference graph is not available, then one may perform sensitivity studies to gain some understanding of the confidence one might have in the result. These can examine the sensitivity of the results to the choice of algorithm, score or CI test, and hyperparameters. Recent work by Kitson and Constantinou (2022) find that some competitive algorithms such as Tabu and MMHC are sensitive to the order of columns in the dataset and therefore this might also warrant investigation. Using a bootstrapping approach whereby graphs are learnt from sub-samples of the data can also cast light on the reliability of the result and provide some indication of whether the sample size is adequate.

To conclude, we note that no structure learning algorithm is perfect and that the literature is quite inconsistent about which algorithms are the most accurate. We note that algorithm ranking is probably quite sensitive to the metrics used, and the scale and nature of the system being modelled. In these circumstances we recommend using a range of suitable algorithms where possible, and considering model averaging across algorithms to identify those features which are most reliably identified. Finally, we note that the involvement of a domain expert to input some constraints into the learning process or review the learnt graphs is liable to be beneficial.

## 7. CONCLUDING REMARKS

Causal discovery is fundamental to understanding our world and predicting the effects of our interventions in it. BNs are an important tool for modelling causal relationships between variables, and hence inferring the effects of our interventions. This paper aims to provide a comprehensive review of the algorithms used to learn the graphical structures of BNs from data, and the ways in which knowledge can be incorporated into that process. We acknowledge that this is a large and rapidly advancing field but aim to have described the major developments over the past thirty years, many of which are still relevant today. Current state-of-the-art and pioneering algorithms are also described.

We begin with a brief introduction to Bayesian Networks including the concepts of DAGs, conditional independence, causal classes, equivalence classes and graphical separation. This aims to be a succinct but accessible introduction for someone relatively new to the field. The bulk of the paper then describes structure learning algorithms broken down into the standard categories of score-based, constraint-based and hybrid algorithms. For each algorithm covered, we present detail for the reader to understand the basic principles that the algorithm employs, and the assumptions it makes. The review is relatively comprehensive and covers 24 constraint-based, 22 score-based and 15 hybrid learning algorithm. Our goal is to highlight the similarities and differences between algorithms and, to that end, we use a consistent set of terms and emphasise the evolution of the algorithms and the relationships between them.

Constraint-based algorithms use CI tests to learn the structure, and we describe these tests next. Constraint-based algorithms that assume causal sufficiency are then covered, categorised into global algorithms that learn the graph skeleton as a whole, and local algorithms that learn the skeleton around each variable and then merge them. The end of this section covers algorithms which do not assume causal sufficiency and the ancestral graphs needed to represent the relationships between observed variables in that context. Score-based algorithms follow a more traditional machine learning route of assigning a score to each graph and exploring possible graphs to find a high, or ideally, the highest, scoring graph. The relevant section describes common choices for the scoring function before discussing exact algorithms which guarantee to return the highest scoring graph for the training dataset, and approximate algorithms which do not offer this guarantee. Sampling algorithms which provide a sample of the higher scoring graphs are also covered under approximate algorithms. We highlight the commonalities across score-based algorithms by grouping them according to the type of search space they explore. We then discuss hybrid algorithms which combine constraint-based and score-based approaches. The most common hybrid approach is to use a "restrict" constraint-based phase to define a reduced search space for the subsequent score-based "maximise" phase to explore. These are described in the first subsection. A second subsection describes other ways in which score-based and constraint-based approaches are combined. As with score-based algorithms, the search space employed is used as a way of categorising hybrid algorithms.

Our final substantive section discusses practical considerations when applying and evaluating the learning performance of these algorithms. We discuss the metrics used to evaluate structure learning, both in terms of graphical discovery and inference, and review the accuracy and runtime performance of algorithms by examining recent papers that focus on large scale comparative analyses. This provides an indication of the increasing scale of problems that can be tackled by the various algorithms and highlights inconsistencies within the literature about which algorithms might be 'best' for a given problem. Whilst papers proposing new algorithms generally include simulations demonstrating superior performance to previous algorithms, generally of the same type, we note that early approaches such as hill-climbing, GES and PC remain competitive in many comparative reviews.

We note that whilst many algorithms offer theoretical guarantees of accuracy such as asymptotically correct behaviour, these guarantees generally rely on unrealistic assumptions about the input data. We note that some or all of these assumptions are generally made for algorithms of all classes: exact and approximate score-based, constraint-based and hybrid ones. The quality of the learnt graphs is dependent on many complex interacting factors including the algorithm chosen, score or CI test employed, hyperparameter values, data size and quality, and the underlying graph. We note, too, that different evaluation metrics can paint different pictures about the superiority of one algorithm over another. This all makes algorithm comparisons rather problematic. Much of the evaluation of the algorithms is performed using synthetic graphs, and even more so, with synthetically generated clean data. Constantinou et al. (2021b) found that forms of noise likely to be found in real world data can have a considerable impact on learning performance. These forms of noise included latent variables, missing data, and measurement or discretisation error. Hence, we include a section on modifications that can be made to algorithms to account for various forms of noise.

Given the huge diversity of algorithms and factors affecting their performance, incorporating expert knowledge into the learning process may be helpful in many situations and hence, we conclude by reviewing approaches for doing this. The relevant subsection discusses hard constraints where expert knowledge is used to restrict the graphs which the algorithm can consider, as well as soft constraints which less rigidly influence the learning process. Approaches to active learning, whereby the algorithm itself identifies which knowledge would be most beneficial, are also discussed.

In conclusion, BN structure learning has become a vibrant research area. However, several key open questions remain, amongst which we would highlight:

- a huge diversity of structure learning approaches with little consensus on the most appropriate algorithm in any given real-world context;
- an absence of real-world datasets with accompanying causal graphs validated by experts, or ideally randomised controlled trials or experiments;
- a need to identify the best ways of incorporating knowledge and interventional data into the learning process;
- many confounding factors making algorithmic comparisons difficult, including sample size, distributional assumptions, faithfulness, linearity, choice of score or CI/test, hyperparameter values, data dimensionality, data noise, and the evaluation metrics.

**ACKNOWLEDGEMENTS**

This research was funded by the EPSRC Fellowship project EP/S001646/1 on Bayesian Artificial Intelligence for Decision Making under Uncertainty, and by the Alan Turing Institute in the UK under the EPSRC grant EP/N510129/1, and by the Royal Thai Government Scholarship offered by Thailand's Office of Civil Service Commission (OCSC).

**DATA AVAILABILITY STATEMENT**

Data sharing not applicable to this article as no datasets were generated or analysed during the current study.

# REFERENCES

[1] Akaike, H., 1974. A new look at the statistical model identification. *IEEE transactions on automatic control*, 19(6), pp.716-723.

[2] Aliferis, C.F., Statnikov, A., Tsamardinos, I., Mani, S. and Koutsoukos, X.D., 2010. Local causal and markov blanket induction for causal discovery and feature selection for classification part i: Algorithms and empirical evaluation. *Journal of Machine Learning Research*, 11(1).

[3] Aliferis, C.F., Tsamardinos, I. and Statnikov, A., 2003. HITON: a novel Markov Blanket algorithm for optimal variable selection. In *AMIA annual symposium proceedings* (Vol. 2003, p. 21). American Medical Informatics Association.

[4] Aliferis, C.F., Tsamardinos, I., Statnikov, A.R. and Brown, L.E., 2003, June. Causal Explorer: A Causal Probabilistic Network Learning Toolkit for Biomedical Discovery. In METMBS (Vol. 3, pp. 371-376).

[5] Amirkhani, H., Rahmati, M., Lucas, P.J. and Hommersom, A., 2016. Exploiting experts' knowledge for structure learning of Bayesian networks. *IEEE transactions on pattern analysis and machine intelligence*, 39(11), pp.2154-2170.

[6] Andersson, S. A., Madigan, D., and Perlman, M. D. (1997). A characterization of Markov equivalence classes for acyclic digraphs. *Annals of Statistics*, vol. 25, pp. 505–541.

[7] Anderson, T.W., 1962. *An introduction to multivariate statistical analysis* (No. 519.9 A53). New York: Wiley.

[8] Andrews, B., Ramsey, J. and Cooper, G.F., 2018. Scoring Bayesian networks of mixed variables. *International journal of data science and analytics*, 6(1), pp.3-18.

[9] Andrews, B., Ramsey, J. and Cooper, G.F., 2019, July. Learning high-dimensional directed acyclic graphs with mixed data-types. In The 2019 ACM SIGKDD Workshop on Causal Discovery (pp. 4-21). PMLR.

[10] Barabási, A.L. and Albert, R., 1999. Emergence of scaling in random networks. *Science*, 286(5439), pp.509-512.

[11] Bartlett, M., and Cussens, J. (2017). Integer linear programming for the Bayesian network structure learning problem. *Artificial Intelligence*, vol. 244, pp. 258–271.

[12] Behjati, S. and Beigy, H., 2020. Improved K2 algorithm for Bayesian network structure learning. Engineering Applications of Artificial Intelligence, 91, p.103617.

[13] Beinlich, I.A., Suermondt, H.J., Chavez, R.M. and Cooper, G.F., 1989. The ALARM monitoring system: A case study with two probabilistic inference techniques for belief networks. In *AIME 89* (pp. 247-256). Springer, Berlin, Heidelberg.

[14] Bernaola, N., Michiels, M., Larrañaga, P. and Bielza, C., 2020. Learning massive interpretable gene regulatory networks of the human brain by merging Bayesian Networks. bioRxiv. doi:10.1101/2020.02.05.935007. https://www.biorxiv.org/content/early/2020/02/05/2020.02.05.935007

[15] Bernstein, D., Saeed, B., Squires, C. and Uhler, C., 2020, June. Ordering-based causal structure learning in the presence of latent variables. In *International Conference on Artificial Intelligence and Statistics* (pp. 4098-4108). PMLR.

[16] Berthold, T., Gamrath, G., Gleixner, A., Heinz, S., Koch, T. and Shinano, Y., 2012. Solving mixed integer linear and nonlinear problems using the SCIP optimization suite.

[17] Blom, T., Klimovskaia, A., Magliacane, S. and Mooij, J.M., 2018. An upper bound for random measurement error in causal discovery. arXiv preprint arXiv:1810.07973.

[18] Borboudakis, G. and Tsamardinos, I., 2012, June. Incorporating causal prior knowledge as path-constraints in Bayesian networks and maximal ancestral graphs. In *Proceedings of the 29th International Conference on Machine Learning* (pp. 427-434).

[19] Borboudakis, G. and Tsamardinos, I., 2013, August. Scoring and searching over Bayesian networks with causal and associative priors. In *Proceedings of the Twenty-Ninth Conference on Uncertainty in Artificial Intelligence* (pp. 102-111).

[20] Bouckaert, R. (1994). Properties of Bayesian belief network learning algorithms. In *Proceedings of 10th Conference on Uncertainty Intelligence*, pp. 102–109.

[21] Bouchaert, R. (1995). Bayesian belief networks: from construction to inference. Ph.D thesis, University of Utrecht.

[22] Buntine, W. (1991). Theory refinement on Bayesian networks. In *Proceedings of the 7th Conference on Uncertainty in Artificial Intelligence*, pp. 52–60.

[23] Cadez, I., Heckerman, D., Meek, C., Smyth, P. and White, S., 2000, August. Visualization of navigation patterns on a web site using model-based clustering. In *Proceedings of the sixth ACM SIGKDD international conference on Knowledge discovery and data mining* (pp. 280-284).

[24] Cai, B., Huang, L. and Xie, M., 2017. Bayesian networks in fault diagnosis. *IEEE Transactions on Industrial Informatics*, 13(5), pp.2227-2240.

[25] Cano, A., Masegosa, A.R. and Moral, S., 2011. A method for integrating expert knowledge when learning Bayesian networks from data. *IEEE Transactions on Systems, Man, and Cybernetics, Part B (Cybernetics)*, 41(5), pp.1382-1394.

[26] Castelo, R. and Siebes, A., 2000. Priors on network structures. Biasing the search for Bayesian networks. *International Journal of Approximate Reasoning*, 24(1), pp.39-57.

[27] Chen, E.Y.J., Choi, A.C. and Darwiche, A., 2016a, May. Enumerating equivalence classes of Bayesian networks using EC graphs. In *Artificial Intelligence and Statistics* (pp. 591-599). PMLR.

[28] Chen, E.Y.J., Shen, Y., Choi, A. and Darwiche, A., 2016b. Learning Bayesian networks with ancestral constraints. *Advances in Neural Information Processing Systems*, 29, pp.2325-2333.

[29] Chen, Y. and Tian, J., 2014, June. Finding the k-best equivalence classes of Bayesian network structures for model averaging. In *Proceedings of the AAAI Conference on Artificial Intelligence* (Vol. 28, No. 1).

[30] Cheng, J., Bell, D.A. and Liu, W., 1997, January. Learning belief networks from data: An information theory based approach. In *Proceedings of the sixth international conference on Information and knowledge management* (pp. 325-331).

[31] Cheng, J. and Greiner, R., 1999, July. Comparing Bayesian network classifiers. In *Proceedings of the Fifteenth conference on Uncertainty in artificial intelligence* (pp. 101-108). Morgan Kaufmann Publishers Inc.

[32] Cheng, J., Greiner, R., Kelly, J., Bell, D., & Liu, W. (2002). Learning Bayesian networks from data: An information-theory based approach. *Artificial Intelligence*, vol. 137, Iss 1-2, pp. 43-90

[33] Chickering, D. (2002). Learning equivalence classes of Bayesian-network structures. *Journal of Machine Learning Research*, vol. 2 pp. 445–498.

[34] Chickering, D.M., Geiger, D. and Heckerman, D., 1994. Learning Bayesian networks is NP-hard (Vol. 196). Technical Report MSR-TR-94-17, Microsoft Research.

[35] Chickering, D.M., Heckerman, D. and Meek, C., 2004. Large-sample learning of Bayesian networks is NP-hard. Journal of Machine Learning Research, 5(Oct), pp.1287-1330.

[36] Chickering, D.M. and Meek, C., 2002, August. Finding optimal Bayesian networks. In *Proceedings of the Eighteenth conference on Uncertainty in artificial intelligence* (pp. 94-102).

[37] Chobtham, K. & Constantinou, A.C., 2020. Bayesian network structure learning with causal effects in the presence of latent variables. Proceedings of the 10th International Conference on Probabilistic Graphical Models, in Proceedings of Machine Learning Research 138:101-112

[38] Chobtham, K., Constantinou, A.C. and Kitson, N.K., 2022. Hybrid Bayesian network discovery with latent variables by scoring multiple interventions. *arXiv* preprint arXiv:2112.10574.

[39] Claassen, T. and Heskes, T., 2012, August. A Bayesian approach to constraint based causal inference. In Proceedings of the Twenty-Eighth Conference on Uncertainty in Artificial Intelligence (pp. 207-216)

[40] Claassen, T., Mooij, J.M. and Heskes, T., 2013, August. Learning sparse causal models is not NP-hard. In *Proceedings of the Twenty-Ninth Conference on Uncertainty in Artificial Intelligence* (pp. 172-181)

[41] Colombo, D. and Maathuis, M.H., 2014. Order-independent constraint-based causal structure learning. *The Journal of Machine Learning Research*, 15(1), pp.3741-3782.

[42] Colombo, D., Maathuis, M.H., Kalisch, M. and Richardson, T.S., 2012. Learning high-dimensional directed acyclic graphs with latent and selection variables. *The Annals of Statistics*, pp.294-321.

[43] Constantinou, A., 2019a. Evaluating structure learning algorithms with a balanced scoring function. arXiv:1905.12666 [cs.LG]

[44] Constantinou, A., 2019b. The Bayesys user manual. Queen Mary University of London, London, UK. [Online]. Software available: http://bayesian-ai.eecs.qmul.ac.uk/bayesys/

[45] Constantinou, A., 2020. Learning Bayesian Networks that enable full propagation of evidence. *IEEE Access*, Vol. 8, pp. 124845-123856

[46] Constantinou, A.C., Guo, Z. and Kitson, N.K., 2021a. The impact of prior knowledge on causal structure learning. arXiv preprint arXiv:2102.00473.

[47] Constantinou, A. C., Liu, Y., Chobtham, K., Guo, Z., and Kitson, N. K. (2020). The Bayesys data and Bayesian network repository. Queen Mary University of London, London, UK. [Online]. Available: http://bayesian-ai.eecs.qmul.ac.uk/bayesys/

[48] Constantinou, A.C., Liu, Y., Chobtham, K., Guo, Z. and Kitson, N.K., 2021b. Large-scale empirical validation of Bayesian Network structure learning algorithms with noisy data. *International Journal of Approximate Reasoning*, 131, pp.151-188.

[49] Constantinou, A.C., Liu, Y., Kitson, N.K., Chobtham, K. and Guo, Z., 2022. Effective and efficient structure learning with pruning and model averaging strategies. *International Journal of Approximate Reasoning*. Volume 151, Pages 292-321.

[50] Cooper, G., and Herskovits, E. (1992). A Bayesian method for the induction of probabilistic networks from data. *Machine Learning*, vol. 9, pp. 309–347.

[51] Cooper, G.F. and Yoo, C., 1999, July. Causal discovery from a mixture of experimental and observational data. In *Proceedings of the Fifteenth conference on Uncertainty in artificial intelligence* (pp. 116-125).

[52] Correia, A.H.C., de Campos, C.P. and van der Gaag, L.C., 2019, June. An Experimental Study of Prior Dependence in Bayesian Network Structure Learning. In International Symposium on Imprecise Probabilities: Theories and Applications (pp. 78-81).

[53] Correia, A.H.C., Cussens, J., and de Campos, C.P., 2020, August. On pruning for score-based Bayesian network structure learning. In International Conference on Artificial Intelligence and Statistics (pp. 2709-2718).

[54] Cussens, J., 2011. Bayesian network learning with cutting planes. In *Proceedings of the 27th Conference on Uncertainty in Artificial Intelligence* (UAI 2011) (pp. 153-160). AUAI Press.

[55] Cussens, J., 2012. An upper bound for bdeu local scores. In *Proceedings of 20th European Conference on Artificial Intelligence, Workshop of Algorithmnic Issues for Inference in Graphical Models*. IOS Press.

[56] Cussens, J., 2020, February. GOBNILP: Learning Bayesian network structure with integer programming. In International Conference on Probabilistic Graphical Models (pp. 605-608). PMLR. Software available: https://bitbucket.org/jamescussens/gobnilp/

[57] Dantzig, G., 2016. Linear programming and extensions. Princeton university press.

[58] Darwiche, A., 2009. Modeling and reasoning with Bayesian networks. Cambridge university press.

[59] de Campos L. (2006). A scoring function for learning Bayesian networks based on mutual information and conditional independence tests. *Journal of Machine Learning Research*, vol. 7, pp. 2149–2187.

[60] de Campos, L.M. and Castellano, J.G., 2007. Bayesian network learning algorithms using structural restrictions. *International Journal of Approximate Reasoning*, 45(2), pp.233-254.

[61] de Campos, L.M., Fernandez-Luna, J.M., Gámez, J.A. and Puerta, J.M., 2002. Ant colony optimization for learning Bayesian networks. *International Journal of Approximate Reasoning*, 31(3), pp.291-311.

[62] de Campos, C.P. and Ji, Q., 2010, July. Properties of Bayesian Dirichlet scores to learn Bayesian network structures. In *Twenty-Fourth AAAI Conference on Artificial Intelligence*.

[63] de Campos, L.M. and Puerta, J.M., 2001. Stochastic local and distributed search algorithms for learning belief networks. In *Proceedings of the III International Symposium on Adaptive Systems: Evolutionary Computation and Probabilistic Graphical Model* (pp. 109-115).

[64] de Campos, C.P., Scanagatta, M., Corani, G. and Zaffalon, M., 2018. Entropy-based pruning for learning Bayesian networks using BIC. *Artificial Intelligence*, 260, pp.42-50.

[65] de Campos, C.P., Zeng, Z. and Ji, Q., 2009, June. Structure learning of Bayesian networks using constraints. In *Proceedings of the 26th Annual International Conference on Machine Learning* (pp. 113-120).

[66] de la Fuente, A., Bing, N., Hoechele, I. and Mendes, P., 2004. Discovery of meaningful associations in genomic data using partial correlation coefficients. *Bioinformatics*, 20(18), pp.3565-3574.

[67] de Jongh, M., Druzdzel, M. J. (2009). A comparison of structural distance measures for causal Bayesian network models, in: Klopotek, M., Przepiorkowski, A., Wierzchon, S. T., and Trojanowski, K. (Eds.), Recent Advances in Intelligent Information Systems, Challenging Problems of Science, Computer Science Series, Academic Publishing House EXIT, pp. 443–456.

[68] Dua, D. and Graff, C., 2019. UCI Machine Learning Repository [http://archive.ics.uci.edu/ml]. Irvine, CA: University of California, School of Information and Computer Science

[69] Eaton, D., and Murphy, K. (2007). Bayesian structure learning using dynamic programming and MCMC. In *Proceedings of the 23rd Conference on Uncertainty in Artificial Intelligence*, pp. 101–108.

[70] Eggeling, R., Viinikka, J., Vuoksenmaa, A. and Koivisto, M., 2019, April. On structure priors for learning Bayesian networks. In The 22nd International Conference on Artificial Intelligence and Statistics (pp. 1687-1695). PMLR.

[71] Erdos, P. and Rényi, A., 1960. On the evolution of random graphs. Publ. Math. Inst. Hung. Acad. Sci, 5(1), pp.17-60.

[72] Fawcett, T., 2004. ROC graphs: Notes and practical considerations for researchers. *Machine learning*, 31(1), pp.1-38.

[73] Franzin, A., Sambo, F. and Di Camillo, B., 2017. bnstruct: an R package for Bayesian Network structure learning in the presence of missing data. Bioinformatics, 33(8), pp.1250-1252. Package available at: https://cran.r-project.org/web/packages/bnstruct/index.html

[74] Friedman, N., 1997, July. Learning belief networks in the presence of missing values and hidden variables. In *ICML* (Vol. 97, No. July, pp. 125-133).

[75] Friedman, N. and Koller, D., 2003. Being Bayesian about network structure. A Bayesian approach to structure discovery in Bayesian networks. Machine learning, 50(1-2), pp.95-125

[76] Friedman, N., Nachman, I. and Peér, D., 1999, July. Learning Bayesian network structure from massive datasets: the "sparse candidate" algorithm. In *Proceedings of the Fifteenth conference on Uncertainty in artificial intelligence* (pp. 206-215).

[77] Gain, A. and Shpitser, I., 2018, Augest. Structure learning under missing data. In *International Conference on Probabilistic Graphical Models* (pp. 121-132). PMLR.

[78] Gasse, M., Aussem, A. and Elghazel, H., 2014. A hybrid algorithm for Bayesian network structure learning with application to multi-label learning. *Expert Systems with Applications*, 41(15), pp.6755-6772

[79] Geiger, D. and Heckerman, D., 1994. Learning gaussian networks. In *Uncertainty Proceedings 1994* (pp. 235-243). Morgan Kaufmann.

[80] Geiger, D. and Heckerman, D., 2002. Parameter priors for directed acyclic graphical models and the characterization of several probability distributions. *The Annals of Statistics*, 30(5), pp.1412-1440.

[81] Gillispie, S.B. and Perlman, M.D., 2002. The size distribution for Markov equivalence classes of acyclic digraph models. *Artificial Intelligence*, 141(1-2), pp.137-155.

[82] Glymour, C., Zhang, K. and Spirtes, P., 2019. Review of causal discovery methods based on graphical models. *Frontiers in genetics*, 10, p.524

[83] Goudie, R., and Mukherjee, S. (2016). A Gibbs sampler for learning DAGs. *Journal of Machine Learning Research*, vol. 17, pp. 1–39.

[84] Gretton, A., Spirtes, P. and Tillman, R., 2009. Nonlinear directed acyclic structure learning with weakly additive noise models. *Advances in neural information processing systems*, 22, pp.1847-1855

[85] Grzegorczyk, M., and Husmeier, D. (2008). Improving the structure MCMC sampler for Bayesian networks by introducing a new edge reversal move. *Machine Learning*, vol. 71, pp. 265–305.

[86] Guo, Z. and Constantinou, A.C., 2020. Approximate learning of high dimensional Bayesian network structures via pruning of Candidate Parent Sets. *Entropy*, 22(10), p.1142.

[87] Hauser, A. and Bühlmann, P., 2012. Characterization and greedy learning of interventional Markov equivalence classes of directed acyclic graphs. The Journal of Machine Learning Research, 13(1), pp.2409-2464.

[88] Heckerman, D., Geiger, D., and Chickering, D. (1995). Learning Bayesian networks: the combination of knowledge and statistical data. *Machine Learning*, vol. 20, pp. 197–243.

[89] Heckerman, D., Meek, C., and Cooper, G. (1997). A Bayesian approach to causal discovery. Technical Report MSR-TR-97-5, Microsoft Research

[90] Horvitz, D.G. and Thompson, D.J., 1952. A generalization of sampling without replacement from a finite universe. *Journal of the American statistical Association*, 47(260), pp.663-685.

[91] Hoyer, P., Janzing, D., Mooij, J.M., Peters, J. and Schölkopf, B., 2008a. Nonlinear causal discovery with additive noise models. *Advances in neural information processing systems*, 21.

[92] Hoyer, P.O., Shimizu, S., Kerminen, A.J. and Palviainen, M., 2008b. Estimation of causal effects using linear non-Gaussian causal models with hidden variables. International Journal of Approximate Reasoning, 49(2), pp.362-378.

[93] Hyttinen, A., Eberhardt, F. and Järvisalo, M., 2014, July. Constraint-based Causal Discovery: Conflict Resolution with Answer Set Programming. In *UAI* (pp. 340-349).

[94] Hyvärinen, A. and Oja, E., 2000. Independent component analysis: algorithms and applications. *Neural networks*, 13(4-5), pp.411-430.

[95] Imoto, S., Higuchi, T., Goto, T., Tashiro, K., Kuhara, S. and Miyano, S., 2004. Combining microarrays and biological knowledge for estimating gene networks via Bayesian networks. *Journal of bioinformatics and computational biology*, 2(01), pp.77-98.

[96] Jaakkola, T., Sontag, D., Globerson, A. and Meila, M., 2010, March. Learning Bayesian network structure using LP relaxations. In *Proceedings of the Thirteenth International Conference on Artificial Intelligence and Statistics* (pp. 358-365).

[97] Jabbari, F., Ramsey, J., Spirtes, P. and Cooper, G., 2017, September. Discovery of causal models that contain latent variables through Bayesian scoring of independence constraints. In *Joint European Conference on Machine Learning and Knowledge Discovery in Databases* (pp. 142-157). Springer, Cham.

[98] Jennings, D., and Corcoran, J. (2018). A birth and death process for Bayesian network structure inference. *Probability in the Engineering and Informational Sciences*, vol. 32, pp.615–625.

[99] Ji, J., Wei, H. and Liu, C., 2013. An artificial bee colony algorithm for learning Bayesian networks. *Soft Computing*, 17(6), pp.983-994.

[100] Kalisch, M., Mächler, M., Colombo, D., Maathuis, M.H. and Bühlmann, P., 2012. Causal inference using graphical models with the R package pcalg. Journal of Statistical Software, 47(11), pp.1-26. Package available: http://CRAN.R-project.org/package=pcalg

[101] Kitson, N. K., & Constantinou, A. (2021). Learning Bayesian networks from demographic and health survey data. *Journal of Biomedical Informatics*, Vol. 113, Article 103588

[102] Kitson, N.K. and Constantinou, A.C., 2022. The Impact of Variable Ordering on Bayesian Network Structure Learning. *arXiv preprint* arXiv:2206.08952.

[103] Koivisto, M. and Sood, K., 2004. Exact Bayesian structure discovery in Bayesian networks. *Journal of Machine Learning Research*, 5(May), pp.549-573.

[104] Koller, D. and Friedman, N., 2009. Probabilistic graphical models: principles and techniques. MIT press.

[105] Kontkanen, P. and Myllymäki, P., 2007. A linear-time algorithm for computing the multinomial stochastic complexity. *Information Processing Letters*, 103(6), pp.227-233.

[106] Korb, K. and Nicholson, A. (2011). *Bayesian Artificial Intelligence* (Second Edition). London, UK: CRC Press.

[107] Kuipers, J. and Moffa, G., 2017. Partition MCMC for inference on acyclic digraphs. *Journal of the American Statistical Association*, 112(517), pp.282-299.

[108] Kuipers, J., Moffa, G. and Heckerman, D., 2014. Addendum on the scoring of Gaussian directed acyclic graphical models. The Annals of Statistics, 42(4), pp.1689-1691.

[109] Kuipers, J., Suter, P. and Moffa, G., 2022. Efficient sampling and structure learning of Bayesian networks. *Journal of Computational and Graphical Statistics*, pp.1-12.

[110] Larranaga, P., Kuijpers, C.M., Murga, R.H. and Yurramendi, Y., 1996a. Learning Bayesian network structures by searching for the best ordering with genetic algorithms. *IEEE transactions on systems, man, and cybernetics-part A: systems and humans*, 26(4), pp.487-493.

[111] Larranaga, P., Poza, M., Yurramendi, Y., Murga, R.H. and Kuijpers, C.M.H., 1996b. Structure learning of Bayesian networks by genetic algorithms: A performance analysis of control parameters. *IEEE transactions on pattern analysis and machine intelligence*, 18(9), pp.912-926.

[112] Lee, C. and van Beek, P., 2017, May. Metaheuristics for score-and-search Bayesian network structure learning. In *Canadian Conference on Artificial Intelligence* (pp. 129-141). Springer, Cham.

[113] Lee, J.D. and Hastie, T.J., 2015. Learning the structure of mixed graphical models. *Journal of Computational and Graphical Statistics*, 24(1), pp.230-253

[114] Li, A. and van Beek, P., 2018, August. Bayesian Network Structure Learning with Side Constraints. In *International Conference on Probabilistic Graphical Models* (pp. 225-236).

[115] Liao, Z.A., Sharma, C., Cussens, J. and van Beek, P., 2019, July. Finding All Bayesian Network Structures within a Factor of Optimal. In *Proceedings of the AAAI Conference on Artificial Intelligence* (Vol. 33, pp. 7892-7899).

[116] Liu, Y. and Constantinou, A.C., 2022. Greedy structure learning from data that contain systematic missing values. *Machine Learning*, 111(10), pp.3867-3896.

[117] Liu, Y., Constantinou, A.C. and Guo, Z., 2020. Improving Bayesian Network Structure Learning in the presence of Measurement Error. *arXiv preprint* arXiv:2011.09776.

[118] Liu, Z., Malone, B., and Yuan, C. (2012). Empirical evaluation of scoring functions for Bayesian network model selection. *BMC Bioinformatics,* vol. 13.

[119] Madigan, D., Andersson, S.A., Perlman, M.D. and Volinsky, C.T., 1996. Bayesian model averaging and model selection for Markov equivalence classes of acyclic digraphs. Communications in Statistics--Theory and Methods, 25(11), pp.2493-2519.

[120] Madigan, D., York, J. and Allard, D., 1995. Bayesian graphical models for discrete data. International Statistical Review/Revue Internationale de Statistique, pp.215-232.

[121] Malone, B.M., Järvisalo, M. and Myllymäki, P., 2015, July. Impact of Learning Strategies on the Quality of Bayesian Networks: An Empirical Evaluation. In *UAI* (pp. 562-571).

[122] Marella, D. and Vicard, P., 2022. Bayesian network structural learning from complex survey data: a resampling based approach. Statistical Methods & Applications, pp.1-33.

[123] Margaritis, D., 2003. Learning Bayesian network model structure from data (No. CMU-CS-03-153). Carnegie-Mellon Univ Pittsburgh Pa School of Computer Science.

[124] Margaritis, D., and Thrun, S., 1999. Bayesian network induction via local neighborhoods. In *Proceedings of the 12[th] International Conference on Neural Information Processing Systems*, pp. 505–511.

[125] Masegosa, A.R. and Moral, S., 2013. An interactive approach for Bayesian network learning using domain/expert knowledge. *International Journal of Approximate Reasoning*, 54(8), pp.1168-1181.

[126] Meek, C. (1995). Causal inference and causal explanation with background knowledge. In *Proceedings of the 11[th] UAI Conference on Uncertainty in Artificial Intelligence*, pp. 403–410.

[127] Michiels, M., Larranaga, P. and Bielza, C., 2021. BayeSuites: An open web framework for massive Bayesian networks focused on neuroscience. Neurocomputing, 428, pp.166-181. Framework available at https://neurosuites.com/morpho/ml_bayesian_networks

[128] Moffa, G., Catone, G., Kuipers, J., Kuipers, E., Freeman, D., Marwaha, S., Lennox, B.R., Broome, M.R. and Bebbington, P., 2017. Using directed acyclic graphs in epidemiological research in psychosis: an analysis of the role of bullying in psychosis. Schizophrenia bulletin, 43(6), pp.1273-1279.

[129] Moraffah, R., Karami, M., Guo, R., Raglin, A. and Liu, H., 2020. Causal interpretability for machine learning-problems, methods and evaluation. *ACM SIGKDD Explorations Newsletter*, 22(1), pp.18-33.

[130] Nandy, P., Hauser, A. and Maathuis, M.H., 2018. High-dimensional consistency in score-based and hybrid structure learning. *The Annals of Statistics*, 46(6A), pp.3151-3183.

[131] Niinimäki, T., Parviainen, P. and Koivisto, M., 2011, July. Partial order MCMC for structure discovery in Bayesian networks. In *Proceedings of the Twenty-Seventh Conference on Uncertainty in Artificial Intelligence* (pp. 557-564).

[132] Nogueira, A.R., Pugnana, A., Ruggieri, S., Pedreschi, D. and Gama, J., 2022. Methods and tools for causal discovery and causal inference. *Wiley Interdisciplinary Reviews: Data Mining and Knowledge Discovery*, 12(2), p.e1449.

[133] Ogarrio, J.M., Spirtes, P. and Ramsey, J., 2016, August. A hybrid causal search algorithm for latent variable models. In *Conference on Probabilistic Graphical Models* (pp. 368-379).

[134] Ott, S., Imoto, S. and Miyano, S., 2003. Finding optimal models for small gene networks. In Biocomputing 2004 (pp. 557-567).

[135] Pearl, J. (1985). Bayesian Networks: A model of self-activated memory for evidential reasoning. In *Proceedings of the $7^{th}$ Conference of the Cognitive Science Society*, pp. 329–334.

[136] Pearl, J., 1988. *Probabilistic Reasoning in Intelligent Systems: Networks of Plausible Inference*. Morgan Kaufmann.

[137] Pearl, J. and Mackenzie, D., 2018. The book of why: the new science of cause and effect. Basic books.

[138] Pearl, J. and Verma, T., 1987, July. The logic of representing dependencies by directed graphs. In Proceedings of the sixth National conference on Artificial intelligence-Volume 1 (pp. 374-379).

[139] Pensar, J., Talvitie, T., Hyttinen, A. and Koivisto, M., 2020, April. A Bayesian approach for estimating causal effects from observational data. In Proceedings of the AAAI Conference on Artificial Intelligence (Vol. 34, No. 04, pp. 5395-5402).

[140] Perrier, E., Imoto, S. and Miyano, S., 2008. Finding optimal Bayesian network given a super-structure. *Journal of Machine Learning Research*, 9(Oct), pp.2251-2286.

[141] Peters, J., and Buhlmann, P. (2015). Structural Intervention Distance (SID) for Evaluating Causal Graphs. *Neural Computation*, vol. 27, Iss. 3, pp. 771–799

[142] Raghu, V.K., Ramsey, J.D., Morris, A., Manatakis, D.V., Sprites, P., Chrysanthis, P.K., Glymour, C. and Benos, P.V., 2018. Comparison of strategies for scalable causal discovery of latent variable models from mixed data. *International journal of data science and analytics*, 6(1), pp.33-45.

[143] Ramsey, J., 2016. Improving accuracy and scalability of the pc algorithm by maximizing p-value. arXiv preprint arXiv:1610.00378.

[144] Ramsey, J.D. and Andrews, B., 2017. A comparison of public causal search packages on linear, gaussian data with no latent variables. *arXiv preprint* arXiv:1709.04240.

[145] Ramsey, J., Glymour, M., Sanchez-Romero, R. and Glymour, C., 2017. A million variables and more: the Fast Greedy Equivalence Search algorithm for learning high-dimensional graphical causal models, with an application to functional magnetic resonance images. *International journal of data science and analytics*, 3(2), pp.121-129.

[146] Ramsey, J., Spirtes, P. and Zhang, J., 2006, July. Adjacency-faithfulness and conservative causal inference. In *Proceedings of the Twenty-Second Conference on Uncertainty in Artificial Intelligence* (pp. 401-408).

[147] Ramsey, J.D., Zhang, K., Glymour, M., Romero, R.S., Huang, B., Ebert-Uphoff, I., Samarasinghe, S., Barnes, E.A. and Glymour, C., 2018. TETRAD—A toolbox for causal discovery. In 8th International Workshop on Climate Informatics.

[148] Raskutti, G. and Uhler, C., 2013. Learning directed acyclic graphs based on sparsest permutations. arXiv preprint arXiv:1307.0366

[149] Raskutti, G. and Uhler, C., 2018. Learning directed acyclic graph models based on sparsest permutations. *Stat*, 7(1), p.e183.

[150] Richardson, T.S., 2009, June. A factorization criterion for acyclic directed mixed graphs. In *Proceedings of the Twenty-Fifth Conference on Uncertainty in Artificial Intelligence* (pp. 462-470).

[151] Richardson, T. and Spirtes, P., 2002. Ancestral graph Markov models. *The Annals of Statistics*, 30(4), pp.962-1030.

[152] Rissanen, J. (1996). Fisher information and stochastic complexity. *IEEE Transactions on Information Theory*, 42(1), pp. 40–47.

[153] Robinson, R. W. (1973). "Counting labeled acyclic digraphs", in Harary, F. (ed.), *New Directions in the Theory of Graphs*, Academic Press, pp. 239–273.

[154] Rubin, D.B. (1976). Inference and missing data. *Biometrika*, 63(3), pp.581-592.

[155] Sachs, K., Perez, O., Pe'er, D., Lauffenburger, D.A. and Nolan, G.P., 2005. Causal protein-signaling networks derived from multiparameter single-cell data. Science, 308(5721), pp.523-529.

[156] Scanagatta, M., Corani, G. and Zaffalon, M., 2017, September. Improved local search in Bayesian networks structure learning. In Advanced Methodologies for Bayesian Networks (pp. 45-56).

[157] Scanagatta, M., de Campos, C.P., Corani, G. and Zaffalon, M., 2015. Learning Bayesian networks with thousands of variables. In *Advances in neural information processing systems* (pp. 1864-1872).

[158] Scheines, R. and Ramsey, J., 2016, June. Measurement error and causal discovery. In CEUR workshop proceedings (Vol. 1792, p. 1). NIH Public Access.

[159] Scutari, M., 2010. Learning Bayesian Networks with the bnlearn R Package. Journal of Statistical Software, 35, pp.1-22. Software available at: https://bnlearn.com/

[160] Scutari, M., 2016, August. An empirical-Bayes score for discrete Bayesian networks. In Conference on probabilistic graphical models (pp. 438-448).

[161] Scutari, M., 2021. Bayesian Network Repository. https://www.bnlearn.com/bnrepository/.

[162] Scutari, M., Graafland, C.E. and Gutiérrez, J.M., 2019a. Who learns better Bayesian network structures: Accuracy and speed of structure learning algorithms. *International Journal of Approximate Reasoning*, 115, pp.235-253.

[163] Scutari, M., Vitolo, C. and Tucker, A., 2019b. Learning Bayesian networks from big data with greedy search: computational complexity and efficient implementation. Statistics and Computing, 29(5), pp.1095-1108.

[164] Sesen, M.B., Nicholson, A.E., Banares-Alcantara, R., Kadir, T. and Brady, M., 2013. Bayesian networks for clinical decision support in lung cancer care. PloS one, 8(12), p.e82349.

[165] Shimizu, S., Hoyer, P.O., Hyvärinen, A. and Kerminen, A., 2006. A linear non-Gaussian acyclic model for causal discovery. *Journal of Machine Learning Research*, 7(Oct), pp.2003-2030.

[166] Silander, T., and Myllymaki, P. (2006). A simple approach for finding the globally optimal Bayesian network structure. In *Proceedings of the 22nd Conference on Uncertainty in Artificial Intelligence*, pp. 445–452.

[167] Silander, T., Roos, T., Kontkanen, P., and Myllymaki, P. (2008). Factorized normalized maximum likelihood criterion for learning Bayesian network structures. In *Proceedings of the 4th European Workshop on Probabilistic Graphical Models*, pp. 257–264.

[168] Silander, T., Roos, T., and Myllymaki, P. (2010). Learning locally minimax optimal Bayesian networks. *International Journal of Approximate Reasoning*, vol. 51, pp. 544–557.

[169] Silander, T., Leppa-aho, J., Jaasaari, E., and Roos, T. (2018). Quotient normalized maximum likelihood criterion for learning Bayesian network structures. In *Proceedings of the 21st International Conference on Artificial Intelligence and Statistics*, pp. 948–957.

[170] Silva, R., Scheines, R., Glymour, C., Spirtes, P. and Chickering, D.M., 2006. Learning the Structure of Linear Latent Variable Models. *Journal of Machine Learning Research*, 7(2).

[171] Singh, A., and Moore, A., 2005. Finding optimal Bayesian networks by dynamic programming. *Technical Report CMU-CALD-05-106*, Carnegie Mellon University.

[172] Singh, K., Gupta, G., Tewari, V. and Shroff, G., 2017. Comparative benchmarking of causal discovery techniques. *arXiv preprint* arXiv:1708.06246.

[173] Singh, M. and Valtorta, M., 1993, January. An algorithm for the construction of Bayesian network structures from data. In *Uncertainty in Artificial Intelligence* (pp. 259-265). Morgan Kaufmann.

[174] Solus, L., Wang, Y. and Uhler, C., 2017. Consistency Guarantees for Greedy Permutation-Based Causal Inference Algorithms. *arXiv preprint* arXiv:1702.03530.

[175] Sorensson, N. and Een, N., 2005. Minisat v1. 13-a sat solver with conflict-clause minimization. SAT, 2005(53), pp.1-2.

[176] Spirtes, P., and Glymour, C., 1991. An Algorithm for Fast Recovery of Sparse Causal Graphs. *Social Science Computer Review*, vol. 9, Iss. 1, 62–72.

[177] Spirtes, P., Glymour, C., and Scheines, R., 1990. Causality from probability. In *Conference Proceedings: Advanced Computing for the Social Sciences*, Williamsburgh.

[178] Spirtes, P., Glymour, C. and Scheines, R., 1993, *Causation, Prediction, and Search: 1st Edition*, Springer-Verlag, New York.

[179] Spirtes, P., Glymour, C., and Scheines, R., 2000. *Causation, Prediction, and Search: 2nd Edition*, The MIT Press, Cambridge Massachusetts and London, England.

[180] Spirtes, P., and Meek, C., 1995. Learning Bayesian Networks with discrete variables from data. In *Proceedings of the 1st Annual Conference on Knowledge Discovery and Data Mining*, pp. 294–299.

[181] Spirtes, P., Meek, C. and Richardson, T., 1995, August. Causal inference in the presence of latent variables and selection bias. In *Proceedings of the Eleventh conference on Uncertainty in artificial intelligence* (pp. 499-506).

[182] Spirtes, P. and Zhang, J., 2014. A Uniformly Consistent Estimator of Causal Effects under the k-Triangle-Faithfulness Assumption. Statistical Science, 29(4), pp.662-678.

[183] Steck, H. and Jaakkola, T.S., 2002, January. On the Dirichlet prior and Bayesian regularization. In *Proceedings of the 15th International Conference on Neural Information Processing Systems* (pp. 713-720).

[184] Strobl, E.V., Visweswaran, S. and Spirtes, P.L., 2018. Fast causal inference with non-random missingness by test-wise deletion. *International journal of data science and analytics*, 6(1), pp.47-62.

[185] Suter, P., Kuipers, J., Moffa, G. and Beerenwinkel, N., 2021. Bayesian structure learning and sampling of Bayesian networks with the R package BiDAG. arXiv preprint arXiv:2105.00488. Package available at: https://CRAN.R-project.org/package=BiDAG

[186] Suzuki, J., 1993, July. A construction of Bayesian networks from databases based on an MDL principle. In *Proceedings of the Ninth international conference on Uncertainty in artificial intelligence* (pp. 266-273).

[187] Suzuki, J., 1999. Learning Bayesian belief networks based on the minimum description length principle: basic properties. *IEICE transactions on fundamentals of electronics, communications and computer sciences*, 82(10), pp.2237-2245.

[188] Suzuki, J., 2017. An efficient Bayesian network structure learning strategy. *New Generation Computing*, 35(1), pp.105-124.

[189] Tan, X., Gao, X., Wang, Z., Han, H., Liu, X. and Chen, D., 2022. Learning the structure of Bayesian networks with ancestral and/or heuristic partition. Information Sciences, 584, pp.719-751.

[190] Teyssier, M., and Koller, D., 2005. Ordering-based search: A simple and effective algorithm for learning Bayesian networks. In *Proceedings of the 21st Conference on Uncertainty in Artificial Intelligence*, pp. 584–590.

[191] Tian, J. and He, R., 2009, June. Computing posterior probabilities of structural features in Bayesian networks. In Proceedings of the Twenty-Fifth Conference on Uncertainty in Artificial Intelligence (pp. 538-547).

[192] Triantafillou, S. and Tsamardinos, I., 2015. Constraint-based causal discovery from multiple interventions over overlapping variable sets. The Journal of Machine Learning Research, 16(1), pp.2147-2205

[193] Triantafillou, S. and Tsamardinos, I., 2016, June. Score-based vs Constraint-based Causal Learning in the Presence of Confounders. In *CFA@ UAI* (pp. 59-67).

[194] Trösser, F., de Givry, S. and Katsirelos, G., 2021, August. Improved Acyclicity Reasoning for Bayesian Network Structure Learning with Constraint Programming. In 30th International Joint Conference on Artificial Intelligence (IJCAI-21).

[195] Tsamardinos, I., Brown, L.E. and Aliferis, C.F., 2006. The max-min hill-climbing Bayesian network structure learning algorithm. *Machine learning*, 65(1), pp.31-78.

[196] Tsamardinos, I., Aliferis, C.F. and Statnikov, A., 2003, August. Time and sample efficient discovery of Markov blankets and direct causal relations. In *Proceedings of the ninth ACM SIGKDD international conference on Knowledge discovery and data mining* (pp. 673-678).

[197] Tsirlis, K., Lagani, V., Triantafillou, S. and Tsamardinos, I., 2018. On scoring maximal ancestral graphs with the max–min hill climbing algorithm. *International Journal of Approximate Reasoning*, 102, pp.74-85.

[198] Tu, R., Zhang, C., Ackermann, P., Mohan, K., Kjellstrom, H. and Zhang, K., 2019, April. Causal discovery in the presence of missing data. In *The 22nd International Conference on Artificial Intelligence and Statistics* (pp. 1762-1770). PMLR.

[199] Ueno, M., 2010. Learning networks determined by the ratio of prior and data. In *Proceedings of the 26thConference on Uncertainty in Artificial Intelligence*, pp. 598–605.

[200] van Beek, P., and Hoffmann, H. F., 2015. Machine learning of Bayesian networks using constraint programming. In *Proceedings of the International Conference on Principles and Practice of Constraint Programming*, pp. 429–445.

[201] Van Haaren, J. and Davis, J., 2012, July. Markov network structure learning: A randomized feature generation approach. In Twenty-Sixth AAAI Conference on Artificial Intelligence.

[202] Verma, T. and Pearl, J., 1990, July. Equivalence and synthesis of causal models. In *Proceedings of the Sixth Annual Conference on Uncertainty in Artificial Intelligence* (pp. 255-270). Elsevier Science Inc..

[203] Viinikka, J., Hyttinen, A., Pensar, J. and Koivisto, M., 2020. Towards scalable bayesian learning of causal dags. *Advances in Neural Information Processing Systems*, 33, pp.6584-6594.

[204] Vitolo, C., Scutari, M., Ghalaieny, M., Tucker, A. and Russell, A., 2018. Modeling air pollution, climate, and health data using Bayesian Networks: A case study of the English regions. *Earth and Space Science*, 5(4), pp.76-88.

[205] Vowels, M.J., Camgoz, N.C. and Bowden, R., 2021. D'ya like DAGs? A Survey on Structure Learning and Causal Discovery. *arXiv preprint* arXiv:2103.02582.

[206] Wang, Z., Gao, X., Yang, Y., Tan, X. and Chen, D., 2021. Learning Bayesian networks based on order graph with ancestral constraints. *Knowledge-Based Systems*, 211, p.106515.

[207] Wong, M.L. and Leung, K.S., 2004. An efficient data mining method for learning Bayesian networks using an evolutionary algorithm-based hybrid approach. *IEEE transactions on evolutionary computation*, 8(4), pp.378-404.

[208] Yang, C., Ji, J., Liu, J., Liu, J. and Yin, B., 2016. Structural learning of Bayesian networks by bacterial foraging optimization. *International Journal of Approximate Reasoning*, 69, pp.147-167.

[209] Yang, J., Li, L. and Wang, A., 2011. A partial correlation-based Bayesian network structure learning algorithm under linear SEM. *Knowledge-Based Systems*, 24(7), pp.963-976.

[210] Yaramakala, S. and Margaritis, D., 2005, November. Speculative Markov blanket discovery for optimal feature selection. In *Fifth IEEE International Conference on Data Mining (ICDM'05)* (pp. 4-pp). IEEE.

[211] Yehezkel, R. and Lerner, B., 2009. Bayesian Network Structure Learning by Recursive Autonomy Identification. *Journal of Machine Learning Research*, 10(7).

[212] Yuan, C., Malone, B. and Wu, X., 2011, June. Learning optimal Bayesian networks using A* search. In *Twenty-Second International Joint Conference on Artificial Intelligence*.

[213] Zanga, A., Ozkirimli, E. and Stella, F., 2022. A survey on causal discovery: Theory and practice. *International Journal of Approximate Reasoning*.

[214] Zhang, J., 2008a. Causal reasoning with ancestral graphs. *Journal of Machine Learning Research*, 9(Jul), pp.1437-1474.

[215] Zhang, J., 2008b. On the completeness of orientation rules for causal discovery in the presence of latent confounders and selection bias. Artificial Intelligence, 172(16-17), pp.1873-1896.

[216] Zhang, J. and Spirtes, P., 2008. Detection of unfaithfulness and robust causal inference. *Minds and Machines*, 18(2), pp.239-271.

[217] Zhang, K., Gong, M., Ramsey, J., Batmanghelich, K., Spirtes, P. and Glymour, C., 2018. Causal Discovery with Linear Non-Gaussian Models under Measurement Error: Structural Identifiability Results. In *UAI* (pp. 1063-1072).

[218] Zhang, K. and Hyvärinen, A., 2009, June. On the Identifiability of the Post-Nonlinear Causal Model. In 25th *Conference on Uncertainty in Artificial Intelligence* (UAI 2009) (pp. 647-655). AUAI Press..

[219] Zhao, J. and Ho, S.S., 2019. Improving Bayesian network local structure learning via data-driven symmetry correction methods. *International Journal of Approximate Reasoning*, 107, pp.101-121.

[220] Zheng, X., Aragam, B., Ravikumar, P.K. and Xing, E.P., 2018. DAGs with NO TEARS: Continuous Optimization for Structure Learning. Advances in Neural Information Processing Systems, 31, pp.9472-9483

# APPENDIX A: GLOSSARY OF SYMBOLS

| Symbol | Meaning |
|---|---|
| $\perp$ | Independent of, e.g. $A \perp B$ means "$A$ is independent of $B$" |
| $\not\perp$ | Not independent of, e.g. $A \not\perp B$ means "$A$ is not independent of $B$" |
| $\mid$ | Given or conditional on, e.g. $A \mid B$ means "$A$ conditional on (or given) $B$" |
| $\setminus$ | Substract from set. e.g. $\boldsymbol{X} \setminus \{A, B\}$ means set $\boldsymbol{X}$ with elements $A$ and $B$ removed |
| $\subset$ | Is a strict subset of, e.g. $\boldsymbol{S} \subset \boldsymbol{X}$ means that all the elements of $\boldsymbol{S}$ are in $\boldsymbol{X}$, and $\boldsymbol{X}$ contains one or more elements not in $\boldsymbol{S}$. |
| $\subseteq$ | Is a subset of, e.g. $\boldsymbol{S} \subseteq \boldsymbol{X}$ means that all the elements of $\boldsymbol{S}$ are in $\boldsymbol{X}$, but $\boldsymbol{X}$ may be the same as $\boldsymbol{S}$. |
| $\mathcal{B}$ | A Bayesian Network |
| $G$ | A graph, typically the DAG in a Bayesian Network |
| $\boldsymbol{\Theta}$ | The set of parameters defining the strength of the relationships between variables |
| $F$ | The number of free parameters in the Bayesian Network, that is, in $\lvert \boldsymbol{\Theta} \rvert$ |
| $n$ | Number of nodes (vertices) in the DAG |
| $\lvert G_n \rvert$ | Number of possible different graphs containing $n$ nodes |
| $\boldsymbol{X}$ | Set of nodes in a DAG, $\boldsymbol{X} = \{X_1, X_2, \dots, X_n\}$ representing the variables being modelled |
| $X_i, A, B, etc$ | An individual node in the Bayesian Network's DAG representing a variable |
| $x_i, A, B$ | The value of the variable $X_i, A, B$ respectively |
| $P(\boldsymbol{X})$ | The joint probability distribution over the variables represented by the nodes |
| $\boldsymbol{E}$ | Set of edges in a graph |
| $\lvert E \rvert$ | Number of edges in the graph |
| $\lvert M \rvert$ | Number of missing (absent) edges, that is, the number of edges in the complete graph minus $\lvert E \rvert$ |
| $\boldsymbol{S}$ | Set of nodes in a separating set, $\boldsymbol{S} = \{s_1, \dots, s_q\}$ |
| $\boldsymbol{Pa}(X_i)$ | The set of nodes that are direct parents of node $X_i$ |
| $\boldsymbol{De}(X_i)$ | The set of nodes that are descendants of node $X_i$ |
| $\boldsymbol{MB}(T)$ | The Markov Blanket of node $T$ |
| $\boldsymbol{PC}(T)$ | The parents and children of node $T$ |
| $\boldsymbol{D}$ | Dataset from which the graph will be learnt |
| $\boldsymbol{d_m}$ | Individual data instance (i.e. row or case) within the dataset |
| $N$ | Number of data instances (cases) in the data set, $\boldsymbol{D}$ |
| $\alpha$ | Significance level used in Conditional Independence tests. |
| $\epsilon$ | Threshold level used in Mutual Information tests |
| $df$ | Degrees of freedom used in statistical tests |
| $i$ | Index over nodes in the DAG, $i = 1..n$ |
| $r_i$ | Number of different values (states) at node (variable) $X_i$ |
| $k$ | Index over possible values at a node, $k = 1..r_i$ at node $X_i$ |
| $q_i$ | Number of unique combinations of values of the parents $\pi_i$ of node $X_i$ |
| $j$ | Index over combinations of parental values, $j = 1..q_i$ at node $X_i$ |
| $N_{ijk}$ | Number of data instances with $k^{th}$ value, and $j^{th}$ combination of parental values at node $X_i$ in data set $D$ |
| $\theta_{ijk}$ | The conditional probability of node $X_i$ having value $x_k$ conditional on the parents of $X_i$ having the $j^{th}$ combination of parental values. |
| $N', N'_{ijk}, \theta'_{ijk}$ | As $N, N_{ijk}, \theta_{ijk}$ but applying to a prior belief of the parameters. |
| $\rho_{ab}$ | Partial correlation between values $\boldsymbol{a}$ and $\boldsymbol{b}$ |
| $TP, FP, TN, FN$ | True positive, false positive, true negative and false negative metrics |
| $P$ | Precision metric |
| $R$ | Recall metric |
| $G^2$ | The G-squared test statistic |
| $\chi^2$ | The Chi-squared test statistic |
| $MI(A, B)$ | The mutual information between random variables $A$ and $B$ |
| $\zeta_N^r$ | The stochastic complexity of $N$ values of a discrete variable with $r$ states |